\documentclass[10pt,journal,compsoc]{IEEEtran}
\newif\ifpeerreview

\peerreviewfalse
\usepackage[nocompress]{cite}
\usepackage{url}
\usepackage{hyperref}
\usepackage{xurl}
\usepackage{amsmath,amssymb,graphicx}
\usepackage{placeins}
\usepackage{amsmath}
\usepackage{amssymb}
\usepackage[nocompress]{cite}
\usepackage{amsmath,amssymb,graphicx}
\usepackage{bm}
\usepackage{cleveref}
\usepackage{xcolor}
\usepackage{algpseudocode}
\usepackage{algorithm}

\usepackage{mathtools}
\usepackage{longtable}
\usepackage{booktabs} 
\usepackage{array}    
\usepackage{verbatim}
\usepackage{cancel}
\usepackage{svg}
\newcommand{\densedots}{\cdot\mkern-5mu\cdot\mkern-5mu\cdot}
\newcommand{\minus}{\mkern-5mu-\mkern-5mu}
\newcommand{\plus}{\mkern-5mu+\mkern-5mu}
\usepackage{siunitx}
\usepackage{etoolbox} 

\robustify\bfseries % Makes bolding work inside S columns

\usepackage{lipsum} % Only used to generate random text.

\usepackage[switch]{lineno}

\newcommand{\paperID}{5}

\title{Fundamental Recovery Bounds for SPAD Signals under Stationary Flux}
\author{Lior Dvir, Nadav Torem, Mohit Gupta, and Yoav Y. Schechner%
\IEEEcompsocitemizethanks{\IEEEcompsocthanksitem Lior Dvir, Nadav Torem, and Yoav Y. Schechner are with the Viterbi Faculty of Electrical and Computer Engineering, Technion-Israel Institute of Technology, Haifa 3200003, Israel.
\IEEEcompsocthanksitem Mohit Gupta is with the Department of Computer Sciences, University of Wisconsin-Madison, Madison, WI 53706, USA.}}

\begin{document}

\IEEEtitleabstractindextext{%
\begin{abstract}
Single-photon avalanche diodes (SPADs) record light as a discrete stream of individual detections. The signal is stochastic. Its statistical structure depends on the sensor's operation mode: binary detection in fixed bins, timestamped detection in fixed bins, or free-running timestamped detection. We derive the likelihood score function for each of these three passive modes. From this single object, stem both fundamental limits of recovery (Cramér–Rao bounds) and practical recovery algorithms based on diffusion posterior sampling. The paper further generalizes fundamental limits 
to Bayesian Cramér–Rao lower bounds. This generalization makes use  of a learned approximation of the score function of signal priors. 
In prior art, analyses and diffusion-based reconstruction for SPAD data have treated individual modes in isolation. Our unified treatment shows a qualitative high-flux gap between modes: binary counts saturate exponentially, while timestamped modes degrade only linearly. We further extend diffusion posterior sampling, previously restricted to binary SPAD data, to a full timestamped case using the suitable score function. We demonstrate experimentally that matching the score to the operation mode is beneficial for high-fidelity reconstruction. By tying the recovery bounds and  diffusion to the % single 
score function, this work aims to establish a common foundation for both asking what is recoverable in single-photon sensing, and building methods that approach %it 
the bound.
\end{abstract}

}

% Make Title
\ifpeerreview
\linenumbers \linenumbersep 15pt\relax 
\author{Paper ID \paperID\IEEEcompsocitemizethanks{\IEEEcompsocthanksitem This paper is under review for ICCP 2026 and the PAMI special issue on computational photography. Do not distribute.}}
\markboth{Anonymous ICCP 2026 submission ID \paperID}%
{}
\fi
\maketitle

% The first section title should be wrapped inside a \IEEEraisesectionheading as follows.
\IEEEraisesectionheading{
  \section{Introduction}\label{sec:introduction}
}

\IEEEPARstart{S}{ingle-photon} avalanche diode (SPAD) arrays offer unique sensing capabilities that extend well beyond conventional sensors. Unlike standard detectors that accumulate charge, SPADs operate in Geiger-mode, enabling the detection of individual photon arrival events with picosecond-level timing precision and negligible read noise~\cite{charbon2014single}. Such extreme sensitivity and high temporal resolution has enabled advances in fluorescence lifetime imaging microscopy (FLIM)~\cite{Zickus2020, Isbaner:16}, LiDAR~\cite{Baek2023, Gupta:2019, Incoronato2021, Lee2023}, non-line-of-sight imaging~\cite{bruschini2019single, otoole2018confocal, FelixHeide2021, Gu2023FastNLOS}, and passive ultra-wideband imaging of dynamic scenes across extreme timescales~\cite{wei2023passive}. 

Single-photon sensitivity has also made passive low-light and high-dynamic-range imaging possible under illumination spanning many orders of magnitude~\cite{fossum2011quanta,vivek2016,ma2020quanta,ingle2019high,ingle2021passive}. As SPAD arrays scale to higher spatial resolution and speeds~\cite{Morimoto:20, Ulku:19}, photon counts per pixel become small.
%and the measurement statistics become strongly non-Gaussian: 
Raw SPAD data are discrete records of detection events, not intensity samples. Recovering a clean image or flux map from such data is a nonlinear inverse problem~\cite{YifanWolfgang2020,roarke2021}, whose structure depends on how the sensor is operated. Current passive SPAD sensors produce raw data in several distinct modes. Binary quanta image sensors report a single detection bit per bin~\cite{fossum2011quanta, ma2020quanta}; free-running timestamped arrays report an arrival time for each detection event~\cite{ingle2019high,wei2023passive}; and bandwidth-limited timestamped arrays report at most one timestamp per fixed time bin~\cite{ingle2021passive}. Each mode yields a different raw signal, with different statistics, hence  a unique relationship between what the sensor records and what can be inferred about the scene.

\begin{figure}[t] 
     \centering
     \includegraphics[width=\columnwidth]{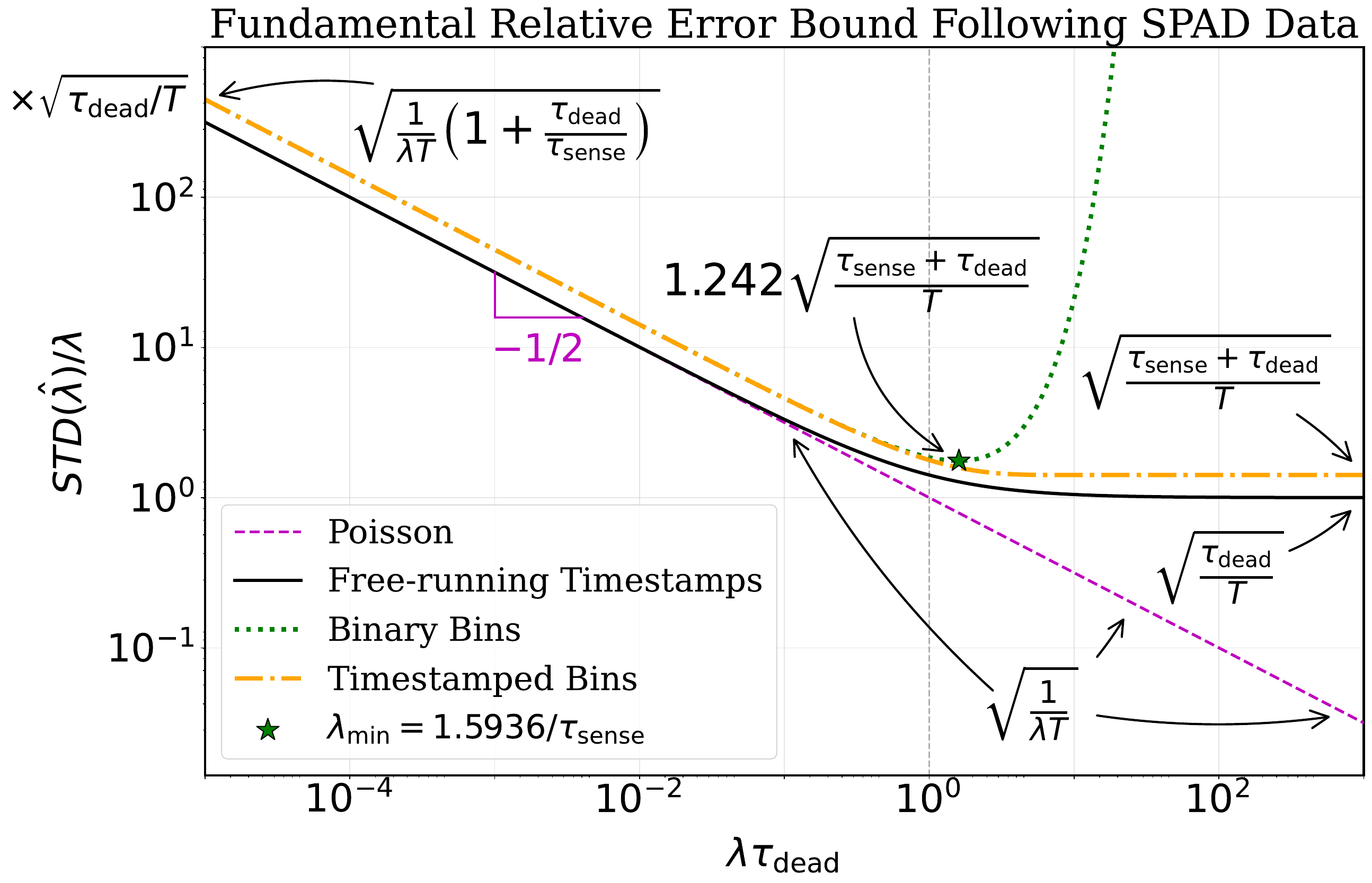}
     \caption{Fundamental relative error bounds for unbiased estimators, based on SPAD measurements of Poissonian light using an independent  pixel. The plots compare  operation modes of {\bf (M1)} {\em Free-running Timestamps}, {\bf (M3)} {\em Timestamped Bins} and {\bf (M2)} {\em Binary Bins}. 
     %For both Discrete Bins and Times in Bins, $\tau_{\rm sense}=\tau_{\rm dead}$. $\textbf{Low event rate:}$ Relative error of Continuous Times and Poisson process converge. Discrete Bins and Times in Bins are bounded from below by Continuous Times. $\textbf{High event rate:}$ Above event rate of $1/\tau_{\rm dead}$, relative error of Continuous Times and Times in Bins converge to the constants. Times in Bins is bounded from below by Continuous times. Discrete Times reached its minimal relative error and then diverges to infinity.
     }
     \label{fig:CRLB}
\end{figure}

Our key observation is that all the modes mentioned above relate to a common statistical object: the likelihood score function — the gradient of the log-likelihood of the raw data with respect to the incident flux. We derive this score function for each of the three modes from first principles. From this single object we derive both  fundamental limits of recovery — Cramér–Rao bounds (Fig.~\ref{fig:CRLB}) — 
and practical recovery algorithms based on diffusion posterior sampling (DPS). Thus, the same score function that characterizes  fundamental limits of recovery also drives modern diffusion-based reconstruction in practice.
This unification allows comparing recovery (Cramér–Rao) bounds for the three modes on one common plot (Fig.~\ref{fig:CRLB}),
thereby revealing a qualitative gap at high flux: the bound for binary counts diverges exponentially, while both timestamped modes grow only linearly. Registering event times is therefore not an incremental gain; it marks a distinction between graceful and rapid saturation at bright illumination. On the algorithmic side, plugging the appropriate score into diffusion posterior sampling gives a reconstruction approach that handles all three modes within a single framework. \smallskip

\noindent {\bf Scope.} In this paper, we restrict attention to passive sensing, where the detector is not synchronized to an active illumination source. The flux is assumed to be stationary; active-mode configurations such as pulsed-laser TCSPC and time-gated LiDAR are beyond our scope. Within this passive setting, the score-based approach provides a compact way to analyze both what is fundamentally recoverable and what an algorithm recovers in practice.

\section{Related Work}\label{sec:related}
\noindent {\bf Passive single-photon imaging}. Passive SPAD sensing has previously been analyzed in the continuous free-running mode~\cite{ingle2019high, ingle2021passive, wei2023passive}, as well as the binary-bin mode with Bernoulli/binomial likelihoods~\cite{fossum2011quanta, ma2020quanta, chan:22}. Wei~et~al.~\cite{wei2023passive} developed a theory for reconstructing time-varying flux from free-running SPAD timestamps, enabling passive imaging across timescales from seconds to picoseconds. Our focus is different: we address static per-pixel flux estimation for spatial image reconstruction, where the unifying object across modes is the likelihood score function. We build on these lines of work and present a unified CRLB analysis across all three passive modes (Sec.~\ref{sec:score}), including the bandwidth-limited timestamped-bin mode which has not received much prior attention.\smallskip
\noindent {\bf Diffusion-based reconstruction for photon-limited data.} Score-based diffusion~\cite{ho2020denoising, song2021score} and diffusion posterior sampling (DPS)~\cite{chung2023diffusion} have become standard tools for nonlinear inverse problems, with recent applications to low-dose CT~\cite{xia2024paralleldiffusionmodelbasedsparseview}, photoacoustic tomography~\cite{dey_2024}, and photon-starved image restoration~\cite{melidonis2025score, Liu_2025_ICCV}. The closest prior work in our setting is~\cite{melidonis2025score}, which develops DPS for binary and count-valued quanta sensing models. We extend DPS to timestamped SPAD data by deriving mode-matched likelihood scores of structurally different form (Eqs.~\ref{eq:HYBRID_binomial_grad_log_likelihood}, \ref{eq:SPAD_FM_grad_log_likelihood}), where, as our CRLB analysis shows, the measurements carry fundamentally more information than in the binary case. \smallskip
\section{Mathematical Preliminaries}
\label{sec:theoback}

We briefly review the SPAD measurement models for the three passive operation modes, and the score-based diffusion approach on which our reconstruction method builds.
\subsection{SPAD Measurement Models} 
\label{sec:photon}

\begin{figure}[t] 
    \centering
    \includegraphics[width=\columnwidth]{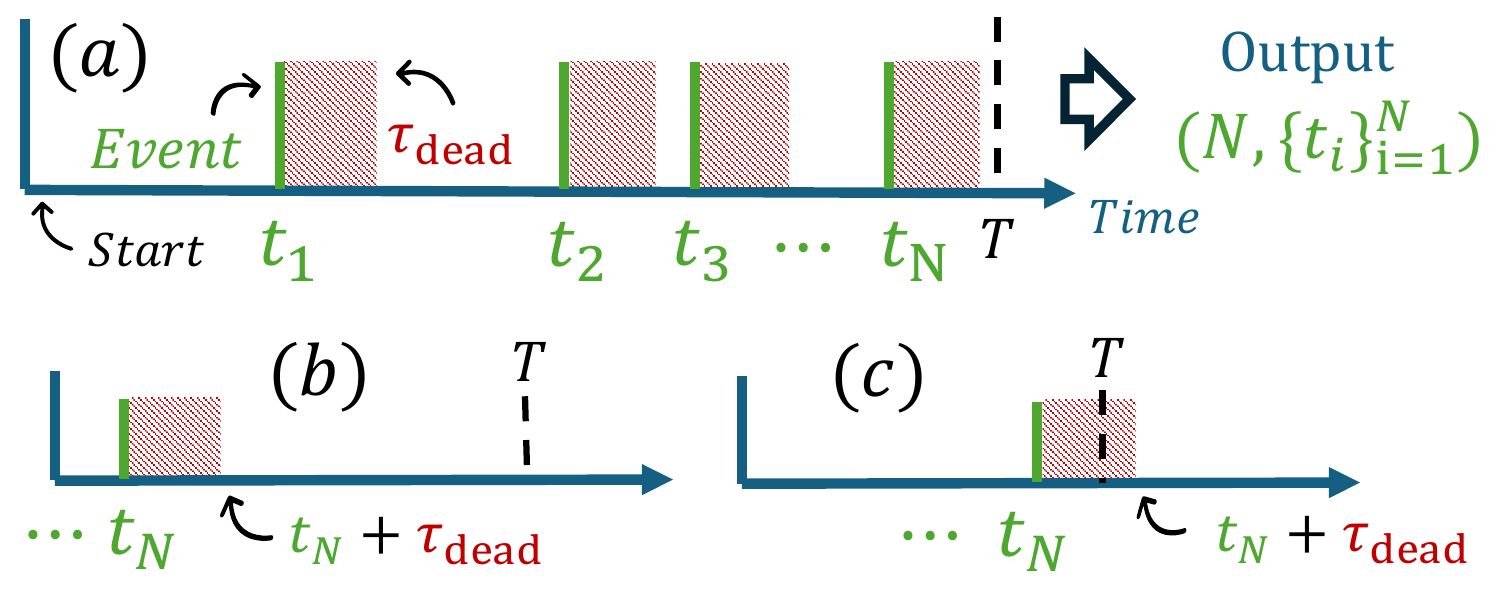}
    \caption{Operation by {\bf (M1)} {\em Free-running Timestamps}. (a) Each green line is a single detected event. A red rectangle is the dead time, spanning $\tau_{\rm{dead}}$. By exposure time $T$, $N$ events are detected, timestamped $\{t_i\}_{i=1}^N$ and reported. (b) {\tt Case I}: The $N$-th event occurs early enough ($t_N +\tau_{\rm{dead}} \leq T$) for further possible detection. (c) {\tt Case II}: Since $t_N +\tau_{\rm{dead}}> T$,  further detection is impossible.}
    \label{fig:temporal_mode}
\end{figure}

\begin{figure}[t] 
    \centering
    \includegraphics[width=\columnwidth]{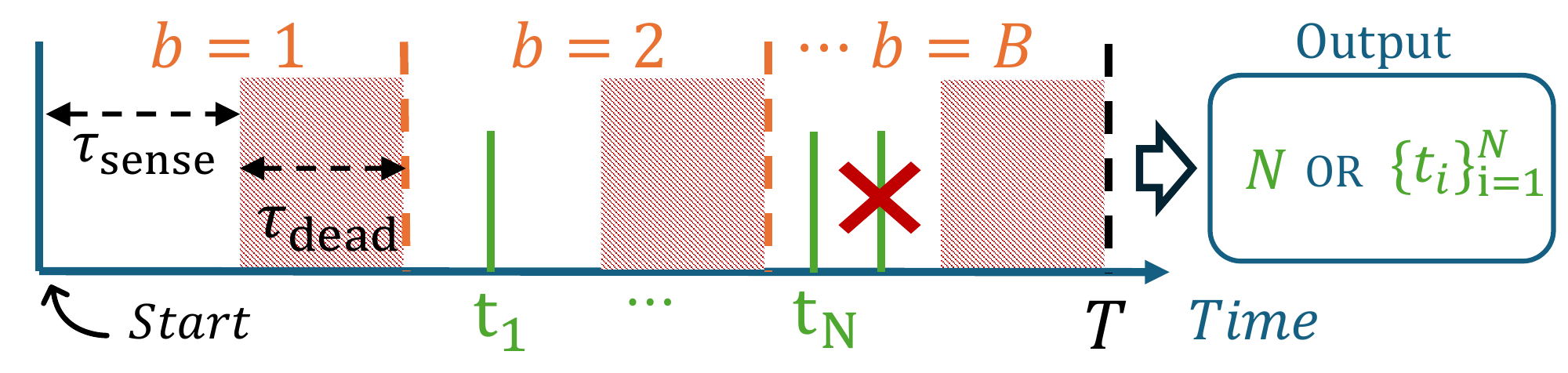}
    \caption{In each discrete time bin $b$, the sensor is sensitive during $\tau_{\rm{sense}}$, then enters a dead time period. Each time bin reports at most one event, no matter the photon flux intensity. The output in mode {\bf (M2)} {\em binary Bins} includes the number of bins with detected events $N$. The output in mode {\bf (M3)} {\em Timestamped Bins} includes the number of bins with detected events $N$, and their associated detection times $\{t_i\}_{i=1}^N$.}
    \label{fig:discrete_mode}
\end{figure}

Consider a pixel exposed to photon flux $\Phi$ [s$^{-1}$] with a known detector quantum efficiency $q \in [0,1]$. Then, incident photon arrivals form a Poisson process with rate $\lambda = q\Phi$. A SPAD detection triggers a dead time $\tau_\text{dead}$ during which the sensor is insensitive. After each dead time, the time to the next detection is exponential with rate $\lambda$, so the inter-detection time $\Delta t$ follows a shifted exponential~\cite{ingle2019high} probability density function (PDF):
\begin{equation}
  f(\Delta t) = \lambda \exp[-\lambda(\Delta t - \tau_\text{dead})], \quad \Delta t \geq \tau_\text{dead}.
  \label{eq:shifted-exp}
\end{equation}
\noindent The three passive operation modes differ in how these detections are reported. Fig.~\ref{fig:temporal_mode}(a) illustrates a detection timeline.
\vspace{0.5em}
\noindent\textbf{(M1) Free-running timestamps (continuous-time).}
The sensor is turned on at $t=0$ and records all detection event times $0 < t_1 < \cdots < t_N \leq T$. Here $T$ is the exposure time and 
\begin{equation}
    N \leq N_\text{max} = \lceil T / \tau_\text{dead} \rceil\;.
    \label{eq:N_max}
\end{equation}
The parameter set is $\mathcal{S}_\text{cont} = \{T, \tau_\text{dead}\}$. The likelihood decomposes into three cases: $N=0$, $N \geq 1$ with room for another detection ($t_N \leq T - \tau_\text{dead}$) as illustrated in Fig.~\ref{fig:temporal_mode}(b), and $N \geq 1$ without (Fig.~\ref{fig:temporal_mode}(c)). The likelihood is~\cite{ingle2021passive} 
\begin{equation}
  p(N, \{t_i\}_{i=1}^N | \lambda, \mathcal{S}_\text{cont}) =
  \begin{cases}
    e^{-\lambda T} & \texttt{Case 0} \\
    \lambda^N e^{-\lambda (T - N\tau_\text{dead})} & \texttt{Case I} \\
    \lambda^N e^{-\lambda [t_N - (N-1)\tau_\text{dead}]} & \texttt{Case II},
  \end{cases}
  \label{eq:likelihood-cont}
\end{equation}
where $\texttt{Case 0}$ denotes $N=0$. We note that $(N, t_N)$ is a sufficient statistic for $\lambda$ --- the earlier timestamps $\{t_i\}_{i<N}$ do not appear in~\eqref{eq:likelihood-cont}.
\vspace{0.5em}
\noindent\textbf{(M2) Binary bins (quanta image sensor mode).}
The exposure $T$ is divided into $B$ bins, each consisting of a sensing window $\tau_\text{sense}$ followed by a dead time, with 
\begin{equation}
    B(\tau_\text{sense} + \tau_\text{dead}) = T\;.
    \label{eq:DISC_T_b}
\end{equation}
 Each bin reports a single bit indicating whether at least one detection occurred during its sensing window. The parameter set is $\mathcal{S}_\text{bin} = \{T, B, \tau_\text{dead}\}$. The per-bin success probability is $p_b$. The total number $N$ of triggered bins is binomial:
\begin{equation}
  p(N| B, p_b) = \binom{B}{N} p_b^N (1 - p_b)^{B-N}.
  \label{eq:likelihood-bin}
\end{equation}
\noindent This operation mode is illustrated in Fig.~\ref{fig:discrete_mode} with output $N$. 
\vspace{0.5em}
\noindent\textbf{(M3) Timestamped bins (bandwidth-limited).}
The bin structure is as in M2, but each triggered bin additionally reports the within-bin arrival time $\tilde{t}_i \in (0, \tau_\text{sense}]$ of its first detection (Fig.~\ref{fig:discrete_mode}). This mode arises when a free-running timestamped channel throttles its output to at most one event per fixed bin due to bandwidth considerations~\cite{seets2020motionadaptivedeblurringsinglephoton}. 

%%%%%%%%%%%%%%%%%%%%%%%%%%%%%%%
\subsection{Score-based Diffusion}
\label{sec:LDbackground}

This paper uses {\em score-based diffusion} (SBD) for image recovery. Background on this method is surveyed in the {\em Supplementary Material}. Here we provide the essential steps.
A true object is expressed by a vector $\bm{x}$, where each element is denoted $x$. 
%For example, $x$ can be the pixel value in a noiseless projected image, and $\bm{x}$ is a two dimensional (2D) noiseless image.
The object $\bm{x}$ is randomly sampled from nature, with a natural PDF denoted $p(\bm{x})$. 
%We do not have hold of this object. 
A forward model $\mathcal{F}$ yields data ${\cal D}=\mathcal{F}(\bm{x})$. We want to recover $\bm{x}$. SBD seeks to generate a sample $\bm{x}_0$ from the posterior distribution $p(\bm{x}|\mathcal{D})$. 
Using Bayes theorem, 
\begin{equation}
    \nabla_{\bm{x}} \log p(\bm{x}|\mathcal{D}) = \nabla_{\bm{x}} \log p(\bm{x}) +  \nabla_{\bm{x}}\log p(\mathcal{D|}\bm{x})\;.
\end{equation}
The term $\nabla_{\bm{x}} \log p({\bm{x}})$ is the {\em score function of the prior term}, and the term $\nabla_{{\bm{x}}} \log p({\cal D}|{\bm{x}})$ is the {\em score function of the likelihood}, also called the log-likelihood gradient. To run an SBD algorithm, both score functions must be computed. 
Computations are in discrete steps indexed  $k\in [K,\dots,1]$  in a countdown. Each step has an intermediate noisy state  $\bm{x}_k$. 
In SBD,  $\nabla_{\bm{x}_k} \log p({\bm{x}}_k)$ is not derived explicitly, but it is approximated by the output $\bm{s}_{\bm{\theta}}({\bm{x}}_k,k)$ of a trained DNN  ({\em score DNN})~\cite{zhu2023denoising}. Its parameters are $\bm{\theta}$. 
Regarding the likelihood score, we use an algorithm termed diffusion posterior sampling (DPS)~\cite{chung2023diffusion}.
Define a schedule $\{\alpha_k\}_{k=1}^K$ and then
\begin{equation}
   \beta_k=1-\alpha_k  \;, ~~~~~~~
    \bar{\alpha}_k=\Pi_{\iota=1}^k\alpha_{\iota}\;.
    \label{eq:BG_alpha_beta}
\end{equation} 
At each iteration, DPS estimates $\bm{x}_0$ as
\begin{equation}
   \hat{\bm{x}}_0({\bm{x}}_k) = \mathbb{E}[\bm{x}_0|\bm{x}_k]=\frac{1}{\sqrt{\bar{\alpha}_k}}\left[ {\bm{x}}_k + (1-\bar{\alpha}_k)\bm{s}_{\bm{\theta}}({\bm{x}}_k,k) \right]\;.
   \label{eq:BG_LIKELIHOOD_x0_estimation}
\end{equation}
%$k$ % \in [K, \dots, 1]$   
Then, $\hat{\bm{x}}_0$ is utilized to approximate the score function of the likelihood:  
$\nabla_{\bm{x}_k} \log p({\mathcal{D}}| \bm{x}_k) \simeq \nabla_{\bm{x}_k} \log p({\mathcal{D}} | \hat{\bm{x}}_0)$. 
This approximation %Computing $\nabla_{\hat{\bm{x}}_0} \log p({\cal D}|\hat {\bm{x}}_0)$ 
is tractable,  because $\hat {\bm{x}}_0$ behaves as a clean object, for which $\log p({\cal D}|\hat {\bm{x}}_0)$ is well defined.
Define a Jacobian for Eq.~(\ref{eq:BG_LIKELIHOOD_x0_estimation}),
%$\bm{J}$ is utilized,
\begin{equation}
    \bm{J}=\frac{\partial\hat{\bm{x}}_0}{\partial\bm{x}_k}=\frac{1}{\sqrt{\bar{\alpha}_k}}\left[ {\bm{I}} + {(1-\bar{\alpha}_k)}\nabla_{\bm{x}_k}\bm{s}_{\bm{\theta}}({\bm{x}}_k,k) \right]\;.
    \label{eq:BG_DPS_J}
\end{equation}
Here $\nabla_{\bm{x}_k}\bm{s}_{\bm{\theta}}({\bm{x}}_k,k)$ is a Jacobian relating the output vector of $\bm{s}_{\bm{\theta}}$ with respect to an input vector ${\bm{x}}_k$. In practice, $\nabla_{\bm{x}_k}\bm{s}_{\bm{\theta}}({\bm{x}}_k,k)$ is calculated through backpropagation, exploiting the fact that  $\bm{s}_{\bm{\theta}}$ is a differentiable DNN. 
Let $\mathcal{N}(\cdot,\cdot)$ be the normal PDF and $\bm{I}$ a unit matrix. 
Let ${\bm \eta}\sim\mathcal{N}(0,\bm{I})$. Define a step size $\rho$.
The iterative rule in DPS follows,
\begin{eqnarray}
        \bm{x'}_{k-1}=
         \frac{\sqrt{\alpha_k}(1-\bar{\alpha}_{k-1})}
              {1-\bar{\alpha}_{k}}\bm{x}_k
        +\frac{\sqrt{\bar{\alpha}_{k-1}}\beta_k}
              {1-\bar{\alpha}_{k}}
        \hat{\bm{x}}_0(\bm{x}_k)+\sigma_k\bm{\eta}
        \\[0.2cm]
        \bm{x}_{k-1}=\bm{x'}_{k-1}+\rho \bm{J} \nabla_{\hat{\bm{x}}_0} \log p[{\cal D}|\hat{\bm{x}}_0(\bm{x}_k)]\;,~~~~~~~~~~~~~~~~~~~~~~
\label{eq:BG_DPS_update}
\end{eqnarray}
where 
$\sigma_K>\sigma_{K-1} \ldots > \sigma_1=0$. %avoiding a collapse to a local maximum throughout the SBD algo
%$\sigma_k\rightarrow 0$. %avoiding a collapse to a local maximum throughout the SBD algorithm. 
Overall, the parameters of the algorithm are 
$\rho,\{ \sigma_k\}_{k=1}^K,\{\alpha_k\}_{k=1}^K$.

%%%%%%%%%%%%%%%%%%%%%%%%%%%%%%%%%%%
\section{Score Functions and Variance Bound}
\label{sec:score}

We now focus on estimation $\hat \lambda$ of the flux, in {\em point-wise} measurements, that can be acquired by a single pixel, without regard to potential spatial statistical dependence. We derive likelihood functions and a couple of their consequences. One consequence is the score function. The score function   is useful for estimation using maximum likelihood (ML), maximum a-posteriori and diffusion models. An additional consequence is a lower bound on the variance of  $\hat \lambda$. This is derived by the Cramér–Rao lower bound (CRLB). Following derivation of the CRLB, we make a fundamental comparison of the different operation modes  and quantify which is better.

Denote raw sensor output data by ${\cal D}$. The sensor has a set of operation parameters ${\cal S}$, as described in Sec.~\ref{sec:photon}.
The likelihood is $p({\cal D}|\lambda,{\cal S})$. The likelihood score function is 
\begin{equation}
    \frac{\partial}{\partial \lambda}
    \log [p({\cal D}|\lambda,{\cal S})]\;.
    \label{eq:whatscore}
\end{equation}
The Fisher information on $\lambda$, based solely on sensor data~\cite{kay1993fundamentals} is
\begin{equation}
    I_{\rm D}(\lambda) = 
      \mathbb{E} 
       \left\{
       \left(  \frac{\partial}{\partial \lambda}
    \log [p({\cal D}|\lambda,{\cal S})] \right)^2 
       \right\} \;,
    \label{eq:fishebasic}
\end{equation}
where $\mathbb{E}$ denotes expectation over random data. Suppose the following conditions hold:\\
i) {Interchangeability}: 
    $~\frac{\partial}{\partial \lambda}
        \int p({\cal D}|\lambda) d{\cal D}
    =\int\frac{\partial}{\partial \lambda} 
       p({\cal D}|\lambda)d{\cal D}$.\\
ii) {Differentiability}: The log-likelihood is twice differentiable with respect to $\lambda$.\\
iii) {Support independence}: Support of 
    $p({\cal D}|\lambda,{\cal S})$ with respect to the data domain
    is independent of $\lambda$.\\
Then, Eq.~(\ref{eq:fishebasic}) can be derived~\cite{kay1993fundamentals} as 
\begin{equation}
     I_{\rm D}(\lambda) =
%     \mathbb{E} \left[ \left( \frac{\partial}{\partial \chi} \log f(\upsilon| \chi) \right)^2 \right] = 
     -\mathbb{E} 
       \left\{
           \frac{\partial^2}{\partial \lambda^2} 
           \log [p({\cal D}|\lambda,{\cal S})] 
        \right\}
    \;.
    \label{eq:fisher}
\end{equation}

Let the estimator $\hat \lambda$ be unbiased~\cite{kay1993fundamentals}, i.e., 
%\begin{equation}
    $\mathbb{E}[\hat{\lambda}]=\lambda$.
%    \label{eq:CRLB_unbias}
%\end{equation}
The CRLB is a lower limit on the variance of any unbiased estimator. It is the reciprocal of the Fisher Information~\cite{kay1993fundamentals}. Thus, for point-wise measurements,  
\begin{equation}
    \text{Var}(\hat{\lambda}) \geq {\rm CRLB} =[I_{\rm D}(\lambda)]^{-1} \;.
    \label{eq:CRLB_fisher}
\end{equation}
We define an estimation signal-to noise ratio (SNR) as the reciprocal of the relative error. The relative error is based on the standard deviation (STD) of the estimate $\hat \lambda$:
\begin{equation}
    \frac{1}{\rm SNR}=
    \frac{{\rm STD}(\hat{\lambda})}{\lambda} \;.
    \label{eq:invSNR}
\end{equation}

%%%%%%%%%%%%%%%%%%%%%
\subsection{Poisson Process}
\label{sec:CRLB_poisson}

A Poissonian process~\cite{feller1968introduction} has no dead time. It is characterized by a discrete probability distribution, to have exactly $N$ events during operation time $T$, for an expected rate $\lambda$: 
\begin{equation}
    %p(N|\lambda,T) = \frac{(\lambda T)^N \exp(- \lambda T)}{N!}.
    p(N|\lambda,T) = (\lambda T)^N \exp(- \lambda T)/{N!} \;.
    \label{eq:SPAD_PAS_poisson_pmf_def}
\end{equation}
%Here $N \in \{0, 1, 2, \dots\}$ represents the number of observed events.
Its expectation is $\mathbb{E}[N]=\lambda T$.
The CRLB of this process should serve as a baseline. 
From~(\ref{eq:SPAD_PAS_poisson_pmf_def}), the score function is 
\begin{equation}
    \frac{\partial\log p(N|T)}
         {\partial\lambda}= \frac{N}{\lambda}-T \;.
    \label{eq:fisher_poisson}
\end{equation}
Differentiating Eq.~(\ref{eq:fisher_poisson}) as a function of $\lambda$ and using  Eqs.~(\ref{eq:fisher},\ref{eq:CRLB_fisher}),
$\text{Var}(\hat{\lambda})\geq {\rm CRLB}= 
        \left[\mathbb{E}\{N\}/\lambda^2
         \right]^{-1} =\lambda/T$. 

The (STD) of the estimate $\hat \lambda$ thus satisfies
\begin{equation}
    {\rm STD}(\hat\lambda)\geq \sqrt{\lambda/T} \;.
    \label{eq:std_poisson}
\end{equation}

%%%%%%%%%%%%%%%%%%%%%%%%%%%%%%%%%%%%%%%%%%
\subsection{Binary Readouts in Discrete Time Bins}
\label{sec:discrete_model}

%\subsubsection{Binary Readouts in Discrete Time Bins}
In this section, we derive the likelihood, score function and CRLB for the
{\bf M2} operation mode of Sec.~\ref{sec:photon}, working with {\em Binary Bins}. The probability to have an event in bin $b$ is 
%From Eq.~(\ref{eq:SPAD_IAES_exp_pdf}), 
\begin{equation}
    p_b=1-\exp\left[-\lambda{\tau_{\rm{sense}}}\right]\;.
    \label{eq:DISC_p}
\end{equation}
Using Eqs.~(\ref{eq:likelihood-bin},\ref{eq:DISC_p}), the discrete log likelihood for $\lambda>0$ is
\begin{equation}
\begin{split}
    \log p(N|\lambda,{\cal S}_{\rm bin})
    %=
%    ~~~~~~~~~~~~~~~~~~~~~~~~~~~~~~~~~~~~~~~~~~~~~
%    \\
    %&=\log \binom{B}{N}p_b^N(1-p_b)^{B-N}\\
    %&
    =\log \binom{B}{N} ~~~~~~~~~~~~~~~~~~~~~~~~~~~~~\\
    + N\log (1-\exp\left[-\lambda{\tau_{\rm{sense}}}\right])
     -(B-N)\lambda{\tau_{\rm{sense}}}\;.
\end{split}
\label{DISC_log_likelihood}
\end{equation}
Hence, for $\lambda>0$, the likelihood score function is
%Using Eqs.~(\ref{eq:DISC_p},\ref{DISC_log_likelihood}) and a some algebra, we derive
%he gradient of the log likelihood is
\begin{equation}
\begin{split}
    \displaystyle 
    \frac{\partial\log p(N|\lambda,{\cal S}_{\rm bin})}
         {\partial\lambda}
    \displaystyle = 
       \frac{N\tau_{\rm{sense}}}
            {\exp{(\lambda \tau_{\rm sense})}-1}
        -\tau_{\rm{sense}}(B-N)\;.
  \end{split}
  \label{DISC_grad_log_likelihood}
\end{equation}

Differentiating %Eq.~
(\ref{DISC_grad_log_likelihood}) as a function of $\lambda$ and using   %Eqs.~
(\ref{eq:fisher},\ref{eq:CRLB_fisher}),
\begin{equation}
    \begin{split}
        %I(\lambda)=
        {\rm CRLB}=
        %-\mathbb{E} \left[ \frac{\partial^2}{\partial \lambda^2} \log p(N,B,T|\lambda) \right]\\[10pt]
        %=-\mathbb{E} \left[ \frac{\partial^2}{\partial \lambda^2} \log \{\binom{B}{N}(1-\exp[-\lambda T_{\rm b}])^N\exp[-\lambda T_{\rm b}]^{B-N}\}\right]\\[10pt]
        %=-\mathbb{E} \left[ \frac{\partial}{\partial \lambda} \left(\frac{N T_{\rm b}}{\exp[\lambda T_{\rm b}]-1}-T_{\rm{b}}\{B-N\}\right)\right]\\[10pt]
         &\left[
         -\mathbb{E} 
          \left\{
               -\frac{N\tau_{\rm sense}^2\exp(\lambda \tau_{\rm sense})}
                     {[\exp(\lambda \tau_{\rm sense})-1]^2}
          \right\}
          \right]^{-1}
          \\[5pt]
        =&\frac{[\exp(\lambda \tau_{\rm sense})-1]^2}
               {\tau_{\rm sense}^2\exp(\lambda \tau_{\rm sense})}             
         \left[ 
            \mathbb{E}\{N\}
         \right]^{-1}\;.
    \end{split}
    \label{eq:fisher_discrete_E_N}
\end{equation}      
 %Binomial distribution with 
In $B$ Bernoulli trials, each with success probability $p_b$, 
the expectation is
$p_bB$. Hence, from Eq.~(\ref{eq:DISC_p}),
\begin{equation}
    \mathbb{E}[N]%=B p_b 
    = B [1-\exp (-\lambda  \tau_{\rm sense})] \;.
    \label{eq:CRLB_disc_expectation}
\end{equation}
Substituting Eq.~(\ref{eq:CRLB_disc_expectation}) in Eq.~(\ref{eq:fisher_discrete_E_N}) yields 
\begin{equation}
       \text{Var}(\hat{\lambda}) \geq {\rm CRLB}
        =\frac{\exp(\lambda \tau_{\rm sense})-1}{\tau_{\rm sense}^2B} \;.
    \label{eq:CRLM_discrete_formula}
\end{equation}

The result is consistent with familiar statistical trends. First, the variance decreases linearly with $B$, which is consistent with statistics of independent measurements. Second, consider the  standard deviation (STD) of $\hat \lambda$ when 
the rate is low, ie., $\lambda\ll 1/\tau_{\rm sense}$.
Then, from Eqs.~(\ref{eq:DISC_T_b},\ref{eq:CRLM_discrete_formula}),
\begin{equation}
       \text{STD}(\hat{\lambda}) 
       \geq \sqrt{\rm CRLB} 
       \xrightarrow
       {\lambda \tau_{\rm sense} \ll 1 }
        \sqrt{\frac{\lambda}{\tau_{\rm sense} B}}
        =
        \sqrt{\frac{\lambda}{T} 
        (1+\frac{\tau_{\rm dead}}{\tau_{\rm sense}})}
     \label{eq:STD_lowdiscrete}
\end{equation}
The increase of STD with $\sqrt\lambda$ follows the trend of Poissonian statistics. It fully agrees with Eq.~(\ref{eq:std_poisson}) when $\tau_{\rm dead}=0$.
Indeed, when the rate is low, the mean time between events is much larger than 
$\tau_{\rm sense}$ and $\tau_{\rm dead}$, yielding 
effectively a Poisson process. Furthermore, the results degrade as $\tau_{\rm sense}$ decreases, which is consistent with effective shortening of the  sensor exposure to light. 
On the other hand, when $\lambda \tau_{\rm sense}\gg 1$, the variance lower bound in Eq.~(\ref{eq:CRLM_discrete_formula}) blows exponentially with $\lambda$. In the limit of very high photon flux, all bins tend to report events, leading (softly) to saturation. Obviously, with advance into saturation, the ability to estimate $\lambda$ greatly diminishes.  

It is useful to bound the optimal SNR in this operation mode.
Based on Eqs.~(\ref{eq:invSNR},\ref{eq:CRLM_discrete_formula}), an optimum is reached when
\begin{equation}
    0=
    \frac{\partial}{\partial({\lambda \tau_{\rm sense}})}
    \frac{\exp(\lambda \tau_{\rm sense})-1}{(\lambda \tau_{\rm sense})^2B}
    \;.
    \label{eq:dsnr}
\end{equation}
It is easy to show that the solution should satisfy 
\begin{equation}
    (\lambda \tau_{\rm sense}-2)
    \exp(\lambda \tau_{\rm sense}-2)
    =-2\exp(-2)
    \;.
    \label{eq:dzmin}
\end{equation}
For an arbitrary $z$, a solution to the equation $z\exp z=a$ is  provided~\cite{Corless1996LambertW} by a branch denoted $W_0$ of the Lambert function $W(a)$. Thus, a non-trivial solution to Eq.~(\ref{eq:dzmin}) is\footnote{We used the $\textbf{lambertw}$ python function from $\textbf{scipy.special}$ package to validate the results.}
\begin{equation}
    \widetilde{\lambda \tau_{\rm sense}}=2+W_0[-2\exp(-2)]\approx1.5936\;.
    \label{eq:MIN_z_min}
\end{equation}
It is simple to show that this solution is the only minimum of the relative error (maximum SNR) bound. Using this value in Eq.~(\ref{eq:CRLM_discrete_formula}), and making use of Eq.~(\ref{eq:DISC_T_b}), the optimal relative error is bounded by
\begin{equation}
     \frac{{\rm STD(\hat \lambda)}}
          {\widetilde \lambda}
      \geq \frac{\sqrt{\exp \widetilde{\lambda \tau_{\rm sense}}-1}}
            {\widetilde{\lambda \tau_{\rm sense}}}
      \frac{1}{\sqrt B}      
    =1.242\sqrt{\frac{\tau_{\rm sense}+\tau_{\rm dead}}{T}}\;.
     \label{eq:TIME_}
\end{equation}

%%%%%%%%%%%%%%%%%%%%%%%%%%%%%%%%%%%%%%%%%%
\subsection{Continuous Time Readouts in Discrete Time Bins}
\label{sec:hybrid_model}

We now derive the likelihood score function and CRLB for the 
{\bf M3} operation mode, working with {\em Timestamped Bins}.
Suppose a bin indexed $i$ has a detection. Within this bin, the detection time is  $0< \tilde t_i\leq\tau_{\rm sense}$. Given that the bin had a detection, the conditional PDF of ${\tilde t}_i$ is 
\begin{equation}
    p({\tilde t}_i|\lambda,{\cal S}_{\rm bin}) 
     =\frac{\lambda\exp ( -\lambda {\tilde t}_i)}{1-\exp (-\lambda \tau_{\rm{sense}})}\;,
    \label{eq:HYBRID_first_photon_prob}
\end{equation}
where the denominator $\int_{0}^{\tau_{\rm{sense}}}\lambda\exp \left[ -\lambda \tau\right]d\tau$ is a normalization factor, because of the truncated bin time. 

There are $N$ time bins with a detected event.
Their probability of $N$ follows  Eqs.~(\ref{eq:likelihood-bin},\ref{eq:DISC_p}).
Therefore, using Eqs.~(\ref{eq:likelihood-bin},\ref{eq:DISC_p},\ref{eq:HYBRID_first_photon_prob}), the data likelihood is 
\begin{equation}
     p(\{{\tilde t}_i\}_{i=1}^N|\lambda,{\cal S}_{\rm bin}) 
     =\binom{B}{N}
      \left(
       e^{-\lambda\tau_{\rm{sense}}}
     \right)^{B-N} 
     \mkern-5mu\left(
       \lambda^N e^{-\lambda \sum_{i=1}^{N}{\tilde t}_i}
     \right).
    \label{eq:HYBRID_binomial_likelihood}
\end{equation}

%\begin{equation}
%\begin{split}
%     p&(\{{\tilde t}_i\}_{i=1}^N|\lambda,{\cal S}_{\rm bin}) 
%     \\
%     &=\binom{B}{N}\left(\exp\left[-\lambda\tau_{\rm{sense}}\right]\right)^{B-N} \left(\lambda\exp \left[ -\lambda {\tilde t}_i\right]\right)^N\;.
%\end{split}
%    \label{eq:HYBRID_binomial_likelihood}
%\end{equation}
From Eq.~(\ref{eq:HYBRID_binomial_likelihood}), the log likelihood for $\lambda>0$ is 
\begin{equation}
    \begin{split}
        &\log p(\{{\tilde t}_i\}_{i=1}^N|\lambda,{\cal S}_{\rm bin})
        \\&=\log \binom{B}{N}
        +N\log \lambda
        %\\&
        -\lambda\left[\sum_{i=1}^{N}{\tilde t}_i+(B-N)\tau_{\rm{sense}}\right].
    \end{split}
    \label{eq:HYBRID_binomial_log_likelihood}
\end{equation}
Hence, for $\lambda>0$, the  likelihood score function is
%Using Eqs.~(\ref{eq:DISC_p},\ref{DISC_log_likelihood}) and a some algebra, we derive
%he gradient of the log likelihood is
\begin{equation}
%    \begin{split}
        \frac{\partial\log p(\{{\tilde t}_i\}_{i=1}^N|\lambda,{\cal S}_{\rm bin})}
         {\partial\lambda}=
         \frac{N}{\lambda}-(B-N)\tau_{\rm{sense}}
         - \sum_{i=1}^{N}{\tilde t}_i\;.
%    \end{split}
    \label{eq:HYBRID_binomial_grad_log_likelihood}
\end{equation}

Differentiating %Eq.~
(\ref{eq:HYBRID_binomial_grad_log_likelihood}) as a function of $\lambda$ and using  %Eqs.~
(\ref{eq:fisher},\ref{eq:CRLB_fisher}),
\begin{equation}
    \begin{split}
        %I(\lambda)=
        {\rm CRLB}=
         &\left[
         \mathbb{E} 
          \left\{
               N/\lambda^2
          \right\}
          \right]^{-1}
        =\lambda^2             
         \left[ 
            \mathbb{E}\{N\}
         \right]^{-1}\;.
    \end{split}
    \label{eq:CRLB_fisher_hybrid}
\end{equation}  
The expected number of detected events $\mathbb{E}[N]$ is the same as Eq.~(\ref{eq:CRLB_disc_expectation}).
Substituting Eq.~(\ref{eq:CRLB_disc_expectation}) in Eq.~(\ref{eq:CRLB_fisher_hybrid}) yields 
\begin{equation}
    \begin{split}
       \text{Var}(\hat{\lambda}) \geq {\rm CRLB}
        =\frac{\lambda^2}{ B [1-\exp (-\lambda  \tau_{\rm sense})]} \;.
    \end{split}
    \label{eq:CRLM_discreteT}
\end{equation}
This result as well is consistent with familiar statistical trends. The variance decreases linearly with $B$. Moreover, when ${\lambda \tau_{\rm sense} \ll 1 }$, Eq.~(\ref{eq:CRLM_discreteT}) leads to 
Eq.~(\ref{eq:STD_lowdiscrete}). 
On the other hand, when $\lambda \tau_{\rm sense}\gg 1$,  Eqs.~(\ref{eq:DISC_T_b},\ref{eq:CRLM_discreteT}) yield
\begin{equation}
       \text{STD}(\hat{\lambda}) 
       \geq \sqrt{\rm CRLB} 
       \xrightarrow
       {\lambda \tau_{\rm sense} \gg 1 }
        %\lambda /\sqrt{B} =
        \lambda
        \sqrt{\frac{\tau_{\rm sense}+\tau_{\rm dead}}{T}}\;.
     \label{eq:STD_highdiscreCt}
\end{equation}
So, the estimation error increases linearly with 
$\lambda$. This is a much more gentle trend than the exponential divergence of Sec.~\ref{sec:discrete_model}. Hence, at high flux, the reporting of event times dramatically improves the estimation statistical error. Even if all bins are saturated (reporting an event), the event times in them carry critical information.

%%%%%%%%%%%%5555
\subsection{Continuous Time Readouts and Domain}
\label{sec:SPAD}

We now derive the likelihood score function and CRLB for the 
{\bf M1} operation mode, working with {\em Free-running Timestamps}.
Following the model of Eq.~(\ref{eq:likelihood-cont}), for $\lambda>0$ 
\begin{equation}
    \begin{split}
        &\log ~p(N, t_N|\lambda,{\cal S}_{\rm cont})=
        \\%[10pt]
        &= \begin{cases} 
       % \displaystyle -\lambda T  & \texttt{Cases 0} \\[10pt]
        \displaystyle N\log \lambda -\lambda (T -N\tau_{\rm{dead}}) &\texttt{Cases 0,I}  
        \\%[10pt]
        \displaystyle N\log \lambda -\lambda \{t_N-(N-1)\tau_{\rm{dead}}\} & \texttt{Case II}  \;.
        \end{cases}
    \end{split}
    \label{eq:SPAD_FM_log_likelihood}
\end{equation}
Hence, for $\lambda>0$, the  likelihood score function is 
\begin{equation}
    \begin{split}
            \frac{\partial\log p(N,t_N|\lambda,{\cal S}_{\rm cont})}
         {\partial\lambda}
        &= \begin{cases} 
        %\displaystyle -T  & \texttt{Case 0} \\[10pt]
        \displaystyle N/\lambda - T+N\tau_{\rm{dead}} & \texttt{0,I} 
        \\%[2pt]
        \displaystyle N/\lambda - t_N+(N-1)\tau_{\rm{dead}} & \texttt{II} 
        \end{cases}
    \end{split}
    \label{eq:SPAD_FM_grad_log_likelihood}
\end{equation}
Differentiating Eq.~(\ref{eq:SPAD_FM_grad_log_likelihood}) as a function of $\lambda$ and using   
Eqs.~(\ref{eq:fisher},\ref{eq:CRLB_fisher}) %,
yields the expression in Eq.~(\ref{eq:CRLB_fisher_hybrid}).

We now explain how we derive the expected number of detected events $\mathbb{E}[N]$ in this operation mode.

%%%%%%%%%%%%%%%%
\subsubsection*{The Erlang Function}
\label{sec:erlang}

\begin{figure}[t]
    \centering
    % % Top Plot
    % \includegraphics[width=1.0\columnwidth]{figures/poisson_pmf.pdf}\\[5pt]
    % Middle Plot
    \includegraphics[width=1.0\columnwidth]{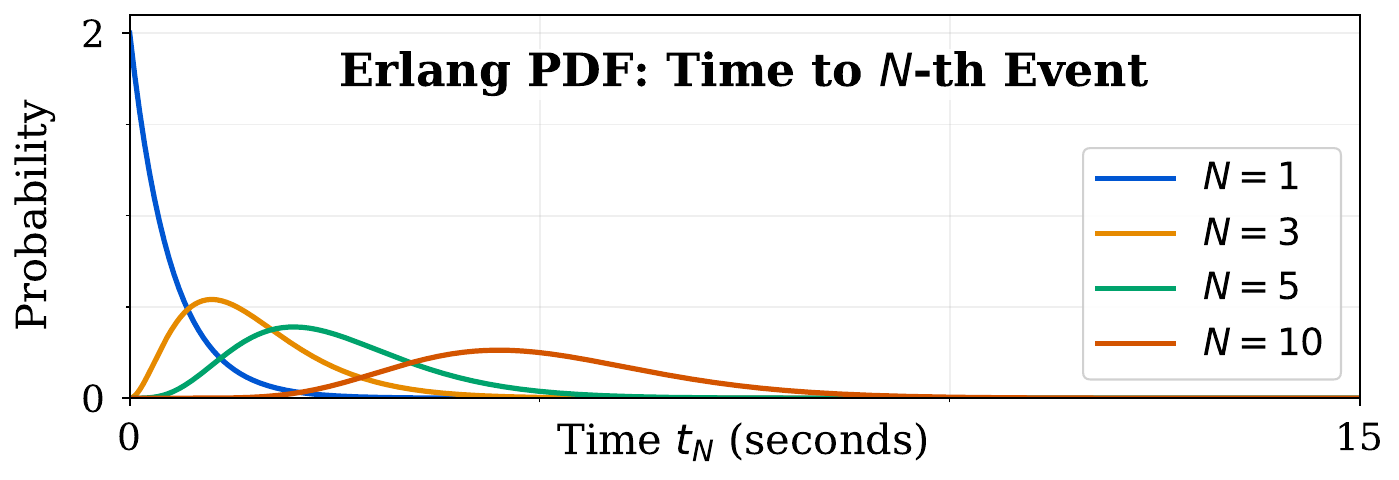} 
    % % Bottom Plot
    % \includegraphics[width=1.0\columnwidth]{figures/poisson_vs_erlang.pdf}
    \caption{
    % [Top] The Poisson PMF for varying expected counts $\lambda T$. It is  probability of observing exactly $N$ events.
    The Erlang PDF of the time of the $N$-th detection.}
    % [Bottom] The Poisson probability (shaded region) is the difference between two Erlang CDFs.}
    \label{fig:poisson_erlang_plots}
\end{figure}

To characterize a Poissonian process, an alternative to Eq.~(\ref{eq:SPAD_PAS_poisson_pmf_def}) is to consider the continuous PDF of time $t$ of the $N$-th event. This is expressed by the Erlang PDF (Fig.~\ref{fig:poisson_erlang_plots}),
\begin{equation}
    f_{\text{Erlang}}(t|N,\lambda) = 
    \frac{\lambda^N t^{N-1} \exp(-\lambda t)}{(N-1)!}, 
    \quad t \ge 0\;.
    \label{eq:SPAD_PAS_erlang_pdf}
\end{equation}
Its corresponding Cumulative Distribution Function (CDF) is
\begin{equation}
    F_{\text{Erlang}}(t|N,\lambda) = 1 - \sum_{i=0}^{N-1} 
     \frac{(\lambda t)^i \exp(-\lambda t)}{i!} \;.
    \label{eq:SPAD_PAS_erlang_cdf_explicit}
\end{equation}

The expressions (\ref{eq:SPAD_PAS_poisson_pmf_def},\ref{eq:SPAD_PAS_erlang_pdf},\ref{eq:SPAD_PAS_erlang_cdf_explicit})
relate to detection events, for $\tau_{\rm dead}=0$. However, due to dead-time, SPAD events are not Poissonian. For a generalized case where there is dead-time, we derive the discrete probability distribution $p(N|\lambda,{\cal S}_{\rm cont})$. The derivation is in the supplementary material, where it is shown to be non-negative and summed to 1. It is
 \begin{equation}
    \begin{split}
        p(N|\lambda,{\cal S}_{\rm cont}) %=
        %p_{\tt{Case I}}(N)+p_{\tt{Case II}}(N) 
        %\\%[10pt]
 %       &=\{F_{\text{Erlang}}[T-N\tau_{\rm{dead}};N]-F_{\text{Erlang}}[T-N\tau_{\rm{dead}};N+1]\} \\[10pt]
 %       &+\{F_{\text{Erlang}}[T-(N-1)\tau_{\rm{dead}};N]-F_{\text{Erlang}}[T-N\tau_{\rm{dead}};N]\}\\[10pt]
        %&
        &=F_{\text{Erlang}}\{[T-(N-1)\tau_{\rm{dead}}]|N,\lambda\} \\
        & -F_{\text{Erlang}}\{[T-N\tau_{\rm{dead}}]|N+1,\lambda \}\;.
    \end{split}
    \label{eq:SPAD_FM_final_erlang}
\end{equation}
%%%%%%%%%%%%%%%%

Based on %Eq.~
(\ref{eq:SPAD_FM_final_erlang}), we show in the Supplementary Material that 

\begin{equation}
%    \begin{split}
        \mathbb{E}\left[ N\right] 
%        &=\sum_{n=1}^\infty 
%           F_{\text{Erlang}}[T-(n-1)\tau_{\rm{dead}}|n,\lambda] \\[10pt] 
        %&
        =\sum_{n=1}^{N^{\rm max}} 
           F_{\text{Erlang}}[T-(n-1)\tau_{\rm{dead}}|n,\lambda] \;.
%    \end{split}
    \label{eq:CRLB_telescopic_cancel}
\end{equation}
Substituting Eq.~(\ref{eq:CRLB_telescopic_cancel}) in Eq.~(\ref{eq:CRLB_fisher_hybrid}), %{eq:CRLB_conT}),
\begin{equation}
    \begin{split}
         \text{Var}(\hat{\lambda}) \geq {\rm CRLB}%&= \frac{1}{I_{\tt temp}(\lambda)}
        =\frac{\lambda^2}
              {\sum_{N=1}^{N^{\rm max}} 
                 F_{\text{Erlang}}[T-(N-1)\tau_{\rm{dead}}|N,\lambda]
              }.
    \end{split}
    \label{eq:CRLM_temp_final}
\end{equation}

Let us study a high-flux limit. Being a CDF, %Eq.~
(\ref{eq:SPAD_PAS_erlang_cdf_explicit}) satisfies
\begin{equation}
    F_{\text{Erlang}}(t|N,\lambda)\xrightarrow
       {\lambda t\to \infty }1\;.
       \label{eq:CRLB_erlang_converge_to_1}
\end{equation}
Since $T>\tau_{\rm{dead}}$, then 
$T-(N-1)\tau_{\rm{dead}} \geq \tau_{\rm{dead}}$, recalling that 
$N\leq N^{\rm max}$. Consequently, for
$\lambda\tau_{\rm{dead}}\rightarrow\infty$, each element in the denominator 
of Eq.~(\ref{eq:CRLM_temp_final}) satisfies 
Eq.~(\ref{eq:CRLB_erlang_converge_to_1}). Overall, the denominator in Eq.~(\ref{eq:CRLM_temp_final}) tends to $N^{\rm max}$.
From Eqs.~(\ref{eq:N_max},\ref{eq:CRLM_temp_final},\ref{eq:CRLB_erlang_converge_to_1}),
\begin{equation}
        \text{STD}(\hat{\lambda}) 
       \geq \sqrt{\rm CRLB} 
       \xrightarrow
       {\lambda \tau_{\rm dead} \to \infty }
%       {\lambda T \to \infty, ~\lambda \tau_{\rm dead} \to \infty }
%        \sqrt{\frac{\lambda^2}{\sum_{n=1}^{N^{\rm{max}}} 1}} 
%         \frac{\lambda}\sqrt{\frac{\lambda^2}{N^{\rm{max}}}}
         %\\ &
        \frac{\lambda}
              {\sqrt{\lceil
                      \frac{T}
                           {\tau_{\rm{dead}}}
                     \rceil}}\;.
\label{eq:STD_high_temporal}
\end{equation}
Let us compare this result to Eq.~(\ref{eq:STD_highdiscreCt}).
Note that
\begin{equation}
    {\left \lceil
      \frac{T}{\tau_{\rm{dead}}}
    \right\rceil}
    \geq  \frac{T} {\tau_{\rm{dead}}}
    >  \frac{T} {\tau_{\rm{sense}}+\tau_{\rm{dead}}}
    ~~~~\forall \tau_{\rm{sense}}>0
    \;.
   \label{eq:comparefraqs}
\end{equation}
Hence, Eq.~(\ref{eq:STD_high_temporal}) is a lower bound for 
Eq.~(\ref{eq:STD_highdiscreCt}).

%%%%%%%%%%%%%%%%%%%%%%%%%%%%%%%%%%%%%%
\section{Maximum Likelihood Estimation}
\label{sec:CRLB_comparison_empirical_ML}

This section derives the ML estimators for {\bf M1, M2} and {\bf M3} operation modes. We use these estimators in the following sections. 
ML nulls the score functions, given respectively in
Eqs.~(\ref{eq:SPAD_FM_grad_log_likelihood},\ref{DISC_grad_log_likelihood},\ref{eq:HYBRID_binomial_grad_log_likelihood}).
%For {\bf M1}, ML nulls  Eq.~(\ref{eq:SPAD_FM_grad_log_likelihood}), 
This yields the corresponding estimators
%yielding an estimator 
%We define $\hat{\lambda}_{\rm (M1)}^{\rm ML}$ as the ML %estimation of Eq.~(\ref{eq:SPAD_FM_log_likelihood}). Using Eq.~(\ref{eq:SPAD_FM_grad_log_likelihood}),
\begin{equation}
     \hat{\lambda}_{\rm M1}^{\rm ML}=
        \begin{cases} 
        \displaystyle \frac{N}{T-N\tau_{\rm dead}}, & \texttt{Case 0,I}
        \\[10pt]
        \displaystyle \frac{N}{t_N-(N-1)\tau_{\rm dead}}, & \texttt{Case II}\;.
        \end{cases}
%    \end{split}
\label{eq:ML_M1}
\end{equation}
%For  {\bf M2}, ML nulls  
%Eq.~(\ref{DISC_grad_log_likelihood}), yielding %an estimator 
\begin{equation}
        \hat{\lambda}_{\rm M2}^{\rm ML}=\frac{1}{\tau_{\rm sense}}\log \left(\frac{B}{B-N}\right) \;.
%    \end{split}
    \label{eq:ML_M2}
\end{equation}
%For  {\bf M3}, ML nulls  
%Eq.~(\ref{eq:HYBRID_binomial_grad_log_likelihood}), yielding an estimator 
\begin{equation}
%    \begin{split}
%         &\frac{N}{\hat{\lambda}_{\rm M3}^{\rm ML}}-(B-N)\tau_{\rm{sense}}
%         - \sum_{i=1}^{N}{\tilde t}_i=0\\[5pt]
%        &\Rightarrow
   \hat{\lambda}_{\rm M3}^{\rm ML}=\frac{N}{\sum_{i=1}^N\tilde{t}_i+(B-N)\tau_{\rm sense}}\;.
%    \end{split}
    \label{eq:ML_M3}
\end{equation}
They are somewhat biased for a low event count: details on handling this are given in the {\em Supplementary Material}.  

%%%%%%%%%%%%%%%%%%%%%%%%%%%%%%%%%%%%%%
\section{Fundamental Bound Plots}
\label{sec:CRLB_comparison}

We summarize the results of Sec.~\ref{sec:score} in  Fig.~\ref{fig:CRLB}.
These are the fundamental performance bounds of unbiased estimators, that are based on SPAD measurements of Poissonian flux, in a single pixel, without regard to potential spatial statistical dependence.  The results, shown in  Fig.~\ref{fig:CRLB}, reflect the derivations in Sections~\ref{sec:CRLB_poisson}-\ref{sec:SPAD}. Specifically, these plots present the application of Eq.~(\ref{eq:invSNR}) on the functions expressed in
Eqs.~(\ref{eq:std_poisson},\ref{eq:CRLM_discrete_formula},\ref{eq:CRLM_discreteT},\ref{eq:CRLM_temp_final}), and are numerically computed. 

In an ideal world, a sensor measures all events with no dead-time. Accordingly, the Poissonian relative error sets the lower bound for any measurement system.
In practice, there is dead-time. Hence, the lowest bound is achieved by a measurement operation mode that senses continuously all events, without discrete time bins, and registers the event time. 
This is the most informative operation mode.

At a low event rate, i.e., a small $\lambda\tau_{\rm dead}$, all the bounds %of all methods 
coincide, while in each discrete bin $\tau_{\rm sense}\gg \tau_{\rm dead}$. Whether event times are %measured as 
continuous measurements inside discrete bins, or in a single, long continuous domain, 
the relative error plateaus at a high event rate,  i.e., a large $\lambda\tau_{\rm dead}$. 
The situation is markedly different if  no event time is registered as a continuous variable, and only the number of events in a discrete time grid is used. Then, the relative error increases exponentially, starting from a moderate rate $\lambda\tau_{\rm dead}\approx 1.6$. 

%%%%%%%%%%%%%%%%%%%%%%%%%%%%%%%%%%%%%%
\subsection*{Numerical Examples}
\label{sec:CRLB_comparison_empirical}

\begin{figure}[t] 
     \centering
     \includegraphics[width=\columnwidth]{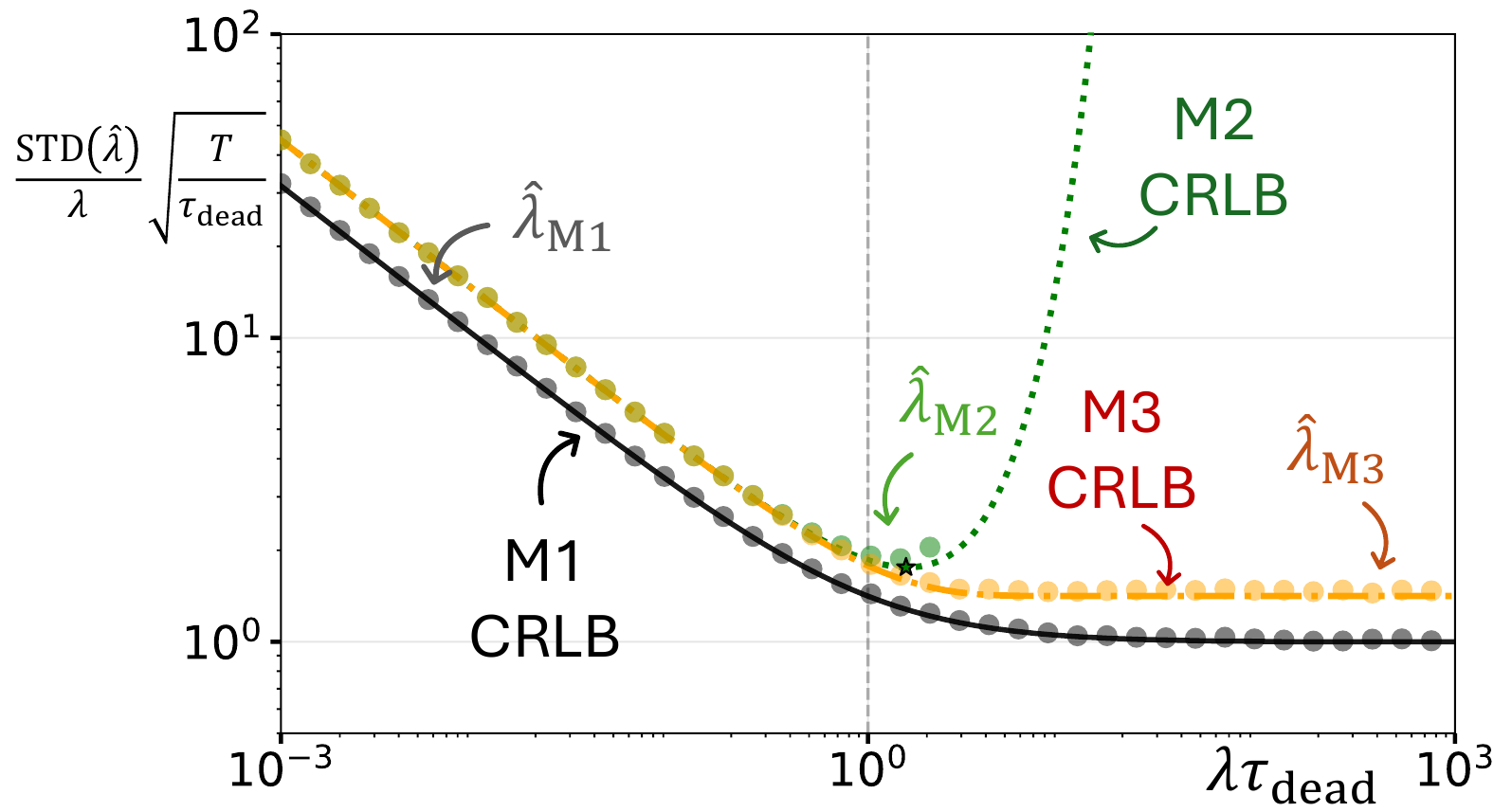}
     \caption{Numerical relative standard deviation, based on  ML estimations. The estimations are based on random data streams.
     The numerical results are overlayed on the bounds from  Fig.~\ref{fig:CRLB}. The plots compare operation modes of {\bf (M1)} {\em Free-running Timestamps}, {\bf (M3)} {\em Timestamped Bins} and {\bf (M2)} {\em Binary Bins}. 
     }
     \label{fig:CRLB_empirical}
\end{figure}

Figure~\ref{fig:CRLB_empirical} plots the relative standard deviation, calculated {\em numerically}. The plot is based on simulated event rate point-wise estimates. 
%The product $\lambda\tau_{\rm dead}\in[ 10^{-3},10^{3}]$ is sampled  in logarithmic steps. 
Each value of $\lambda\tau_{\rm dead}$  passes to several simulators of ${\cal F}$, depending on the operation mode of Sec.~\ref{sec:photon}. For {\bf (M1)} {\em Free-running Timestamps}, a sequence of detection events is simulated by an algorithm we implemented based on Ref.~\cite{suonsivu2021time}. This algorithm also 
incorporates a dark count rate (DCR), probability of after-pulsing (PAP) and timing jitter. We created simulations that implement the operation modes {\bf (M3)} {\em Timestamped Bins} and {\bf (M2)} {\em binary Bins}.
They account for $\tau_{\rm sense}$ in addition to $\tau_{\rm dead}$. The simulators use parameters labeled as {\em medium event rate} in the {\em Supplementary Material}.
Then, we use the corresponding\footnote{The {\em Supplementary Material} shows the consequence of estimators not corresponding the operation mode data. The study shows that matching the analysis operation mode to its corresponding data is beneficial for signal recovery.} ML estimators of Sec.~\ref{sec:CRLB_comparison_empirical_ML}. The simulated data are stochastic, and so are the estimates. For error statistics, each simulator ran $65536$ times for the lowest value $\lambda\tau_{\rm dead}$ value, and $4096$ times for the highest value $\lambda\tau_{\rm dead}$ value. 

In Fig.~\ref{fig:CRLB_empirical}, generally, the numerical relative standard deviation is above the value derived from the corresponding CRLB. 
% A few points lie slightly below the bound. We attribute this to the approximation of theoretical expectation by a finite mean. 
The plot corresponding to  {\bf M2}  terminates near $\lambda \tau_{\rm dead}=3$, for the following reason. A high event rate increases the probability of saturation, where all binary bins contain an event detection ($N=B$). When $N=B$, the logarithmic term in Eq.~(\ref{eq:ML_M2}) is singular and cannot be used. 

Therefore, we terminate the plot, there, where at least $1\%$ of the numerical tests saturate.  

%%%%%%%%%%%%%%%%%%%%%%%%%%%%%%%%%%%%%%%%%%
\section{Bound Under The Effect of a Prior}
\label{sec:BCRLB}

Sec.~\ref{sec:score} applies to estimation in individual pixels, without regard to spatial correlation. However, in natural images, there are spatial trends. These trends are partly known, being {\em prior information}. Exploiting a good prior improves recovery, i.e., it decreases the estimation variance. This is a practical effect, with a theoretical bound that generalizes Sec.~\ref{sec:score} to a  {\em Bayesian} CRLB (BCRLB)~\cite{TichavskyBCRLB, Scope_Crafts_2025}. Note: this prior is {\em unrelated to SPADs}. It is a prior on 2D images, projecting radiance of scene objects. The radiance of objects is oblivious to the timing mechanisms of a SPAD sensor. 

First, point measurements generalize to an array. Represent the two dimensional (2D) image ${\bm\lambda}$ as a column-stack vector. Its estimation is ${\hat {\bm\lambda}}$. Define the likelihood score as a column vector

\begin{equation}
\begin{split}
       &\nabla_{{\bm \lambda}} 
    \log p\!\left(\mathcal{\bm D}|\bm{\lambda},{\cal S}\right)
    \\&~~~=\left[\frac{\partial \log p\!\left(\mathcal{\bm D}|{\lambda}_1,{\cal S}\right)}{\partial {\lambda}_1},\ldots,\frac{\partial \log p\!\left(\mathcal{\bm D}|{\lambda}_r,{\cal S}
                 \right)}
              {\partial {\lambda}_r},         
    \ldots \right]^{\top},
\end{split}
    \label{eq:gradLambda_split}
\end{equation}
where ${\top}$ denotes transposition. 
Similarly, define the score function of the prior term as a column vector $\nabla_{{\bm \lambda}} \log p\!(\bm{\lambda})$.
Eq.~(\ref{eq:fishebasic}) generalizes to a Fisher information {\em matrix}, based solely on sensor data 
\begin{equation}
    {\bf I}_{\rm D} = 
      \mathbb{E} 
       \left\{
        [\nabla_{{\bm\lambda}} 
                \log p({\cal D}|{\bm\lambda},{\cal S})]~
              [\nabla_{{\bm\lambda}} 
                \log p({\cal D}|{\bm\lambda},{\cal S})]^{\top}
       \right\} \;,
    \label{eq:fisheDmat}
\end{equation}
where $\mathbb{E}$ denotes expectation over random data {\em and}
over ${\bm\lambda}$. 

Photon flux is Poissonian, and dead-time onset in any pixel is independent of other pixels. So, for any pixels $r,r'$, data are statistically independent.
Since data are independent, it can be shown that for our SPAD data,  ${\bf I}_{\rm D}$ is {\em diagonal}. Each element on the diagonal is Eq.~(\ref{eq:fishebasic}), when $\lambda$ is fixed.

Define a prior information matrix 
\begin{equation}
    {\bf I}_{\rm P} = 
      \mathbb{E} 
       \left\{
        [\nabla_{{\bm\lambda}} 
                \log p({\bm\lambda})]~
        [\nabla_{{\bm\lambda}} 
                \log p({\bm\lambda})]^{\top}
       \right\} \;,
    \label{eq:fishePmat}
\end{equation}
with expectation over ${\bm\lambda}$. Then, using expectation over
the data and ${\bm\lambda}$, the BCRLB is
\begin{equation}
    %\left[
      \mathbb{E} 
       \left\{
       {\rm Cov}({\hat {\bm\lambda}})
        \right\}
    %\right]^{-1}
    \succeq 
    [{\bf I}_{\rm D} + {\bf I}_{\rm P}]^{-1} 
    \;,
    \label{eq:Cov}
\end{equation}
where $\mathbb{E} 
       \left\{
       {\rm Cov}({\hat {\bm\lambda}})
        \right\}$ is the Mean Square Error (MSE) covariance matrix.
The diagonal yields the bound on the expected recovery MSE, per pixel:
\begin{equation}
    {\rm BCRLB}_{\rm MSE}={\rm diag}\left([{\bf I}_{\rm D} + {\bf I}_{\rm P}]^{-1}\right)\;.
    \label{eq:BCRLB_var}
\end{equation}
Consider the case where the prior term is ineffective. This is equivalent to stating that all possible event-rate maps ${\bm\lambda}$ are equally probable {\em a-priori}. Then,  it can be shown that
${\bf I}_{\rm P}=0$.  Then, since ${\bf I}_{\rm D}$ is diagonal, Eq.~(\ref{eq:BCRLB_var}) degenerates to the  CRLB~(\ref{eq:CRLB_fisher}) for a fixed 
${\bm\lambda}$.
 %, when $\lambda$ is fixed.
Generally, prior information is not zero, and ${\bf I}_{\rm P}\succeq 0$. So, the MSE covariance in Eq.~(\ref{eq:Cov}) shrinks. The error bound is lower than the CRLB. This motivates the use of strong priors, as used in SBD. 

%%%%%%%%%%%%%%%%%%%%%%%%%%%%%%%%%%%%%%%%%%
\subsection*{Computing a Practical Bound}
\label{sec:BCRLB_calc}

In this section we derive practical computation of the BCRLB, based on a training dataset ${\cal T}$. Eq.~(\ref{eq:fisheDmat}) is estimated using the empirical mean over  ${\cal T}$:
\begin{equation}
%\begin{split}
        {\bf I}_{\rm D} %&
        \!\approx\!\frac{1}{|{\cal T}|}\sum_{{\bm \lambda}\in {\cal T}}\big[\nabla_\lambda\log  p({\cal D}|{\bm \lambda},{\cal S})\big]\big[\nabla_\lambda\log  p({\cal D}|{\bm \lambda},{\cal S})\big]^{\top}.    
%    \end{split}
    \label{eq:Id_approx}
\end{equation}
For ${\bf I}_{\rm P}$ we use the following process.  
Let ${\bm x}$ be a column vector representing a 2D map.  Recall from Sec.~\ref{sec:LDbackground}, that a score function of the prior  
$\nabla_{\bm x}\log p(\bm x)$ is approximated by a learned vector function
$\bm{s}_{\bm{\theta}}({\bm x},k)$. This function trains using 
${\cal T}$ for SBD. We use a method described in Ref.~\cite{Scope_Crafts_2025}. The last step ($k=1$) in SBD uses in Eq.~(\ref{eq:BG_DPS_J}) a function $\bm{s}_{\bm{\theta}}({\bm x},1)$, which is applicable to clean images, hence can be used on ground-truth images. 

SBD models commonly operate in a $[-1,1]$ graylevel range per element (pixel) of ${\bm x}$. In contrast, $\lambda$ is proportional to a non-negative flux $[{\rm photons/s}]$ that is not limited to a particular supremum.  We address the discrepancy by domain adaptation:
\begin{equation}
    \bm{x} = (2/\zeta) {\bm\lambda} -1
    \;,   
    \label{eq:DAPS_affine_inv}
\end{equation}
where $\zeta>0$ is the supremum over the train data. We later explain how we set it in simulations and real data. Applying Eq.~(\ref{eq:DAPS_affine_inv}) to any 
${\bm\lambda}\in{\cal T}$ yields a corresponding ground-truth image denoted ${\bm x}\in{\cal T}$ in the graylevel range $[-1,1]$. Then, 
\begin{equation}
    \nabla_{{\bm \lambda}}\log p({\bm \lambda})=
    \nabla_{{\bm \lambda}}\log 
       \left[p({\bm x})\cdot
          \left|
             \det\frac{\partial {\bm x}}{\partial {\bm \lambda}}
         \right| 
       \right] \;.
\label{eq:conversionP}
\end{equation}
From Eq.~(\ref{eq:DAPS_affine_inv}), 
$\partial {\bm x}/\partial {\bm \lambda}=(2/\zeta){\bm{I}}$, independent of 
${\bm \lambda}$. Therefore, Eq.~(\ref{eq:conversionP}) leads to 

\begin{eqnarray}
    &\nabla_{{\bm \lambda}}\log p({\bm \lambda})=
    ~~~~~~~~~~~~~~~~~~~~~~~~~~~~~~~~~~~~~~~~~~~~~~~~~~~~~~~~~~~~~~
    \\  \nonumber
    &\left(\frac{\partial {\bm x}}{\partial {\bm \lambda}}\right)^{\top}\nabla_{{\bm x}}\log p({\bm x})+\nabla_{{\bm \lambda}}\log \left|\det\frac{\partial {\bm x}}{\partial {\bm \lambda}}\right| 
    % &=\left(\frac{\partial {\bm x}}{\partial {\bm \lambda}}\right)^{\top}\nabla_{{\bm x}}\log p({\bm x})+\nabla_{{\bm \lambda}}\log \left|\det\frac{\partial {\bm x}}{\partial {\bm \lambda}}\right|\;.
    =\frac{2}{\zeta}\bm{s}_{\bm{\theta}}({\bm x},1)\;.
\label{eq:conversion_jacobian}
\end{eqnarray}

Similarly to Eq.~(\ref{eq:Id_approx}), we estimate ${\bf I}_{\rm P}$ of Eq.~(\ref{eq:fishePmat}) using the mean over ${\cal T}$: 

\begin{equation}
    \begin{split}
        {\bf I}_{\rm P}&\approx\frac{1}{|{\cal T}|}\sum_{{\bm \lambda}\in {\cal T}}\left[\nabla_{{\bm \lambda}} 
                \log p({\bm \lambda})\right]
        \left[\nabla_{{\bm \lambda}} 
                \log p({\bm \lambda})\right]^{\top}\\[5pt]
        &\approx\frac{4}{\zeta^2|{\cal T}|}\sum_{{\bm x}\in {\cal T}}
        [\bm{s}_{\bm{\theta}}({\bm x},1)]
        [\bm{s}_{\bm{\theta}}({\bm x},1)]^{\top}
        \;. 
    \end{split}
    \label{eq:Ip_approx}
\end{equation}
From Eqs.~(\ref{eq:Id_approx},\ref{eq:Ip_approx}), we compute Eqs.~(\ref{eq:Cov},\ref{eq:BCRLB_var}).

%%%%%%%%%%%%%%%%%%%%%%%%%%%%%%%%%%%%%%%%%%
\section{Diffusion based on SPAD data}
\label{sec:domain_adaptation}

As we explain in Sec.~\ref{sec:LDbackground}, derivation of the likelihood function is a key for solving inverse problems. Specifically, the likelihood  score function is required for SBD. Ref.~\cite{melidonis2025score} explores SBD-based reconstruction of simulated SPAD data, assuming a mode similar to Sec.~\ref{sec:discrete_model}, but without consideration of $\tau_{\rm sense}$ and $\tau_{\rm dead}$. We account for these time constants and also handle the operation modes of Secs.~\ref{sec:hybrid_model} and~\ref{sec:SPAD}. 
%To our knowledge, these operational modes were not demonstrated with diffusion models. 
We now employ SBD using score functions we derive in Sec.~\ref{sec:score}. 

% A discrepancy exists between  ranges of SPAD data and  common SBD approaches, specifically DPS. 
Recalling from Sec.~\ref{sec:LDbackground}, in SBD, a score function of the prior term $\nabla_{\bm{x}} \log p_{{\bm{x}}}({\bm{x}}_k)$ is approximated by a {\em score DNN} $s_{\bm{\theta}}$. The DPS estimate $\hat{\bm{x}}_0$ and $s_{\bm{\theta}}$ use the $[-1,1]$ graylevel range. So, 
the inverse of Eq.~(\ref{eq:DAPS_affine_inv}) converts  $\hat{\bm{x}}_0$ to a non-negative estimate of the image
%Eq.~(\ref{eq:DAPS_affine}) for ${\hat {\bm\lambda}}$.
\begin{equation}
    {\hat {\bm\lambda}}=\zeta(\hat{\bm{x}}_0+1)/2 \;.
    \label{eq:DAPS_affine}
\end{equation}

In the domain of $\lambda$, we use, {\em per pixel} indexed $r$, the likelihood score. This score is either one of Eqs.~(\ref{DISC_grad_log_likelihood},\ref{eq:HYBRID_binomial_grad_log_likelihood},\ref{eq:SPAD_FM_grad_log_likelihood}), depending on the SPAD operation mode. We use Eq.~(\ref{eq:DAPS_affine}) to compute
$\frac{\zeta}{2}\nabla_{{\bm \lambda}} \log p\!\left(\mathcal{\bm D}|\bm{\lambda},{\cal S}\right)$. This gradient is multiplied by the coefficient $\rho$, and the Jacobian ${\bm J}$ of Eqs.~(\ref{eq:BG_DPS_J},\ref{eq:BG_DPS_update}).
Overall, the SBD method for reconstruction of scenes based on SPAD data is listed in Algorithm~\ref{alg:recon_alg}.
The algorithm uses the sets $\{\alpha_k, \sigma_k\}_{k=1}^K$.
We used without adaptation these sets, as quoted in the DPS paper~\cite{chung2023diffusion}.
\begin{algorithm}[t]
\caption{SPAD Signals Reconstruction Algorithm}
\begin{algorithmic}[1]
\Require Measurements $\mathcal{\bm D}$, trained score DNN $\bm{s}_{\bm{\theta}}$ of the prior, forward model parameters ${\cal S}$, schedules $\{\alpha_k, \sigma_k\}_{k=1}^K$, $\rho$.
\State Initialize $\bm{x}_K \sim \mathcal{N}(\bm{0}, \bm{I})$ 
\Ensure Reconstructed Event Rate 2D map $\hat{\bm{\lambda}}$.
\For{$k = K \textbf{ to } 1$}
    \State \textbf{1. Estimation of the score of the prior term:}
    \State $\hat{\bm{s}} \leftarrow \bm{s}_{\bm{\theta}}(\bm{x}_k, k)$
    
    \State \textbf{2. Tweedie's estimation of a clean image:}
    \State $\bar{\alpha}_k\leftarrow\Pi_{i=1}^k\alpha_i$
    \State $\hat{\bm{x}}_0 \leftarrow 
    \frac{1}{\sqrt{\bar{\alpha}_k}} 
    \left[ \bm{x}_k + (1 - \bar{\alpha}_k)\hat{\bm{s}} \right]$
        \State $\bm{J}\leftarrow \frac{1}{\sqrt{\bar{\alpha}_k}}\left[ {\bm{I}} + {(1-\bar{\alpha}_k)}\nabla_{\bm{x}_k}\hat{\bm{s}} \right]$ 
    
    \State \textbf{3. Reverse diffusion:}
    \State $\bm{\eta} \sim \mathcal{N}(\bm{0}, \bm{I})$
    \State $\bm{x}'_{k-1} \leftarrow \frac{\sqrt{{\alpha}_k}(1-\bar{\alpha}_{k-1})}{1-\bar{\alpha}_k}\bm{x}_k + \frac{\sqrt{\bar{\alpha}_{k-1}}(1-\alpha_k)}{1-\bar{\alpha}_k}\hat{\bm{x}}_0 + \sigma_k \bm{\eta}$

    \State \textbf{4. Domain adaptation:}
    \State $\hat{\bm{\lambda}} \leftarrow \frac{\zeta}{2}(\hat{\bm{x}}_0+1)$
    
    \State \textbf{5. Likelihood gradient per pixel:}
    \State $\frac{\partial}{\partial {\lambda}} \log p\left(\mathcal{\bm D}| {\lambda},\mathcal{S}\right)\leftarrow \text{One of Eqs.~}(\ref{DISC_grad_log_likelihood},\ref{eq:HYBRID_binomial_grad_log_likelihood},\ref{eq:SPAD_FM_grad_log_likelihood})$
    \State \textbf{6. Domain Re-adaptation:}
    \State $\bm{x}_{k-1} \leftarrow \bm{x}'_{k-1} + 
    \rho(\zeta/2)\bm{J}\nabla_{\bm{\lambda}} \log p\left(\mathcal{\bm D}| {\bm {\lambda}},\mathcal{S}\right)$
\EndFor
\State $\hat{\bm{\lambda}}\leftarrow \frac{\zeta}{2}({\bm{x}}_0+1)$
\State \Return $\bm{\lambda}$
\end{algorithmic}
\label{alg:recon_alg}
\end{algorithm}

Any good, differentiable image  prior score 
$\bm{s}_{\bm{\theta}}$ can be used. However, most existing  DNN-based priors have trained on color images as inputs and outputs. Our scenes are taken by grayscale sensors. So, we trained $\bm{s}_{\bm{\theta}}$  from scratch on grayscale images described below. We opted to use a U-Net~\cite{ronneberger2015unet} architecture as a basis for the score DNN. We changed this architecture, to work with grayscale images. We optimized ${\bm{\theta}}$ using the code of Ref.~\cite{dhariwal2021diffusion}.

%%%%%%%%%%%%%%%%%%%%%%%%%%%%%%%%%%%%%%%%%%%%%%%
\section{Simulations}
\label{sec:Experiments}

%The experiments above are limited, due to a current shortage of real SPAD data in the different operation modes. So, 
We describe now simulated tests. Further simulation details and results appear in the {\em Supplementary Material}. {\bfseries Source code is available at:\\
\url{https://doi.org/10.5281/zenodo.20858183}}. %In this section, we describe main aspects. 

Training and testing are based on Flickr-Faces-HQ  (FFHQ)~\cite{karras2019style} data of color face images, converted to gray.

Generating event data has several steps. First, a method from~\cite{suonsivu2021time} simulates
the flux $\Phi$. 

Then we apply $\lambda = q\Phi$ at the pixel.
%Eq.~(\ref{eq:SPAD_FM_eff_flux}) converts $\Phi$ to  $\lambda$ at the pixel. 
The expected event rate {\em per pixel} is then used in one of several simulators mentioned in Sec.~\ref{sec:CRLB_comparison_empirical}.

We detail in the {\em Supplementary Material} how to set $\zeta$ and tune the parameter $\rho$, mentioned in Sec.~\ref{sec:LDbackground} and in Alg.~\ref{alg:recon_alg}.
A test set ${\cal T}$ contains 900 scenes, that had been used for neither training nor tuning. 
On each test scene, %rate scenario and sensor operation mode, 
we applied the respective simulators. 
%described in Sec.~\ref{sec:simulation}. 
Then, we applied reconstruction by SBD, as described in Sec.~\ref{sec:domain_adaptation}.  
% Setting  $\zeta$ is done by substituting $I=1$ in Eq.~(\ref{eq:SIM_lambda}), then the using the value yielded by Eq.~(\ref{eq:SPAD_FM_eff_flux}).
For each operation mode, we used the corresponding likelihood 
score functions (\ref{DISC_grad_log_likelihood},\ref{eq:HYBRID_binomial_grad_log_likelihood},\ref{eq:SPAD_FM_grad_log_likelihood}). 
In addition, we apply ML from Ref.~\cite{ingle2021passive}, relating to the mode of {\bf (M1)} {\em Free-running Timestamps}. 

\begin{figure}[t!] 
    \centering
    \includegraphics[width=\columnwidth]{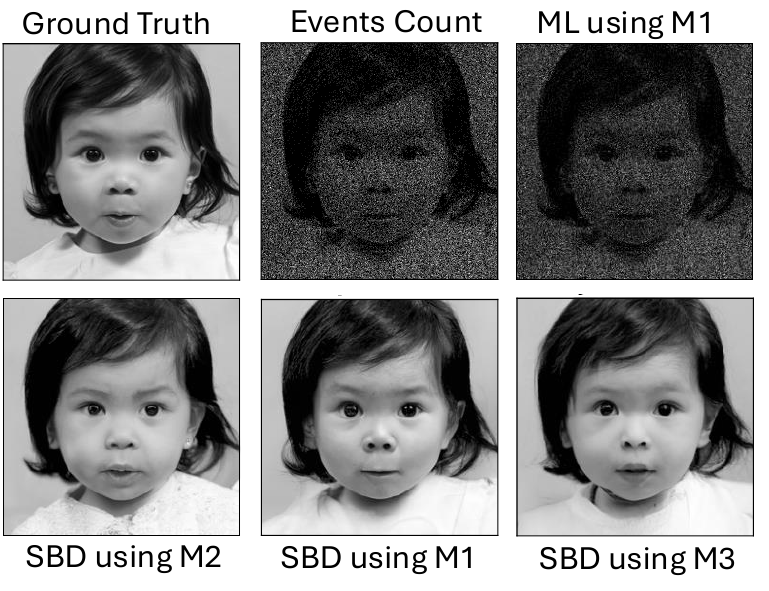}
    \caption{Simulation results for a medium event rate. 
    From left to right: Ground truth image; Raw events count; ML based on {\bf (M1)} {\em Free-running Timestamps}; SBD using {\bf (M2)} {\em binary Bins}. SBD using {\bf (M1)} {\em Free-running Timestamps}. SBD using {\bf (M3)} {\em Timestamped Bins}.}
    \label{fig:all_results}
\end{figure}

Examples are shown in Fig.~\ref{fig:all_results}. Statistics across the test set for a medium event rate are listed in Table~\ref{tab:sim_results}. %More results appear in the {\em Supplementary Material}. 

Generally, the mode {\bf (M1)} {\em Free-running Timestamps} yields the best results, both visually and statistically, followed by {\bf (M3)} {\em Timestamped Bins}.
Statistics as in Table~\ref{tab:sim_results} are affected by the event rate of each scene sample, the stochastic nature of the detection events and random sampling of reverse diffusion by DPS.

Sec.~\ref{sec:BCRLB_calc} leads to a bound in Eq.~(\ref{eq:Cov}),
based on  Eqs.~(\ref{eq:Id_approx},\ref{eq:Ip_approx}). The bound requires inversion of the matrix  $[{\bf I}_{\rm D} + {\bf I}_{\rm P}]$.
For an image having $N^{\rm pixels}$, the matrix had dimensions
$N^{\rm pixels}\times N^{\rm pixels}$. For the FFHQ images used above, $N^{\rm pixels}=256^2$, challenging matrix inversion. So, we demonstrate the bound for FFHQ images resized to $32\times 32$ pixels. We trained ${\bm s}_{\bm{\theta}}$ for the resized data. For this test, we simulate {\bf (M1)} {\em Free-running Timestamps} data  high event rate (See Supplementary Material).
Then, recovery uses the {\bf (M1)} operation mode.

\begin{figure}[t!] 
    \centering
    \includegraphics[width=\columnwidth]{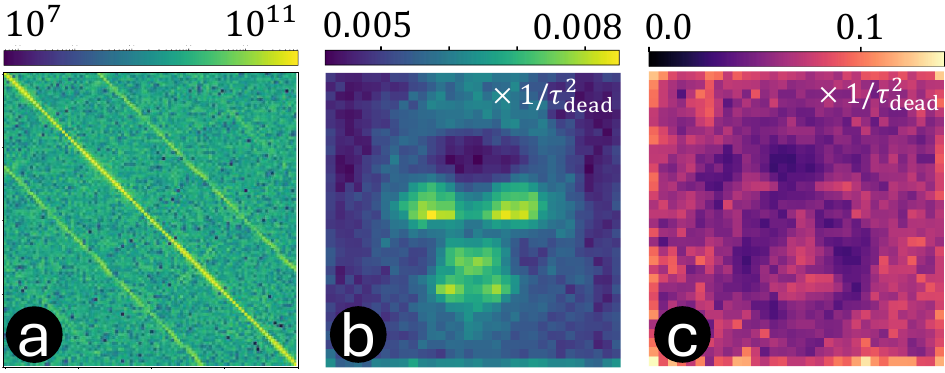}
    \caption{Empirical BCRLB for $32\times32$ scene recovery. 
    (a) The matrix $[{\bf I}_{\rm D} + {\bf I}_{\rm P}]^{-1}$.
    %    right-hand side of Eq.~(\ref{eq:Cov}) 
    This is a zoom-in on its $100\times100$ center patch. We display the absolute values in logarithmic scale. (b) Bound on the expected recovery MSE; (c) $\Delta_{\rm error}$ from Eq.~(\ref{eq:empirical_diff}). 
    The scale in (b,c) is $1/\tau_{\rm dead}^2$.}
    \label{fig:BCRLB_all}
\end{figure}

The inverted matrix  $[{\bf I}_{\rm D} + {\bf I}_{\rm P}]^{-1}$ from the right-hand side of Eq.~(\ref{eq:Cov}) is shown in Fig.~\ref{fig:BCRLB_all}(a), zoomed-in and in log-scale of absolute values. As expected, its main diagonal dominates, and its values decay with distances from each image pixel. 
% The diagonal yields the bound on the expected recovery variance, per pixel:
% \begin{equation}
%     {\rm BCRLB}_{\rm var}={\rm diag}\left([{\bf I}_{\rm D} + {\bf I}_{\rm P}]^{-1}\right)\;.
%     \label{eq:BCRLB_var}
% \end{equation}
Eq.~(\ref{eq:BCRLB_var}) is depicted in Fig.~\ref{fig:BCRLB_all}(b). The difference between the empirical test-set MSE and Eq.~(\ref{eq:BCRLB_var}) is
\begin{equation}
    \Delta_{\rm error}=\left[\frac{1}{|{\cal T}|}\sum_{{\bm \lambda}\in{\cal T}}({\bm \lambda}-\hat{{\bm \lambda}})^2\right]- {\rm BCRLB}_{\rm MSE}\;.
    \label{eq:empirical_diff}
\end{equation}
The map of $\Delta_{\rm error}$ is depicted in Fig.~\ref{fig:BCRLB_all}(c). This map is positive for all pixels. Therefore, Eq.~(\ref{eq:BCRLB_var}) indeed bounds the empirical MSE.

\begin{table}[t]
\centering
\caption{Performance measures.  Continuous Times stand for {\bf (M1)} {\em Free-running Timestamps}. Discrete Bins stand for {\bf (M2)} {\em binary Bins}. Times in Bins stand for {\bf (M3)} {\em Timestamped Bins}.}
\label{tab:sim_results}
% Reduced from 5.0pt to 4.2pt to pull the right edge away from the column gutter
\setlength{\tabcolsep}{4.2pt}
\small 
\begin{tabular}{@{} l 
    S[table-format=2(1), table-number-alignment=center] 
    S[table-format=1.2(1), table-number-alignment=center] 
    S[table-format=1.2(1), table-number-alignment=center] 
    S[table-format=3, table-number-alignment=center] @{}}
\toprule
Method & {PSNR\makebox[0pt][l]{$\uparrow$}} & {SSIM\makebox[0pt][l]{$\uparrow$}} & {LPIPS\makebox[0pt][l]{$\downarrow$}} & {FID\makebox[0pt][l]{$\downarrow$}} \\ 
% \midrule
% ML: Continuous Times & 6(2)  & 0.02(4)  & 0.75(4)  & 467 \\
% SBD: Discrete Bins      & 15(2) & 0.45(9)  & 0.57(7)  & \bfseries 75  \\
% SBD: Times in Bins      & 17(2) & 0.56(8)  & 0.51(8)  & 78  \\ 
% SBD: Continuous Times   & \bfseries 18(2) & \bfseries 0.61(8) & \bfseries 0.49(7) & \bfseries 75 \\

\specialrule{1.2pt}{2pt}{2pt} 
ML: Continuous Times & 8(2) & 0.11(4) & 0.76(5) & 305 \\
SBD: Discrete Bins      & 22(2) & 0.61(8) & 0.36(5) & 36 \\
SBD: Times in Bins      & 22(2) & 0.61(8) & 0.36(5) & 36  \\ 
SBD: Continuous Times   & \bfseries 23(2) & \bfseries 0.67(7) & \bfseries 0.33(5) & \bfseries 33 \\

% \specialrule{1.2pt}{2pt}{2pt} 
% ML: Continuous Times & 11(1) & 0.08(2) & 0.89(4) & 407 \\
% SBD: Discrete Bins      & 7(2)  & 0.21(5) & 0.80(4) & 373 \\
% SBD: Times in Bins      & 15(3) & 0.44(8) & 0.60(6) & 113 \\ 
% SBD: Continuous Times   & \bfseries 17(2) & \bfseries 0.59(8) & \bfseries 0.49(7) & \bfseries 71 \\
\bottomrule
\end{tabular}
\end{table}

%%%%%%%%%%%%%%%%%%%%%%%%%%%%%%%%%%%%%%%
\section{Demonstrations on real SPAD data}
\label{sec:spad_measurements}

%%%%%%%%%%%%%%%%%%%%%%%%%%%%%%%%%%%%%%%
%\subsection{Fan Scene}
%\label{sec:fan_data}

In this section, we show SBD-based recovery from real {\bf M1} and {\bf M3} SPAD data. The data are of a {\em Fan} and a {\em Tunnel} scene.

{\bf Fan scene:} Data of a 3-blade fan are from  Ref.~\cite{seets2020motionadaptivedeblurringsinglephoton, seets2024dataset}. Data are in the mode of Sec.~\ref{sec:hybrid_model}: 
{\bf (M3)} {\em Timestamped Bins}. The sensor has $32\times 32$ SPAD pixels, $\tau_{\rm sense}=2[\mu s]$, $\tau_{\rm dead}=18[\mu s]$.
We use the ML estimator from Eq.~(\ref{eq:ML_M3}). It yields $\hat{\lambda}^{\rm ML}_{\rm M3}$ per pixel. We use the maximum value across the image to set $\zeta=\max\{{\hat{\lambda}^{\rm ML}_{\rm M3}}\}$ for Eq.~(\ref{eq:DAPS_affine}).

\begin{figure}[t] 
    \centering
    \includegraphics[width=0.7\columnwidth]{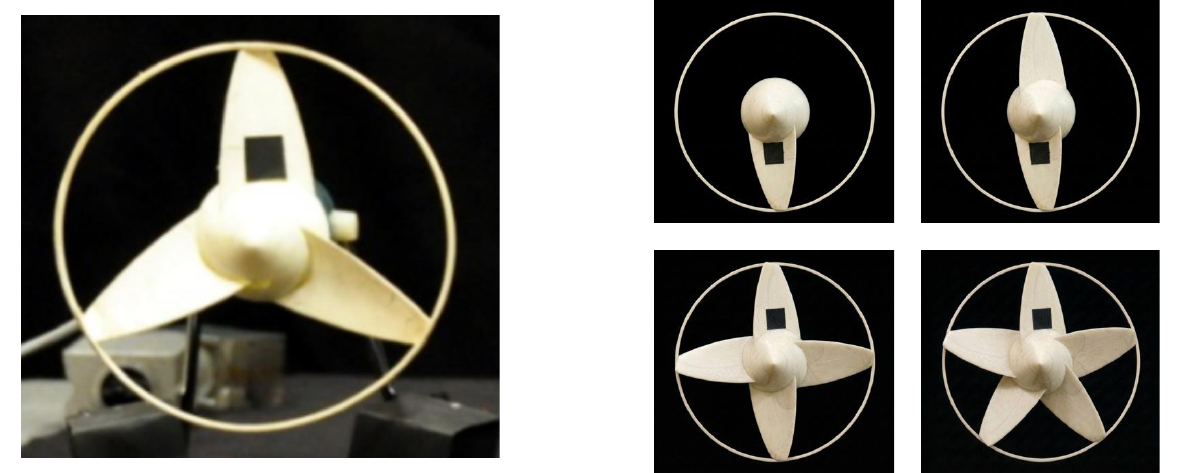}
    \caption{[Left]: Image of the fan, from~\cite{seets2020motionadaptivedeblurringsinglephoton}. 
    [Right]: Fan variants, synthesized by {\em Gemini}, a language model by Google.}
    \label{fig:fan_data}
\end{figure}

\begin{figure}[t] 
    \centering
    \includegraphics[width=\columnwidth]{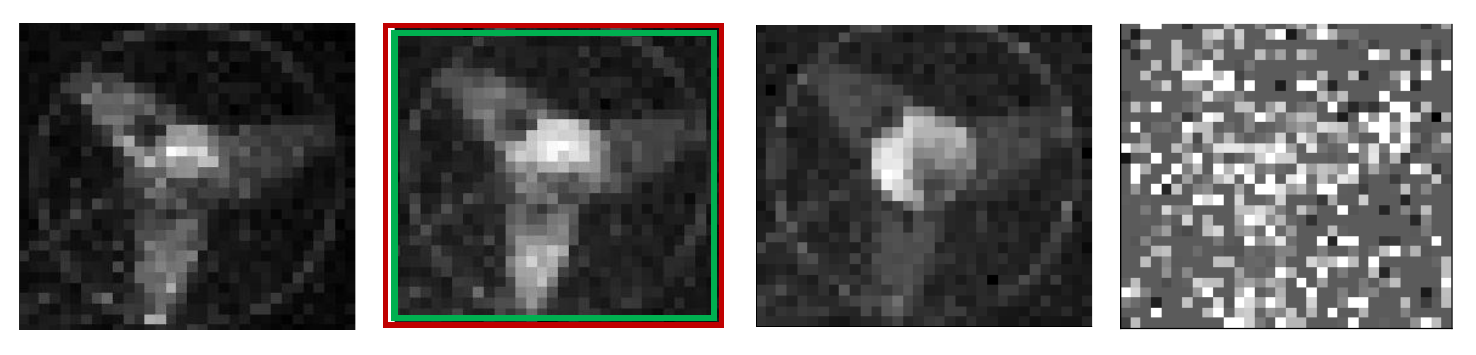}
    \caption{Reconstruction from real SPAD data, by plugging different likelihood functions in the recovery computations. Except for ML, all results are by a diffusion model, each using a different data likelihood score. 
    [Left to Right]: ML from Eq.~(\ref{eq:ML_M3}); {\bf (M3)} {\em Timestamped Bins}, which is the {\bf true model} of this data source;
    {\bf (M2)} {\em binary Bins}; 
    and {\bf (M1)} {\em Free-running Timestamps}. The latter fails as it ignores 
    ${\tau_{\rm sense}}$.}
    \label{fig:fan_recon}
\end{figure}

%First, we performed training. 
We did not find sufficient training sets of fans.  
We thus created synthetic data of $10^{4}$ fan variants using the language model 
{\em Gemini} by Google. They are similar to those depicted in Fig.~\ref{fig:fan_data}, 
with  different numbers of blades and rotation angles, yet {\em no} sample has three blades, to challenge the test. Samples are scaled to $32\times 32$ pixels and converted to gray.
%We then downsized the synthetic images down to $32\times 32$ pixels and converted them to grayscale. 
We trained the U-Net for $10^5$ steps, over three hours on NVIDIA Tesla V100-DGXS GPU with 32GB HBM2 memory. 
We manually set $\rho=10^{-3}$.
%for Algorithm~\ref{alg:recon_alg}. 
Fig.~\ref{fig:fan_recon} shows reconstructions corresponding to  
Eqs.~(\ref{DISC_grad_log_likelihood},\ref{eq:HYBRID_binomial_grad_log_likelihood},\ref{eq:SPAD_FM_grad_log_likelihood}), with $T=600$~[$\mu$s]. Additional results with $T=300$~[$\mu$s] are in the supplementary material.
The result based on the score function of the operation mode this data source is marked in a green frame. In contrast, failure is seen when using the score function of {\bf (M1)} {\em Free-running Timestamps}, because it ignores $\tau_{\rm sense}$, and wrongly interprets single detections 
over bins, as if $\lambda$ is very small.\\

\begin{figure}[t] 
    \centering
    \includegraphics[width=\columnwidth]{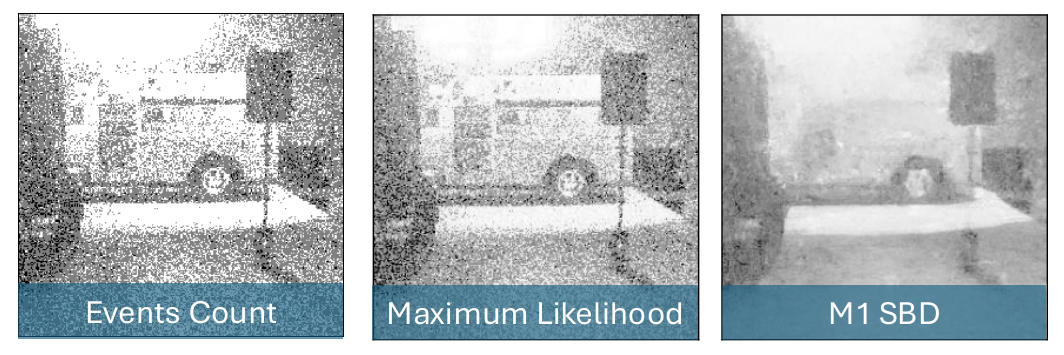}
    \caption{Reconstruction of tunnel data. For display purposes only, Gamma-correction of $\gamma=1/2$ was applied. From left to right: Events count; ML Reconstruction.~\cite{ingle2021passive}; {\bf (M1)} {\em Free-running Timestamps}-based SBD reconstruction.}
    \label{fig:truck_data}
\end{figure}

%%%%%%%%%%%%%%%%%%%%%%%%%%%%%%%%%%%%%%%
%\subsection{Tunnel Scene}
%\label{sec:tunnel_data}
\vspace{-0.35cm}
\noindent {\bf Tunnel scene:} Data are from Ref.~\cite{ingle2021passive} and are in the mode of Sec.~\ref{sec:SPAD}: 
{\bf (M1)} {\em Free-running Timestamps}.  We used $T\approx250$~[ns], $\tau_{\rm dead}\approx110$~[ns] and $400\times 400$ pixels. 
%Specifications per scene pixel are provided with the data. Records show slight variation in parameters, but generally: 
%. The tunnel scene include signs, dolls and a truck. 
It has a high dynamic range. We set $\zeta$ using ML suited for {\bf (M1)} {\em Free-running Timestamps}, along Ref.~\cite{ingle2021passive}. 
We trained the U-Net on 2 sub-classes of ImageNet~\cite{deng2009imagenet} (Fire trucks, Ambulances) for $4\cdot 10^4$ steps, over 6 days, using 4 NVIDIA A100-SXM4-80GB GPUs. We cropped the data to $256\times256$, to comply with the architecture of Ref.~\cite{dhariwal2021diffusion}. Each SPAD pixel detected at most 2 events. We set $\rho=5\cdot10^{-3}$. Reconstructions appear in Fig.~\ref{fig:truck_data}. With so few detected events, ML reconstruction is very noisy. The SBD reconstruction using {\bf (M1)} {\em Free-running Timestamps} likelihood is smoother as result of its prior.  

%%%%%%%%%%%%%%%%%%%%%%%%%%%%%%%%%%%%%%%%%%
\section{Discussion}
\label{sec:conclusions}

Operating modes typically reside on different sensors rather than coexisting on a single device. This motivates performance prediction.
There is a cost/performance tradeoff. For a given application with constraints on power, latency, cost, and accuracy, a designer must choose an operating point. Our analysis provides information for this decision: a designer may assess what performance to expect from each mode, so this can be weighed against the costs.
Costs span sensor hardware (per-pixel TCSPC for {\bf M1},{\bf M3}  increases power and fabrication complexity), compute and bandwidth ({\bf M1},{\bf M3} data is more voluminous than M2), and algorithmic complexity (each mode yields a different raw data format, requiring different algorithms). For example, {\bf M2} SPADs currently offer substantially higher spatial resolutions than {\bf M1},{\bf M3} arrays, but offer lower dynamic range --- a tradeoff our analysis helps a designer reason about.

%We analyze three SPAD operational modes. We present likelihood score functions for each. We calculate their CRLB and analyzed their fundamental relative errors. We show reconstruction of SPAD measurements with a diffusion model, using the operational modes. We present results both on simulated and real SPAD measurement data.
The fundamental relative error comparison reveals how a mode reporting {\bf (M1)} {\em Free-running Timestamps} yields a lower bound to the other modes, both for low and high event rates. The reconstruction statistics and visual results further support this operation mode  as a preferred configuration. 
The bounds particularly show the value of this mode, compared to {\bf (M2)} {\em Binary Bins}. As indicated in Ref.~\cite{ingle2021passive},  only the last timestamp $t_N$ needs to be reported by a SPAD sensor, to enjoy the full benefit of the M1 operation mode.

A SPAD sensor measures event times, which are noisy. {\em Timing jitter}~\cite{xu2025efficient} is the temporal uncertainty of recorded event. System time is measured by a clock based an oscillator~\cite{becker2005advanced}. An oscillator has random fluctuations leading to $\textbf{time drift}$ due to thermal changes and aging~\cite{allan1966statistics}.
Jitter and time drift may prove important to include in the analysis, in some cases, especially for exposure times longer than seconds.
However, this work neglects these  effects, for the following reasons. Current commercial sensors report typical timing jitter in picoseconds~\cite{piimaging2026spadlambda}.  Ref.~\cite{brown1992introduction}, provides typical parameters for temperature-compensated crystal oscillator (TCXO), found in many GPS receivers~\cite{bruggemann_csac}. After $1$[s], time drift is in picoseconds. In contrast, our work focuses on exposure times ranging between nanoseconds to milliseconds, such as in Ref.~\cite{seets2020motionadaptivedeblurringsinglephoton}. 

This work can possibly be generalized to determine bounds and optimal operation modes in active time-of-light SPAD sensing.
There, the operation degrees of freedom would include also the transmitted signal sequence.

In this work we used DPS to solve the reconstruction inverse problem. 
Currently, insufficient prior information is one of the main limiting factors in {DNN}-based reconstruction methods such as DPS. 
Better, more advanced methods to resolve likelihood score-based inverse problems may become available. Nevertheless, the fundamental analysis still stands.

% Any acknowledgments to only be included in camera ready
\ifpeerreview \else
\section*{Acknowledgments}
We thank Andreas Velten, Atul Ingle and Trevor Seets for their hard work providing the Fan and Tunnel data, and the support. We thank Vadim Holodovsky, Ina Talmon and Miri Haramati for their invaluable assistance. Yoav Schechner is the Mark and Diane
Seiden Chair in Science at the Technion. He is a Landau
Fellow supported by the Taub Foundation. His work was
conducted in the Ollendorff Minerva Center. Minvera is
funded through the BMBF. Mohit Gupta was supported by the National Science Foundation (CAREER Award \#1943149), the Office of Naval Research (N000142412155), and by a Sony Faculty Innovation Award. This work was supported by the PMRI – Peter Munk Research Institute - Technion, and by KLA. 
\fi

\bibliographystyle{IEEEtran}
\bibliography{references}

\clearpage

% 3. Supplementary Material Title
\begin{center}
    \textbf{\Large Supplementary Material}
\end{center}
\vspace{1em} % Adds a little breathing room after the title

% 4. Restart the section counter
\setcounter{section}{0}

% 5. Restart other counters (Highly Recommended)
\setcounter{equation}{0}
\setcounter{figure}{0}
\setcounter{table}{0}

\renewcommand{\thesection}{S\arabic{section}}
\renewcommand{\theequation}{S\arabic{equation}}
\renewcommand{\thefigure}{S\arabic{figure}}
\renewcommand{\thetable}{S\arabic{table}}

\peerreviewfalse

% Configuration for siunitx to allow bolding and +/- alignment
\sisetup{
    detect-all,
    separate-uncertainty,
    tight-spacing=true,
    table-align-uncertainty=true,
}
\robustify\bfseries % Makes bolding work inside S columns

% Enter your paper title below
\title{Fundamental Recovery Bounds for SPAD
Signals under Stationary Flux}

% Enter your author information before
% Note this is only necessary for the camera review. Submissions are anonymized.
% \author{Michael~Shell,~\IEEEmembership{Member,~IEEE,}
%         and~Jane~Doe,~\IEEEmembership{Life~Fellow,~IEEE}% <-this % stops a space
% \IEEEcompsocitemizethanks{\IEEEcompsocthanksitem M. Shell is with the Department
% of Electrical and Computer Engineering, Georgia Institute of Technology, Atlanta,
% GA, 30332.\protect\\
% % note need leading \protect in front of \\ to get a newline within \thanks.
% E-mail: see http://www.michaelshell.org/contact.html
% \IEEEcompsocthanksitem J. Doe is with Anonymous University.}% <-this % stops an unwanted space
% }
\author{Lior Dvir, Nadav Torem, Mohit Gupta, and Yoav Y. Schechner%
\IEEEcompsocitemizethanks{\IEEEcompsocthanksitem Lior Dvir, Nadav Torem, and Yoav Y. Schechner are with the Viterbi Faculty of Electrical and Computer Engineering, Technion-Israel Institute of Technology, Haifa 3200003, Israel.
\IEEEcompsocthanksitem Mohit Gupta is with the Department of Computer Sciences, University of Wisconsin-Madison, Madison, WI 53706, USA.}}

\IEEEtitleabstractindextext{%
\begin{abstract}
This supplementary material relates to the main manuscript on Fundamental Recovery Bounds for SPAD Signals under Stationary Flux. We present herein the duality between Erlang and Poisson functions. We expand on the background of Score-Based Diffusion. We show the validity of the PDF in the case of {\em Continuous Time Readouts and Domain}, and calculate the modes's expectation over the number of detected events. This document shows how the Fisher conditions are met for each operation mode. Finally, we provide more detail on the simulation process, training and results. 
\end{abstract}

\begin{IEEEkeywords} % Enter keywords here
 Single-photon Sensors, Single-photon Avalanche Diodes, Computational Photography, Diffusion Models, Quanta Imaging
\end{IEEEkeywords}
}

% Make Title
\ifpeerreview
\linenumbers \linenumbersep 15pt\relax 
\author{Paper ID \paperID\IEEEcompsocitemizethanks{\IEEEcompsocthanksitem This paper is under review for ICCP 2026 and the PAMI special issue on computational photography. Do not distribute.}}
\markboth{Anonymous ICCP 2026 submission ID \paperID}%
{}
\fi
\maketitle

\section{Erlang-Poisson Duality}
\label{sec:erlang_poisson}

\begin{figure}[t]
    \centering
    % Top Plot
    \includegraphics[width=1.0\columnwidth]{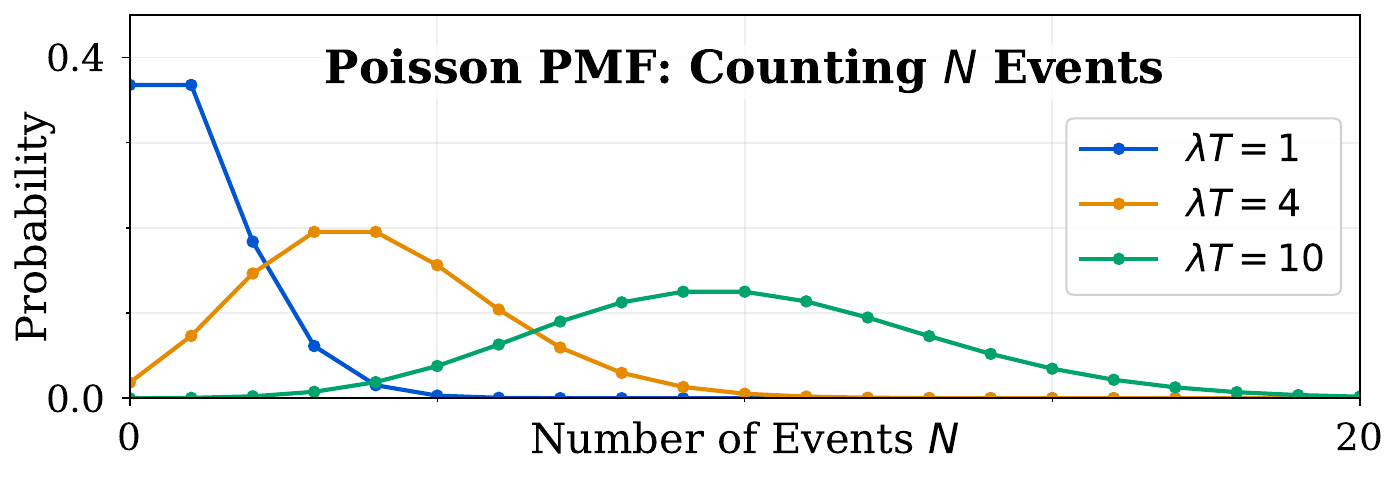}\\[5pt]
    % Middle Plot
    \includegraphics[width=1.0\columnwidth]{figures/erlang_pdf.pdf}\\[5pt]
    % Bottom Plot
    \includegraphics[width=1.0\columnwidth]{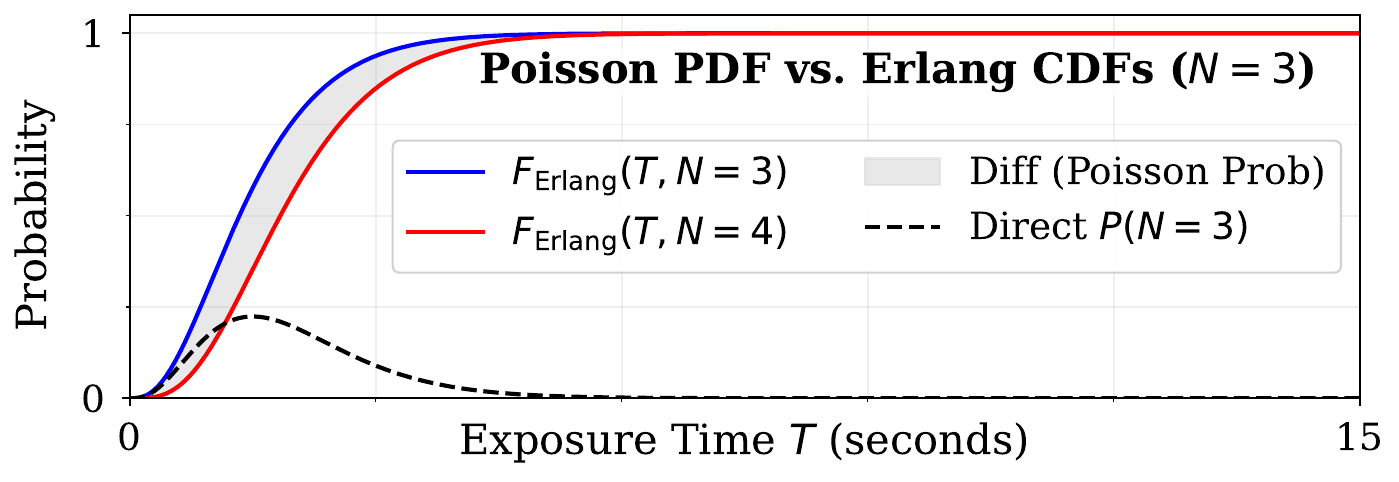}
    \caption{
    [Top] The Poisson PMF for varying expected counts $\lambda T$. It is  probability of observing exactly $N$ events.
    [Middle] The Erlang PDF of the detection time of the $N$-th event.
    [Bottom] The Poisson probability (shaded region) is the difference between two Erlang CDFs.}
    \label{fig:SUPP_poisson_erlang_plots}
\end{figure}

In this section we show the duality between Poisson process and Erlang function.
The Poisson PMF (Fig.~\ref{fig:SUPP_poisson_erlang_plots} herein, Top) models the probability of observing exactly $N$ independent events occurring in an interval ${t}$, given a known expected rate ${\lambda}$:
\begin{equation}
    p(N) = \frac{({\lambda}{t})^N \exp[-{\lambda}{t}]}{N!}.
    \label{eq:SUPP_SPAD_PAS_poisson_pmf_def}
\end{equation}
Here $N \in \{0, 1, 2, \dots\}$ represents the number of observed events (See Ref.~\cite{feller1968introduction} herein).

The Erlang PDF (Fig.~\ref{fig:SUPP_poisson_erlang_plots} herein, Middle) describes the distribution of the waiting time ${t}$ until the $N$-th event occurs in a Poisson process with rate ${\lambda}$. It is a special case of the Gamma distribution, where the shape parameter is an integer $N$ (See Ref.~\cite{ross2014introduction} herein):
\begin{equation}
    f_{\text{Erlang}}({t}; N) = \frac{{\lambda}^N {t}^{N-1} \exp[-{\lambda} {t}]}{(N-1)!}, \quad {t} \ge 0.
    \label{eq:SUPP_SPAD_PAS_erlang_pdf_def}
\end{equation}

From Ref.~\cite{ross2014introduction} herein, having exactly $N$ events by time $T$ is equivalent to having the $N$-th event before time $T$, while the $(N+1)$-th event occurs after time $T$. Therefore, the Poisson PMF can be expressed as the difference between two Erlang CDFs (Fig.~\ref{fig:SUPP_poisson_erlang_plots} herein, Bottom):
\begin{equation}
\begin{split}
    p_{\text{Poisson}}&(N) \\[10pt]
    &= \displaystyle \int_{0}^{T} f_{\text{Erlang}}(t;N) \, dt - \int_{0}^{T} f_{\text{Erlang}}(t;N+1) \, dt \\[10pt]
    &= F_{\text{Erlang}}(T;N) - F_{\text{Erlang}}(T;N+1) \;,
\end{split}
    \label{eq:SUPP_SPAD_PAS_poisson_erlang_duality}
\end{equation}
where 
\begin{equation}
    F_{\text{Erlang}}(T; N) = 1 - \sum_{i=0}^{N-1} \frac{(\lambda T)^i \exp[-\lambda T]}{i!} \;.
    \label{eq:SUPP_SPAD_PAS_erlang_cdf_explicit}
\end{equation}

\section{Score-based diffusion}
\label{sec:Supp_LDbackground}

A true object is expressed by a vector $\bm{x}$, where each element is denoted $x$. For example, $x$ can be the pixel value in a noiseless projected image, and $\bm{x}$ is a two dimensional (2D) noiseless image.
The object $\bm{x}$ is randomly sampled from nature, with a natural PDF denoted $p(\bm{x})$. We do not have hold of this object. Let $\mathcal{F}$ be a forward model that can be applied on $\bm{x}$. Let data ${\cal D}$ be the output ${\cal D}=\mathcal{F}(\bm{x})$. We want to digitally reconstruct $\bm{x}$ using a {\em score-based diffusion} (SBD) process. A reconstructed object is denoted $\bm{x}_0$.

In diffusion models, a {\em noising process} perturbs a signal by iterative addition of random noise. This process eventually reduces the signal-to-noise ratio (SNR) towards zero. Specifically, Ref.~\cite{song2021score} herein defines a noising process in continuous time, using a stochastic-differential equation (SDE),\footnote{For simplicity, in this work, we follow the standard Denoising Diffusion Probabilistic Models (DDPM) framework, of Ref.~\cite{ho2020denoising} herein.}
\begin{equation}
    d\bm{x} = -\beta_t\bm{x}_t dt/2 + \sqrt{\beta_t}d\bm{w}\;.
    \label{eq:SUPP_sde}
\end{equation}
Here $\bm{w}$ is a standard Wiener process (See Ref.~\cite{Wiener1923} herein), $dt>0$ and $\beta_t$ is a schedule of the noising process. 

A diffusion model defines a generative process that produces $\bm{x}_0$ as the reverse of a noising process. 

For inverse problems, the goal is to reconstruct $\bm{x}$ by generating a sample $\bm{x}_0$ from the posterior distribution $p(\bm{x}|\mathcal{D})$. In this case, $dt<0$ and the {\em reverse process} SDE is given (See Ref.~\cite{song2021score} herein) by,
\begin{equation}  
     d\bm{x} = 
     \left[ 
     -\beta_t\bm{x}_t/2 
     - \beta_t\,\nabla_{\bm{x}_t} \log p(\bm{x}_t|\mathcal{D}) 
     \right] dt + \sqrt{\beta_t}\, d\bar{\bm{w}}\;.
\label{eq:SUPP_reverse_diff}
\end{equation}
Using Bayes theorem, 
\begin{equation}
    \nabla_{\bm{x}_t} \log p(\bm{x}_t|\mathcal{D}) = \nabla_{\bm{x}_t} \log p(\bm{x}_t) +  \nabla_{\bm{x}_t}\log p(\mathcal{D|}\bm{x}_t)\;.
\end{equation}
The term $\nabla_{\bm{x}_t} \log p({\bm{x}}_t)$ is the {\em score function of the prior term}, and the term $\nabla_{{\bm{x}}_t} \log p({\cal D}|{\bm{x}_t})$ is the {\em score function of the likelihood}, also called the log-likelihood gradient. To run an SBD algorithm, both score functions must be computed. 

Computations are in discrete steps indexed $k$, thus ${\bm{x}}_k$ stands for ${\bm{x}}_t$. 
%In essence, this DNN learns and then expresses, approximately,  priors of true objects. 
Let $k$ % \in [K, \dots, 1]$  
be an iteration %count-down 
index.
Define 
\begin{equation}
   \alpha_k=1-\beta_k  \;, ~~~~~~~
    \bar{\alpha}_k=\Pi_{\iota=1}^k\alpha_{\iota}\;.
    \label{eq:SUPP_BG_alpha_beta}
\end{equation} 
In discrete settings, the noising process (\ref{eq:SUPP_sde}) herein is
\begin{equation}
    \bm{x}_k = \sqrt{{\alpha}_k}\,\bm{x}_{k-1} + \sqrt{1 - {\alpha}_k}\,\bm{\epsilon}_{k-1}\;.
\label{eq:SUPP_too_lazy}
\end{equation}
Here $\bm{\epsilon}_{k-1}\sim\mathcal{N}(0,\bm{I})$, where $\mathcal{N}(\cdot,\cdot)$ is the normal PDF and $\bm{I}$ is a unit matrix. Eq.~(\ref{eq:SUPP_too_lazy}) herein is solved by
\begin{equation}
    \bm{x}_k = \sqrt{\bar{\alpha}_k}\,\bm{x}_{0} + \sqrt{1 - \bar{\alpha}_k}\,\bm{\epsilon}\;.
\label{eq:SUPP_too_lazy_2}
\end{equation}
Here $\bm{\epsilon}$, modeled as $\bm{\epsilon}\sim\mathcal{N}(0,\bm{I})$,
is the discrepancy (up to scale) between a sample 
$\bm{x}_0$ from the posterior and  %temporary 
state  $\bm{x}_k$. 

In SBD,  $\nabla_{\bm{x}_k} \log p({\bm{x}}_k)$ is not derived explicitly, but it is approximated by the output $\bm{s}_{\bm{\theta}}({\bm{x}}_k,k)$ of a trained DNN  ({\em score DNN}). Its parameters are $\bm{\theta}$. 
The function  $\bm{s}_{\bm{\theta}}({\bm{x}}_k,k)$ 

essentially learns to extract the random noise sample ${\bm \epsilon}$, given a noisy object ${\bm{x}}_k$ (See Ref.~\cite{zhu2023denoising} herein). 
Regarding $\nabla_{{\bm{x}_k}} \log p({\cal D}|{\bm{x}_k})$, computation may be a significant challenge for a non-linear forward model $\mathcal{F}$. The dependency of ${\cal D}$ on a {\em true} object $\bm{x}$ is often known through $\mathcal{F}$. However, this is not necessarily true regarding the statistical dependency of ${\cal D}$ on the noisy state ${\bm{x}}_k$. Often, this dependency is intractable, when $\mathcal{F}(\cdot)$ is nonlinear and random, as is the case with SPAD signals. 

This difficulty is addressed by 
an algorithm termed diffusion posterior sampling (DPS) (See Ref.~\cite{chung2023diffusion} herein).
Here $k\in [K,\dots,1]$  in a countdown. 
At each iteration, DPS estimates $\bm{x}_0$ as
\begin{equation}
   \hat{\bm{x}}_0({\bm{x}}_k) = \mathbb{E}[\bm{x}_0|\bm{x}_k]=\frac{1}{\sqrt{\bar{\alpha}_k}}\left[ {\bm{x}}_k + (1-\bar{\alpha}_k)\bm{s}_{\bm{\theta}}({\bm{x}}_k,k) \right]\;.
   \label{eq:SUPP_BG_LIKELIHOOD_x0_estimation}
\end{equation}
%$k$ % \in [K, \dots, 1]$   
Then, $\hat{\bm{x}}_0$ is utilized to approximate the score function of the likelihood:  
$\nabla_{\bm{x}_k} \log p({\mathcal{D}}| \bm{x}_k) \simeq \nabla_{\bm{x}_k} \log p({\mathcal{D}} | \hat{\bm{x}}_0)$. 
This approximation %Computing $\nabla_{\hat{\bm{x}}_0} \log p({\cal D}|\hat {\bm{x}}_0)$ 
is tractable,  because $\hat {\bm{x}}_0$ behaves as a clean object, for which $\log p({\cal D}|\hat {\bm{x}}_0)$ is well defined.

%To calculate the derivative $\nabla_{\bm{x}_k} \log p({\mathcal{D}} | \hat{\bm{x}}_0)$ 
Define a Jacobian for Eq.~(\ref{eq:BG_LIKELIHOOD_x0_estimation}) herein,
%$\bm{J}$ is utilized,
\begin{equation}
    \bm{J}=\frac{\partial\hat{\bm{x}}_0}{\partial\bm{x}_k}=\frac{1}{\sqrt{\bar{\alpha}_k}}\left[ {\bm{I}} + {(1-\bar{\alpha}_k)}\nabla_{\bm{x}_k}\bm{s}_{\bm{\theta}}({\bm{x}}_k,k) \right]\;.
    \label{eq:SUPP_BG_DPS_J}
\end{equation}
Here $\nabla_{\bm{x}_k}\bm{s}_{\bm{\theta}}({\bm{x}}_k,k)$ is a Jacobian relating the output vector of $\bm{s}_{\bm{\theta}}$ with respect to an input vector ${\bm{x}}_k$. In practice, $\nabla_{\bm{x}_k}\bm{s}_{\bm{\theta}}({\bm{x}}_k,k)$ is calculated through backpropagation, exploiting the fact that  $\bm{s}_{\bm{\theta}}$ is a differentiable DNN. Let ${\bm \eta}\sim\mathcal{N}(0,\bm{I})$. Define a step size $\rho$.
The iterative rule in DPS follows,
\begin{eqnarray}
        \bm{x'}_{k-1}=
         \frac{\sqrt{\alpha_k}(1-\bar{\alpha}_{k-1})}
              {1-\bar{\alpha}_{k}}\bm{x}_k
        +\frac{\sqrt{\bar{\alpha}_{k-1}}\beta_k}
              {1-\bar{\alpha}_{k}}
        \hat{\bm{x}}_0(\bm{x}_k)+\sigma_k\bm{\eta}
        \\[0.2cm]
        \bm{x}_{k-1}=\bm{x'}_{k-1}+\rho \bm{J} \nabla_{\hat{\bm{x}}_0} \log p[{\cal D}|\hat{\bm{x}}_0(\bm{x}_k)]\;,~~~~~~~~~~~~~~~~~~~~~~
\label{eq:SUPP_BG_DPS_update}
\end{eqnarray}
where 
$\sigma_K>\sigma_{K-1} \ldots > \sigma_1=0$. %avoiding a collapse to a local maximum throughout the SBD algo
%$\sigma_k\rightarrow 0$. %avoiding a collapse to a local maximum throughout the SBD algorithm. 
Overall, the parameters of the algorithm are 
$\rho,\{ \sigma_k\}_{k=1}^K,\{\alpha_k\}_{k=1}^K$.

\section{M1 PDF Validity}
\label{sec:forward_model}

In this section we justify the validity of the {\em Continuous Time Readout and Domain} as a PDF.
In Sec. 3.1 of the main manuscript, the likelihood model of {\em Continuous Time Readouts and Domain} was presented as is. We will now derive it. Although the derivation is not the same as in Ref.~\cite{ingle2021passive} herein, the resulting log-likelihood function is similar.
Let $T$ be the exposure time of a SPAD pixel. 
The total number of photons detected by the pixel during this interval is denoted by the random variable $N$. The detection event times are denoted by a set of random variables $\{t_i\}_{i=1}^{N}$,  where $0 \le t_1 < \dots < t_N \le T$.
%Let $\tau_{\rm{dead}}$ be the sensor dead time.
Due to the dead time, the time intervals satisfy
\begin{equation}
    t_{i} - t_{i-1} \ge \tau_{\rm{dead}} \quad \forall i \in \{2, \dots, N\} \;.
    \label{eq:SUPP_TEMP_time_diff}
\end{equation}

The PDF for detecting the first photon at time $t_1$ is 
\begin{equation}
   \lambda \exp[- \lambda t_1]\;.
   \label{eq:SUPP_SPAD_FM_first_photon_arrival}
\end{equation}
The PDF of each of the consequent event times is  derived by Eq.~(1) of the main manuscript. So, their joint PDF is
\begin{equation}
    \Psi = \prod_{i=2}^{N} \lambda \exp\left[ -\lambda (t_i-t_{i-1}-\tau_{\rm{dead}}) \right] \;.
    \label{eq:SUPP_SPAD_FM_photon_inter_arrivals}
\end{equation}
The probability that no detection events happened in the remaining exposure time is derived by,
\begin{equation}
    \exp\left[-\lambda (T-t_N-\tau_{\rm{dead}})\right]\;.
    \label{eq:SUPP_SPAD_FM_tail}
\end{equation}
The probability no detection event happens at all during exposure time $T$ is also derived by,
\begin{equation}
    \exp\left[-\lambda T\right]\;.
    \label{eq:SUPP_SPAD_FM_no_events}
\end{equation}
Eqs.~(\ref{eq:SUPP_SPAD_FM_first_photon_arrival},\ref{eq:SUPP_SPAD_FM_photon_inter_arrivals},\ref{eq:SUPP_SPAD_FM_tail}) herein form the PDF, or likelihood, to have a set of detection events described by $(N,\{t_i\}_{i=1}^{N})$, given $\lambda$ and $N>0$:
\begin{equation}
\begin{split}
    p(N,&\{t_i\}_{i=1}^{N}|\lambda) \\[10pt]
    &= \lambda \exp[- \lambda t_1] 
       \Psi 
       \exp[-\lambda (T-t_N-\tau_{\rm{dead}})] \;.
    \label{eq:SUPP_SPAD_FM_forward_model}
\end{split}
\end{equation}
In the instance of no detection events $N=0$, the likelihood is described by Eq.~(\ref{eq:SUPP_SPAD_FM_no_events}) herein
\begin{equation}
    p(N=0|\lambda)=\exp\left[-\lambda T\right]\;.
    \label{eq:SUPP_SPAD_FM_forward_model_no_events}
\end{equation}
To verify that Eq.~(3) of the main manuscript constitutes a valid PDF, it must satisfy two conditions: non-negativity and normalization.
The first condition, 
\begin{equation}
    \begin{split}
        p(N, \{t_i\}_{i=1}^{N}|\lambda,{\cal S}_{\rm cont}) &\ge 0, \quad N\in\{1,2,\dots\} \\
        p(N=0|\lambda) &\ge 0
    \end{split}
    \label{eq:SUPP_first_cond}
\end{equation}
is satisfied by definition. Since the rate parameter $\lambda$ is positive and the exponential function is strictly positive for all real arguments, the likelihood function, constructed as a product of these terms, is strictly positive.\\
The second condition requires that a sum over all $N$, of the respective integrals over the domain of event times equal unity:
\begin{equation}
\begin{split}
        \sum_{N=0}^{\infty}&\left[\int_{t_N=0}^{T}\densedots
    \int_{t_2=0}^{t_3} 
    \int_{t_1=0}^{t_2} p(N, \{t_i\}_{i=1}^{N}|\lambda,{\cal S}_{\rm cont}) \, dt_1 \densedots dt_N\right] \\[10pt] &= 1
    \;.
\end{split}
    \label{eq:SUPP_SPAD_FM_model_proof}
\end{equation}
The proof is split into two cases according to the arrival time $t_N$ of the last measured event. 

%%%%%%%%%%%%%%%%%%%%%%%%%%%%%%%%%%%%%%
\subsection{Case I}
\label{sec:case_I}
In this section we formulate the likelihood in the case where the time left after the $N$-th event is longer than $\tau_{\rm{dead}}$. Then we present the result in the Erlang form.

In $\texttt{Case I}$, 
\begin{equation}
    p(N, \{t_i\}_{i=1}^{N}|\lambda,{\cal S}_{\rm cont})=\lambda^{N} \exp\left[-\lambda \left(T -N\tau_{\rm{dead}}\right)\right]\;,
    \label{eq:SUPP_p_case_i}
\end{equation}
and
\begin{equation}
    0<t_N\leq T-\tau_{\rm{dead}}
    \;.
    \label{eq:SUPP_SPAD_FM_case_I}
\end{equation}
For easier integration bounds, we define
\begin{equation}
\begin{split}
    &u_0=0 \\
    &u_i =t_i +(1-i)\tau_{\rm{dead}} \quad \quad \text{for $i \in \{1, 2, \dots\, N\}$}\;. 
\end{split}
\label{eq:SUPP_SPAD_case_I_change_vars_formula}
\end{equation}
Then,
\begin{equation}
\begin{split}
    %u_0&=0\;. \\[5pt]
    u_1 &= t_1\;. \\[5pt]
    u_2 &= t_2-\tau_{\rm{dead}}\;. \\
    \vdots \\
    u_N&=t_N-(N-1)\tau_{\rm{dead}}\;. \\[5pt]
\end{split}
\label{eq:SUPP_SPAD_FM_case_I_change_of_vars}
\end{equation}
The variables $\{u_i\}_{i=2}^N$ satisfy
\begin{equation}
    \begin{split}
    u_i - u_{i-1}&=[t_i-(i-1)\tau_{\rm{dead}}] - [t_{i-1}-(i-2)\tau_{\rm{dead}}] \\[10pt] 
    &= t_i-t_{i-1}-\tau_{\rm{dead}}\;.
    \end{split}
    \label{FM_change_of_vars_diff}
\end{equation}
Assigning back to Eq.~(\ref{eq:SUPP_SPAD_FM_photon_inter_arrivals}) herein, define
\begin{equation}
\begin{split}
    \Omega &= \lambda\exp[-\lambda u_1]\prod_{i=2}^{N} \lambda \exp\left[ -\lambda (u_i-u_{i-1})\right] \\
    &= \lambda\exp[-\lambda (u_1-u_0)]\prod_{i=2}^{N} \lambda \exp\left[ -\lambda (u_i-u_{i-1})\right] \\
    &=\prod_{i=1}^{N} \lambda \exp\left[ -\lambda (u_i-u_{i-1})\right]
\end{split}
\label{eq:SUPP_SPAD_FM_psi_u}
\end{equation}
Following Eq.~(\ref{eq:SUPP_SPAD_FM_forward_model}) herein,
\begin{equation}
    \begin{split}
        p&(N, \{u_i\}_{i=1}^{N}|\lambda,{\cal S}_{\rm cont}) \\[10pt]
        &=\Omega \exp[-\lambda \{T-(u_{N}+[N-1]\tau_{\rm{dead}})-\tau_{\rm{dead}}\}] \\[10pt]
        &=\lambda^{N} \exp[-\lambda u_{N}]\exp[-\lambda(T-N\tau_{\rm{dead}}-u_{N})] \\[10pt]
        &=\lambda^{N} \exp[-\lambda(T-N\tau_{\rm{dead}})]\;.
    \end{split}
    \label{eq:SUPP_SPAD_FM_case_I_assigning_u}
\end{equation}
Integrating over $\{u_i\}_{i=1}^{N}$ to calculate $p_{\tt I}(N|\lambda,{\cal S}_{\rm cont})$ under the integral bounds $0<u_1<u_2<\dots<u_N<(T-N\tau_{\rm{dead}})$,
\begin{equation}
    \begin{split}
        &p_{\tt I}(N|\lambda,{\cal S}_{\rm cont})\\[10pt]
        &=\int_{u_N=0}^{T-\tau_{\rm{dead}}}\densedots
        \int_{u_1=0}^{u_2}\lambda^{N} \exp[-\lambda(T-N\tau_{\rm{dead}})]du_1\densedots du_N \\[10pt]
        &=\lambda^{N} \exp[-\lambda(T-N\tau_{\rm{dead}})]\int_0^{T-\tau_{\rm{dead}}}\densedots \int_0^{u_2}1\ du_1\densedots du_N.
    \end{split}
    \label{eq:SUPP_SPAD_FM_case_I_integral_over_u}
\end{equation}
We denote
\begin{equation}
    \mathcal{A}_N=\int_{u_{N-1}=0}^{u_N}
    \dots\int_{u_1=0}^{u_2}
    1\ du_1\dots du_{N-1}\;.
    \label{eq:SUPP_SPAD_CASE_I_inner_integral}
\end{equation}
Integrating Eq.~(\ref{eq:SUPP_SPAD_CASE_I_inner_integral}) herein for different values of $N$ gives:
\begin{equation}
\begin{split}\\
    \mathcal{A}_1 &=  1 \\[10pt]
    \mathcal{A}_2 &=\displaystyle \int_0^{u_2} 1 \, du_1 = u_2 \\[10pt]
    \mathcal{A}_3 &=\displaystyle \int_0^{u_3} u_2 \, du_2 = \frac{u_3^2}{2} \\
        &\vdots \\
    \mathcal{A}_N &= \displaystyle \int_0^{u_N} \frac{u_{N-1}^{N-2}}{(N-2)!} \, du_{N-1} = \frac{u_N^{N-1}}{(N-1)!} \;.
\end{split}
\label{eq:SUPP_SPAD_FM_case_I_integral_volume}
\end{equation}
The final integral over $\mathcal{A}_N$ follows
\begin{equation}
\begin{split}
    \displaystyle \int_0^{T-N\tau_{\rm{dead}}} \mathcal{A}_N \, du_{N} &=
    \displaystyle \int_0^{T-N\tau_{\rm{dead}}} \frac{u_N^{N-1}}{(N-1)!} \, du_{N} \\[10pt]
    &= \frac{(T-N\tau_{\rm{dead}})^{N}}{N!} \;. 
\end{split}
   \label{eq:SUPP_lastint}
\end{equation}
Substituting Eq.~(\ref{eq:SUPP_lastint}) herein back to Eq.~(\ref{eq:SUPP_SPAD_FM_case_I_integral_over_u}) herein,
\begin{equation}
    \begin{split}
        p_{\tt I}(N|\lambda,{\cal S}_{\rm cont})&=\lambda^{N} \exp[-\lambda(T-N\tau_{\rm{dead}})]\frac{(T-N\tau_{\rm{dead}})^{N}}{N!} \\[10pt]
        &=\frac{[\lambda(T-N\tau_{\rm{dead}})]^{N}\exp[-\lambda(T-N\tau_{\rm{dead}})]}{N!}\;.
    \end{split}
    \label{eq:SUPP_SPAD_FM_case_I_result_poisson}
\end{equation}
We note that for $N=0$, Eq.~(\ref{eq:SUPP_SPAD_FM_case_I_result_poisson}) herein coincides with Eq.~(\ref{eq:SUPP_SPAD_FM_forward_model_no_events}) herein,
\begin{equation}
\begin{split}
     &\frac{[\lambda(T-0\cdot\tau_{\rm{dead}})]^{0}\exp[-\lambda(T-0\cdot\tau_{\rm{dead}})]}{0!}\\[10pt]&=\exp{[-\lambda T]}\\[10pt]&=p_{\tt 0}(N|\lambda,{\cal S}_{\rm cont})\;.   
\end{split}
    \label{eq:SUPP_SPAD_FM_N_coincide}
\end{equation}
Therefore, Eq.~(\ref{eq:SUPP_SPAD_FM_case_I_result_poisson}) herein is true for $N\in\{0,1,2,\dots\}$. Using Eq.~(\ref{eq:SUPP_SPAD_PAS_poisson_erlang_duality}) herein,
\begin{equation}
\begin{split}
     p_{\tt I}&(N|\lambda,{\cal S}_{\rm cont})\\[10pt]&=F_{\text{Erlang}}(T-N\tau_{\rm{dead}}|N)-F_{\text{Erlang}}(T-N\tau_{\rm{dead}}|N+1)\;.
    \label{eq:SUPP_SPAD_FM_case_I_result_erlang}   
\end{split}
\end{equation}

%%%%%%%%%%%%%%%%%%%%%%%%%%%%%%%%%%%%%%
\subsection{Case II}
\label{sec:case_II}
In this section we formulate the likelihood in the case where the time left after the $N$-th event is shorter than $\tau_{\rm{dead}}$. We then present the result in the Erlang form along Eq.~(\ref{eq:SUPP_SPAD_PAS_poisson_erlang_duality}) herein.
In $\texttt{Case II}$,
\begin{equation}
    p(N, \{t_i\}_{i=1}^{N}|\lambda,{\cal S}_{\rm cont})=\lambda^{N} \exp\left[-\lambda \{t_N-(N-1)\tau_{\rm{dead}}\}\right]\;,
\end{equation}
and
\begin{equation}
    T-\tau_{\rm{dead}}<t_N\leq T\;.
    \label{eq:SUPP_SPAD_FM_case_II}
\end{equation}
Using change of variables from Eq.~(\ref{eq:SUPP_SPAD_case_I_change_vars_formula}) herein,
\begin{equation}
    \begin{split}
        p(N, \{u_i\}_{i=1}^{N}|\lambda,{\cal S}_{\rm cont}) &=\prod_{i=1}^{N}\lambda \exp[-\lambda (u_i-u_{i-1})] \\[10pt]
        &=\lambda^{N}\exp[-\lambda\sum_{i=1}^{N}(u_i-u_{i-1})] \\[10pt]
        &=\lambda^{N}\exp[-\lambda u_N]\;.
    \end{split}
    \label{eq:SUPP_SPAD_FM_case_II_forward_model_cahnged_vars}
\end{equation}
Under change of variables from Eq.~(\ref{eq:SUPP_SPAD_case_I_change_vars_formula}) herein, Eq.~(\ref{eq:SUPP_SPAD_FM_case_II}) herein becomes
\begin{equation}
    T-N\tau_{\rm{dead}}<u_N\leq T-(N-1)\tau_{\rm{dead}}\;.
    \label{eq:SUPP_SPAD_FM_case_II_changed_vars}
\end{equation}
Integrating over $\{u_i\}_{i=1}^{N}$ to calculate $p(N|\lambda,{\cal S}_{\rm cont})$ under the integral bounds $0<u_1<u_2<\densedots<u_N<[T-(N-1)\tau_{\rm{dead}}]$, with the same methodology as in Eqs.~(\ref{eq:SUPP_SPAD_CASE_I_inner_integral}, \ref{eq:SUPP_SPAD_FM_case_I_integral_volume}) herein,
\begin{equation}
    \begin{split}
        p_{\tt II}&(N|\lambda,{\cal S}_{\rm cont})\\[10pt]
        &=
        \int_{u_N=T-N\tau_{\rm{dead}}}^{T-(N-1)\tau_{\rm{dead}}}\dots
        \int_{u_1=0}^{u_2}\lambda^{N} \exp[-\lambda u_N]du_1\dots du_N \\[10pt]
        &=\int_{T-N\tau_{\rm{dead}}}^{T-(N-1)\tau_{\rm{dead}}}\lambda^{N} \exp[-\lambda u_N]\mathcal{A}_Ndu_N \\[10pt]
        &=\int_{T-N\tau_{\rm{dead}}}^{T-(N-1)\tau_{\rm{dead}}}\lambda^{N} \exp[-\lambda u_N]\frac{u_N^{N-1}}{(N-1)!}du_N\;.
    \end{split}
    \label{eq:SUPP_SPAD_FM_case_II_integral_over_u}
\end{equation}
The integrand in Eq.~(\ref{eq:SUPP_SPAD_FM_case_II_integral_over_u}) herein is the Erlang PDF shown in Eq.~(40) of the main manuscript. Hence,
\begin{equation}
\begin{split}
     p_{\tt II}&(N|\lambda,{\cal S}_{\rm cont})\\[10pt]&=F_{\text{Erlang}}[T-(N-1)\tau_{\rm{dead}}|N]-F_{\text{Erlang}}[T-N\tau_{\rm{dead}}|N]\;.   
\end{split}
    \label{eq:SUPP_SPAD_FM_case_II_result_erlang}
\end{equation}
In the instance of no detection events $N=0$, there is no dead time $\tau_{\rm{dead}}$ affecting the model. Therefore, the situation described by $\texttt{case II}$ cannot occur and 
\begin{equation}
    p_{\tt 0}(N=0|\lambda,{\cal S}_{\rm cont})=0\;.
    \label{eq:SUPP_SPAD_FM_case_II_no_events}
\end{equation}
We note that $F_{\text{Erlang}}[T|N=0]$ and $F_{\text{Erlang}}[T+\tau_{\rm{dead}}|N=0]$ are the probabilities to detect at least $0$ events by times $T$ and $T+\tau_{\rm{dead}}$ respectively:
\begin{equation}
    \begin{split}
        F_{\text{Erlang}}&(T| N=0) \\
        &= 1 - \cancelto{0}{\sum_{i=0}^{0-1}} \frac{(\lambda T)^i \exp[-\lambda T]}{i!}=1 \\[10pt]
        F_{\text{Erlang}}&(T+\tau_{\rm{dead}}|N=0) \\ 
        &= 1 - \cancelto{0}{\sum_{i=0}^{0-1}} \frac{[\lambda (T+\tau_{\rm{dead}})]^i \exp[-\lambda (T+\tau_{\rm{dead}})]}{i!}=1\;.
    \end{split}
    \label{eq:SUPP_SPAD_FM_case_II_erlang_no_events}
\end{equation}
Assigning $N=0$ in Eq.~(\ref{eq:SUPP_SPAD_FM_case_II_result_erlang}) herein and using Eq.~(\ref{eq:SUPP_SPAD_FM_case_II_erlang_no_events}) herein gives
\begin{equation}
\begin{split}
    p_{\tt 0}&(N=0|\lambda,{\cal S}_{\rm cont})\\[10pt]
    &=F_{\text{Erlang}}(T+\tau_{\rm{dead}}|N=0)-F_{\text{Erlang}}(T| N=0)\\
    &=1-1=0\;,
\end{split}
\label{eq:SUPP_SPAD_FM_case_II_erlang_no_events_coincide}
\end{equation}
which coincides with Eq.~(\ref{eq:SUPP_SPAD_FM_case_II_no_events}) herein. Therefore, Eq.~(\ref{eq:SUPP_SPAD_FM_case_II_result_erlang}) herein is true for $N\in\{0,1,2,\dots\}$.
%%%%%%%%%%%%%%%%%%%%%%%%%%%%%%%%%%%%%%
\subsection{Combined Case}
\label{sec:combined_case}
In this section we add up Eqs.~(\ref{eq:SUPP_SPAD_FM_case_I_result_erlang},\ref{eq:SUPP_SPAD_FM_case_II_result_erlang}) herein  to calculate the whole $p(N|\lambda,{\cal S}_{\rm cont})$. Then, we integrate to 1 to prove it is a valid PMF.
We note $p_{\tt{I}}(N|\lambda,{\cal S}_{\rm cont})$ and $p_{\tt{II}}(N|\lambda,{\cal S}_{\rm cont})$ represent the integral of the same PMF over two non-overlapping regions that together span the entire possible range of event sequences.
The combined PMF $p(N|\lambda,{\cal S}_{\rm cont})$ can be calculated:
\begin{equation}
    \begin{split}
        &p(N|\lambda,{\cal S}_{\rm cont})=p_{\tt{ I}}(N|\lambda,{\cal S}_{\rm cont})+p_{\tt{ II}}(N|\lambda,{\cal S}_{\rm cont}) \\[10pt]
        &=\{F_{\text{Erlang}}[T-N\tau_{\rm{dead}}|N]-F_{\text{Erlang}}[T-N\tau_{\rm{dead}}|N+1]\} \\[10pt]
        &+\{F_{\text{Erlang}}[T-(N-1)\tau_{\rm{dead}}|N]-F_{\text{Erlang}}[T-N\tau_{\rm{dead}}|N]\}\\[10pt]
        &=F_{\text{Erlang}}[T-(N-1)\tau_{\rm{dead}}|N] \\
        &\ \ \ \ \ \ \ \ \ \ \ \ \ \ \ \ \ \ \ \ \ \ \ \ \ \ \ \ \ \ \ \ \ \ \ \ \ \ \ \ \ \ \ \ -F_{\text{Erlang}}[T-N\tau_{\rm{dead}}|N+1]\;.
    \end{split}
    \label{eq:SUPP_SPAD_FM_final_erlang}
\end{equation}
The maximum number of event detections $N^{\rm max}$ by exposure time $T$ is
\begin{equation}
    N^{\rm max}\triangleq
    \left\lceil \frac{T}{\tau_{\rm{dead}}} 
    \right\rceil\;.
    \label{eq:SUPP_SPAD_FM_final_erlang_M}
\end{equation}
Let $S_{N}$ be defined as
\begin{equation}
    S_{N}\triangleq F_{\text{Erlang}}[T-(N-1)\tau_{\rm{dead}}|N]\;.
    \label{eq:SUPP_SPAD_FM_final_erlang_s}
\end{equation}
By the definition of the Erlang CDF,
\begin{equation}
\begin{split}
    &S_0=p(\text{At least 0 events happened by time} \ T)=1 \\[10pt]
    &S_{N^{\rm max}+1}\\[10pt]&=p(\text{At least} \ N^{\rm max} \text{+ 1 events happened by time} \ T)=0\;.
\end{split}
\label{eq:SUPP_SPAD_FM_erlang_proof}
\end{equation}
Using Eqs.~(\ref{eq:SUPP_SPAD_FM_final_erlang}, \ref{eq:SUPP_SPAD_FM_final_erlang_M}, \ref{eq:SUPP_SPAD_FM_final_erlang_s}, \ref{eq:SUPP_SPAD_FM_erlang_proof}) herein then proves Eq.~(\ref{eq:SUPP_SPAD_FM_model_proof}) herein,
\begin{equation}
    \begin{split}
        \sum_{N=0}^{M}&{p(N|\lambda,{\cal S}_{\rm cont})}\\[10pt]
        &=(S_0-S_1)+(S_1-S_2)+\dots+(S_{N^{\rm max}}-S_{N^{\rm max}+1}) \\[10pt]
        &=S_0-S_{N^{\rm max}+1} \\[10pt]
        &=1-0 \\[10pt]
        &=1\;.
    \end{split}
    \label{eq:SUPP_SPAD_FM_end_of_forward_model_proof}
\end{equation}

%%%%%%%%%%%%%%%%
\section{Expected Number of Events in M1}
\label{sec:Nerlang}

We use the survival function identity from Ref.~\cite{feller1968introduction} herein,
\begin{equation}
   \mathbb{E}\left[ N\right] = \sum_{n=0}^\infty n\cdot p(N=n|\lambda,{\cal S}_{\rm cont}) = \sum_{n=1}^\infty p(N\geq n|\lambda,{\cal S}_{\rm cont})\;.
   \label{eq:SUPP_CRLB_survival}
\end{equation}
By definition of $p(N\geq n|\lambda,{\cal S}_{\rm cont})$,
\begin{equation}
   \begin{split}
       \mathbb{E}\left[ N\right] &= \sum_{n=1}^\infty \sum_{i=n}^\infty p(N = i|\lambda,{\cal S}_{\rm cont}) \;.
   \end{split}
   \label{eq:SUPP_CRLB_sub_p}
\end{equation}
We note for $N > N^{\rm max}$ from %Eq.~(\ref{eq:SUPP_SPAD_FM_final_erlang_M}):
\begin{equation}
   p(N|\lambda,{\cal S}_{\rm cont})=0 \;.
\label{eq:SUPP_CRLB_bigger_than_M}
\end{equation}
Therefore, we get the telescopic term
\begin{equation}
   \begin{split}
       &\mathbb{E}\left[ N\right] = 
       \sum_{n=1}^\infty \sum_{i=n}^{N^{\rm max}} p(N = i|\lambda,{\cal S}_{\rm cont})\;.
   \end{split}
   \label{eq:SUPP_CRLB_max_M}
\end{equation}
From Eq.~(\ref{eq:SUPP_SPAD_FM_final_erlang}) herein,
\begin{equation}
   \begin{split}
       &\mathbb{E}\left[ N\right]
       \mkern-5mu \\
       \mkern-5mu&=\sum_{n=1}^\infty\{F_{\text{Erlang}}[T\minus(n\minus1)\tau_{\rm{dead}}|n]
       \minus F_{\text{Erlang}}[T\minus\tau_{\rm{dead}}|n\plus1] \\[10pt]
       &+ F_{\text{Erlang}}[T\minus n\tau_{\rm{dead}}|n\plus 1]
       \minus F_{\text{Erlang}}[T\minus (n\plus 1)\tau_{\rm{dead}}|n\plus2] \\
       &\vdots \\
       &+ F_{\text{Erlang}}
       [T\minus (N^{\rm max}\minus 1)\tau_{\rm{dead}}|N^{\rm max}]\\[10pt]
       &~~~~~~~~~~~~~~~~~~~~~~~~~~~~~~~~~~~\minus F_{\text{Erlang}}
       [T\minus N^{\rm max}\tau_{\rm{dead}}|N^{\rm max}\plus1]\} \;.
   \end{split}
\label{eq:SUPP_CRLB_telescopic}
\end{equation}
The inner telescopic terms cancel each other out. 
From Eq.~(\ref{eq:SUPP_SPAD_FM_erlang_proof}) herein,
\begin{equation}
 F_{\text{Erlang}}[T-N^{\rm max}\tau_{\rm{dead}}|N^{\rm max}+1,\lambda]=0
   \label{eq:SUPP_FNmax}
\end{equation}
%From Eqs.~(\ref{eq:SUPP_CRLB_telescopic},\ref{eq:SUPP_FNmax}), 
\begin{equation}
%    \begin{split}
        \mathbb{E}\left[ N\right] 
%        &=\sum_{n=1}^\infty 
%           F_{\text{Erlang}}[T-(n-1)\tau_{\rm{dead}}|n,\lambda] \\[10pt] 
        %&
        =\sum_{n=1}^{N^{\rm max}} 
           F_{\text{Erlang}}[T-(n-1)\tau_{\rm{dead}}|n,\lambda] \;.
%    \end{split}
    \label{eq:SUPP_CRLB_telescopic_cancel}
\end{equation}

%%%%%%%%%%%%%%%%%%%%%%%%%%%%%

\section{Fisher Conditions}
\label{sec:conditions}
In this section we justify the conditions are met to calculate the Fisher Information for the three SPAD operational configurations detailed in the paper.
For the identity in Eq.~(14) of the main manuscript to be true, three conditions need to be met (See Ref.~\cite{kay1993fundamentals} herein):
\begin{enumerate}[label=\arabic*)]
    \item \textbf{Support independence}\\
    Support of $p({\cal D}|\lambda,{\cal S})$ with respect to the data domain is independent of $\lambda$.
    \item \textbf{Differentiability}\\
    The log-likelihood is twice differentiable with respect to $\lambda$.
    \item \textbf{Interchangeability}\\
    $~\frac{\partial}{\partial \lambda}\int p({\cal D}|\lambda) d{\cal D}=\int\frac{\partial}{\partial \lambda} p({\cal D}|\lambda)d{\cal D}$.
    \label{eq:SUPP_fisher_conditions}
\end{enumerate}
The log-likelihood of all three operational configurations is twice continuously differentiable with respect to $\lambda$, as it consists solely of polynomial and exponential terms. We now justify $\textbf{Support independence}$ and $\textbf{Interchangeability}$ for each.

%%%%%%%%%%%%%%%%%%%%%%%%%%%%%

\subsection{Fisher Conditions for M1}
\label{sec:CRLB_temporal_conditions}

$\textbf{Support independence}$\\
The possible number of events $N$ is the set $\{0,1,2,\dots,N^{\rm max}\}$. The last detection time $t_N$ must satisfy $(N-1)\tau_{\rm dead}\leq t_N\leq T$. Both $N^{\rm max}=\lceil\frac{T}{\tau_{\rm dead}}\rceil$ and $t_N$ depend on ${\cal S}_{\rm cont}$, regardless of $\lambda$.\\ 
$\textbf{Interchangeability}$\\
Since $\{t_i\}_{i=1}^N,N^{\rm max}$ for $N\leq N^{\rm max}$ are finite, 
we have a finite sum and a definite integral. Then, linearity of differentiation  allows to swap the sum and derivative signs. Then, because the integration bounds are independent of $\lambda$, the Leibniz Integral Rule (See Ref.~\cite{kaplan2002advanced} herein) lets us swap the integral and derivative signs:  
\begin{equation}
\begin{split}
    &\frac{\partial}{\partial \lambda}\left[\sum_{n=0}^{N^{\rm max}} \int_{{t}_N}\densedots\int_{{t}_1} p(N=n, \{{ t}_i\}_{i=1}^N|\lambda,{\cal S}_{\rm cont})d{t}_1\densedots d{t}_N\right]\\
    &~~~=\sum_{n=0}^{N^{\rm max}} \int_{{t}_N}\densedots\int_{{t}_1} \frac{\partial}{\partial \lambda}p(N=n, \{{ t}_i\}_{i=1}^N|\lambda,{\cal S}_{\rm cont})d{t}_1\densedots d{t}_N.
\end{split}
    \label{eq:SUPP_CRLB_temporal_interchangeability}
\end{equation}

%%%%%%%%%%%%%%%%%%%%%%%%%%%%%

\subsection{Fisher Conditions for M2}
\label{sec:CRLB_discrete_conditions}

$\textbf{Support independence}$\\
The possible number of events $N$ is the set $\{0,1,2,\dots,B\}$. B depends on ${\cal S}_{bin}$, regardless of $\lambda$.\\
$\textbf{Interchangeability}$\\
Since $B$ is finite, we have a finite sum. Then, linearity of differentiation  allows to swap the sum and derivative signs:  
\begin{equation}
    \frac{\partial}{\partial \lambda}\sum_{n=0}^B p(N=n|\lambda, {\cal S}_{\rm bin})=\sum_{n=0}^B \frac{\partial}{\partial \lambda}p(N=n|\lambda, {\cal S}_{\rm bin})\;.
    \label{eq:SUPP_CRLB_discrete_interchangeability}
\end{equation}

%%%%%%%%%%%%%%%%%%%%%%%%%%%%%

\subsection{Fisher Conditions for M3}
\label{sec:CRLB_hybrid_conditions}

$\textbf{Support independence}$\\
The possible number of events $N$ is the set $\{0,1,2,\dots,B\}$. B depends on ${\cal S}_{bin}$, regardless of $\lambda$. Each timestamp ${\tilde{t}}_i\in\{{\tilde{t}}_i\}_{i=1}^N$ is within the range $[0,\tau_{\rm sense}]$, regardless of $\lambda$.\\
$\textbf{Interchangeability}$\\
Since $B,\{\tilde{t}_i\}_{i=1}^N$ for $N\leq B$ are finite, 
we have a finite sum and a definite integral. Then, linearity of differentiation  allows to swap the sum and derivative signs. Then, because the integration bounds are independent of $\lambda$, the Leibniz Integral Rule (See Ref.~\cite{kaplan2002advanced} herein) lets us swap the integral and derivative signs:
\begin{equation}
\begin{split}
    &\frac{\partial}{\partial \lambda} \sum_{n=0}^B \int_{{\tilde{t}}_N}\densedots\int_{{\tilde{t}}_1}p(N=n, \{{ \tilde{t}}_i\}_{i=1}^N|\lambda,{\cal S}_{\rm bin})d{\tilde{t}}_1\densedots d{\tilde{t}}_N\\
    &~~~=\sum_{n=0}^B \int_{{\tilde{t}}_N}\densedots\int_{{\tilde{t}}_1}\frac{\partial}{\partial \lambda} p(N=n, \{{ \tilde{t}}_i\}_{i=1}^N|\lambda,{\cal S}_{\rm bin})d{\tilde{t}}_1\densedots d\tilde{t}_N \;.
\end{split}
\label{eq:SUPP_CRLB_hybrid_interchangeability}
\end{equation}  

%%%%%%%%%%%%%%%%%%%%%%%%%%%%%%%%%%%%%%
\section{Mismatched Recovery}
\label{sec:CRLB_comparison_empirical_mismatch}

\begin{figure}[t] 
     \centering
     \includegraphics[width=\columnwidth]{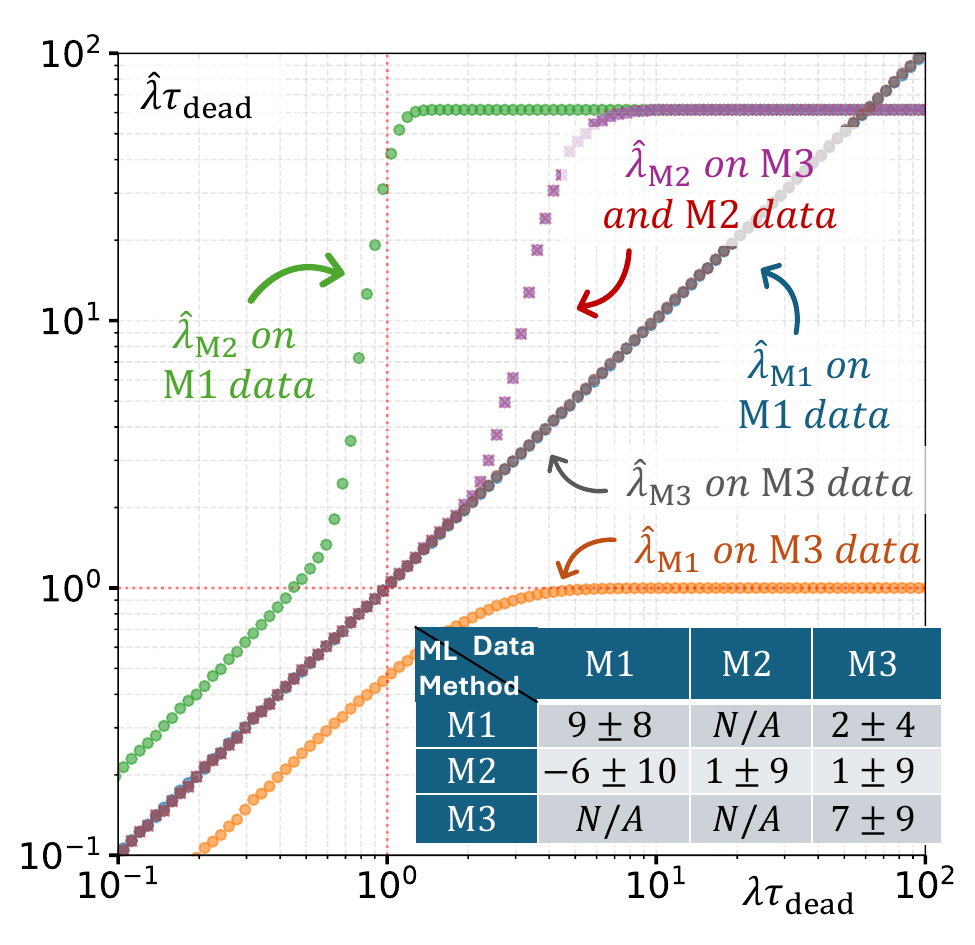}
     \caption{Numerical results. Different ML estimation methods are 
     applied to data acquired in different combinations. PSNR comparison is attached.
     }
     \label{fig:empirical_mismatch}
\end{figure}

In this section, we compare recovery of the event rate, when pairing Maximum Likelihood (ML) estimation methods with inconsistent operation mode data. We use ML estimation formulas from Sec.~5 of the main manuscript for results shown in Sec.~6 of the main manuscript. Then, we calculate the mean recovered value ${\hat \lambda}\tau_{\rm dead}$. We pair all possible combinations of estimation modes and data. {\bf M2} data lack time measurements needed for $\hat \lambda_{\rm M1},\hat \lambda_{\rm M3}$. {\bf M1} data lack time measurements needed for $\hat \lambda_{\rm M3}$. 
However, $\hat \lambda_{\rm M2}$ analysis can ignore timing data and run on {\bf M1} data. Comparison is depicted in Fig.~\ref{fig:empirical_mismatch} herein. 

In two cases, mismatch creates a very large bias relative to a properly matched solution, with an early onset of saturation for $\hat \lambda_{\rm M2}$ when run on {\bf M1} data. 
We now consider analysis using the {\bf M2} operation mode. A high event rate increases the probability of saturation, where all binary bins contain an event detection ($N=B$). When $N=B$, the logarithmic term in Eq.~(49) of the main manuscript is singular and cannot be used. For a sufficiently high event rate, all data are saturated. For this rate and beyond, no point is plotted in Fig.~\ref{fig:empirical_mismatch} herein. Right before the point of saturation, when $N=B-1$, the ML estimate $\hat \lambda_{\rm M2}$ converges to
\begin{equation}
\begin{split}
    \hat \lambda_{\rm M2}\tau_{\rm dead}&\xrightarrow{\lambda\tau_{\rm dead}\gg1}\frac{\tau_{\rm dead}}{\tau_{\rm sense}}\log B\\[5pt]
    &=\frac{\tau_{\rm dead}}{\tau_{\rm sense}}\log\left(\frac{T}{\tau_{\rm sense}+\tau_{\rm dead}}\right).    
\end{split}
    \label{eq:SUPP_M2_converge}
\end{equation}
Let $\tau_{\rm sense}=\tau_{\rm dead}$. Consider a {\em medium event rate} configuration from Tab.~\ref{tab:event_configs} herein. Then, $\hat \lambda_{\rm M2}\tau_{\rm dead}\xrightarrow{\lambda\tau_{\rm dead}\gg1}3.9$. This result coincides with the converging values of $\hat \lambda_{\rm M2}\tau_{\rm dead}$ in Fig.~\ref{fig:empirical_mismatch} herein.

\begin{table}[t]
\centering
\small{
\caption{Event rate configurations and parameters}
\label{tab:event_configs}
\begin{tabular}{l|c|r}
%\hline
Scenario & lux$_{\rm ref}$~[lm/m$^2$] & $T$~[s]~~~~~ \\ \hline
High Event Rate     & 400                  & $200\times10^{-9}$       \\ %\hline
Medium Event Rate    & 4                    & $10\times10^{-6}$    \\ %\hline
Low Event Rate  & $5 \times 10^{-3}$   & $10\times10^{-3}$       % \\ \hline
\end{tabular}
}
\end{table}

We now look at analysis by the {\bf M1} mode, when applied on {\bf M3} data. When the normalized event rate $\lambda\tau_{\rm dead}$ becomes very high, the probability rises for all binary bins to contain an event ($N=B$). Moreover, the probability to detect an event early during $\tau_{\rm sense}$ rises. Particularly, the probability to detect the last event at time $T-(\tau_{\rm sense}+\tau_{\rm dead})$ rises. We note that $t_N=T-(\tau_{\rm sense}+\tau_{\rm dead})<T-\tau_{\rm dead}$. 
This means that {\bf M1} analysis operates in {\tt Case I}. From Eq.~(48) of the main manuscript, 
\begin{equation}
\begin{split}
    \hat{\lambda}_{\rm M1}\tau_{\rm dead}&\xrightarrow{\lambda\tau_{\rm dead} \gg1}\frac{B\tau_{\rm dead}}{T-B\tau_{\rm dead}}=\frac{\tau_{\rm dead}}{T/B-\tau_{\rm dead}}\\[5pt]
    &=\frac{\tau_{\rm dead}}{(\tau_{\rm sense}+\tau_{\rm dead})-\tau_{\rm dead}}=\frac{\tau_{\rm dead}}{\tau_{\rm sense}} 
    \;.
\end{split}
\label{eq:SUPP_M1_converge}
\end{equation}
In the simulation configuration, $\tau_{\rm sense}=\tau_{\rm dead}$. So, from Eq.~(\ref{eq:SUPP_M1_converge}) herein: $\hat{\lambda}_{\rm M1}\tau_{\rm dead}\xrightarrow{\lambda\tau_{\rm dead}\gg1}1$. This coincides with Fig.~\ref{fig:empirical_mismatch} herein. A table summarizing PSNR results for the mismatched recovery is attached to Fig.~\ref{fig:empirical_mismatch} herein. The table shows how quality  is statistically highest when analysis methods match the data mode.  

%%%%%%%%%%%%%%%%%%%%%%%%%%%%%%%%%%%%%%%%%%%%%%%
\section{ML Estimation Bias}
\label{sec:ml_bias}

In this section we show that ML based on SPAD data (Sec.~5 of the main manuscript) is slightly biased. Then, we describe a method to compensate for the bias. Similarly to Sec.~6~({\em Numerical Examples}) of the main manuscript, we use 
%create $100$ 
 different values of $\lambda\tau_{\rm dead}$, the range of $[ 10^{-2},10^{2}]$. Recall the three operation modes from Sec.~3.1 of the main manuscript. For each value of $\lambda$ and operation mode (sensor system) ${\cal S}$, we simulate $N^{\rm sample}$ random data streams. A random data stream is denoted
${\cal D}_{\iota}(\lambda|{\cal S})$, where $\iota=[1\ldots N^{\rm sample}]$. 

Then, we recover corresponding values 
${\hat \lambda}^{({\iota})}_{\cal S}$, using the matching ML estimators from Sec.~5 of the main manuscript. The numerical mean 
of the estimators is
\begin{equation}
    \langle {\hat \lambda}_{\cal S} \rangle 
    =\frac{1}{N^{\rm sample}}
    \sum_{\iota=1}^{N^{\rm sample}}
    {\hat \lambda}^{({\iota})}_{\cal S}
    \;.
    \label{eq:SUPP_meanML}
\end{equation}
The relative bias is 
\begin{equation}
    \chi(\lambda,{\cal S})=
    \frac{\langle {\hat \lambda}_{\cal S} \rangle - \lambda}
        {\lambda}
    \;.
    \label{eq:SUPP_chi}
\end{equation}

\begin{figure}[t]
    \centering
    % Top Plot
    \includegraphics[width=1.0\columnwidth]{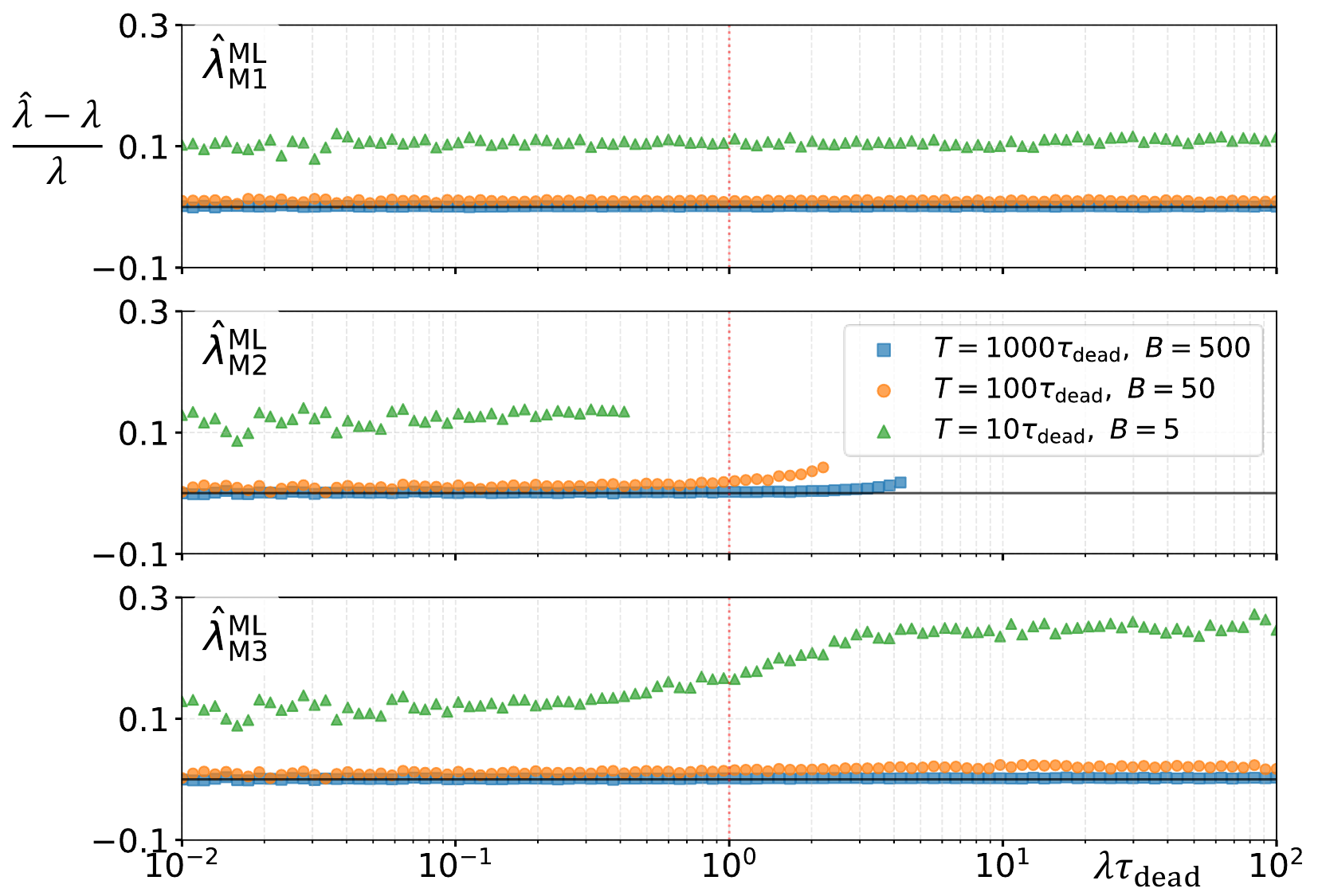}\\
    % Middle Plot
    \caption{
    Relative error of ML estimators recovery. From top to bottom: {\bf M1} estimation from Eq.~(48) of the main manuscript; {\bf M2} estimation from Eq.~(49) of the main manuscript; {\bf M3} estimation from Eq.~(50) of the main manuscript.}
    \label{fig:ml_bias}
\end{figure}

The relative bias is plotted in Fig.~\ref{fig:ml_bias} herein, for various settings. The bias is non-negative, small, and it becomes negligible when the number of events increases. Moreover, it is {\em systematic}. Furthermore, from the examples here, the bias is monotonous: estimation bias increases $\hat \lambda$ as $\lambda$ increases.  From monotonicity, this bias is an invertible function. Therefore, it is possible to {\em de-bias} an ML estimator, i.e., compensate for the bias. The de-biased estimate is denoted 
${\hat \lambda}^{\rm DB}_{\cal S}$. The de-biasing function can be pre-calibrated from examples.  

\begin{figure}[t]
    \centering
    % Top Plot
    \includegraphics[width=1.0\columnwidth]{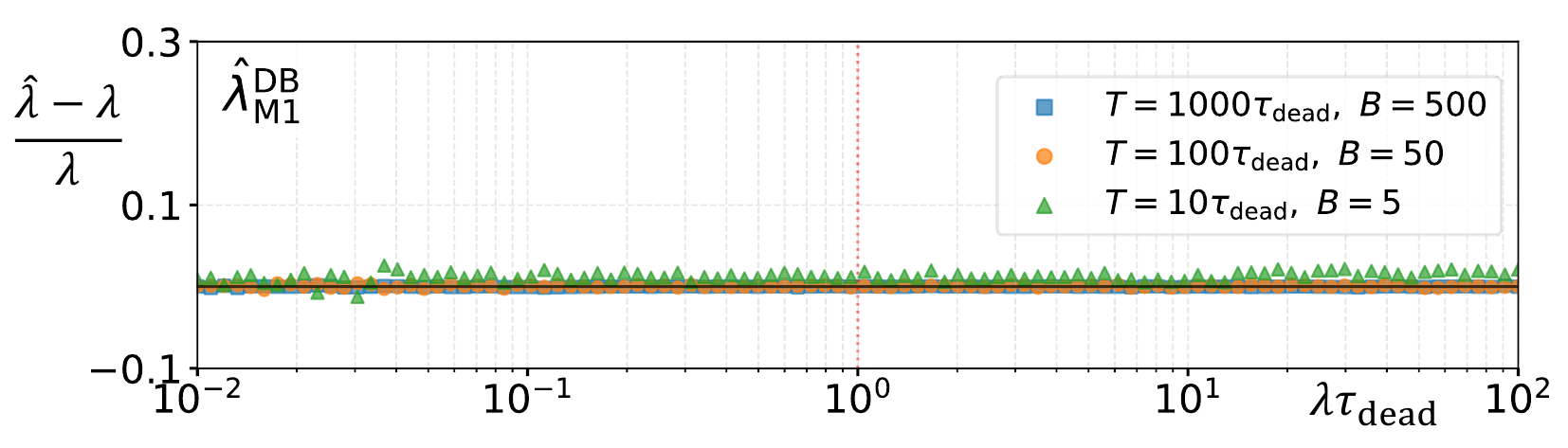}\\
    % Middle Plot
    \caption{
    {\bf M1}-based ML estimators are slightly biased
    in Fig.~\ref{fig:ml_bias} herein. So, here they are de-biased using a model learned from $T=1000\tau_{\rm dead}$ data. The relative bias of the de-biased results show generalization of de-bias to data having $T=\{100\tau_{\rm dead},10\tau_{\rm dead}\}$.}
    \label{fig:ml_bias_fix}
\end{figure}

In Fig.~\ref{fig:ml_bias_fix} herein, we show an example of de-biased ${\hat \lambda}^{\rm DB}_{\rm M1}$. We model the bias function as $c_{\rm M1}{\tau_{\rm dead}}{/T}$, where ${c}_{\rm M1}$ is a parameter. The relative bias examples $\chi(\lambda,{\rm M1})$ for $T=1000\tau_{\rm dead}$ are used to train this parameter.  Specifically, we calculate the mean over $\chi(\lambda,{\rm M1})$, and extract ${c}_{\rm M1}$:
\begin{equation}
    \langle\chi(\lambda,{\rm M1})\rangle=c_{\rm M1}\frac{\tau_{\rm dead}}{T}
    ~\Rightarrow~ c_{\rm M1}=\langle\chi(\lambda,{\rm M1})\rangle\frac{T}{\tau_{\rm dead}}\;.
    \label{eq:SUPP_M1_scale}
\end{equation}
Then, we show that this function generalizes: Using  $\{T=100\tau_{\rm dead}, T=10\tau_{\rm dead}\}$ as test sets, bias diminishes.

%%%%%%%%%%%%%%%%%%%%%%%%%%%%%%%%%%%%%%%%%%%%%%%
\section{Simulation Details}
\label{sec:sim_detail}

In this section we detail the simulation process. We describe  simulation in training, testing and the chosen configurations.

%%%%%%%%%%%%%%%%%%%%%%%%%%%%%%%%%%%%%%%%%%%%%%%
\subsection{Training}
\label{sec:sim_train}

Training was based on Flickr-Faces-HQ  (FFHQ) data, from Ref.~~\cite{karras2019style} herein. This set contains 70,000 RGB face images in $1024\times 1024$ resolution. We resized each image to $256\times 256$ resolution, and converted to a $[0,1]$ grayscale range. Out of this data, we used 69,000 images to train the score DNN of the prior, $\bm{s}_{\bm{\theta}}$. Training used $8.5\cdot 10^5$ steps, on the hardware described in Sec.~10 of the main manuscript. We used the other images for testing. 

%%%%%%%%%%%%%%%%%%%%%%%%%%%%%%%%%%%%%%%%%%%%%%%
\subsection{Simulated Data}
\label{sec:sim_data}

Testing requires event data. Generating event data has several steps. First, a method from Ref.~\cite{suonsivu2021time} herein simulates an expected event  rate. Take a grayscale image pixel, whose intensity $I$ is in the range $[0,1]$. Let the speed of light, Plank constant and the pixel area be denoted, respectively, $c,h$ and $A$. Set a reference wavelength $l$ and a corresponding luminous efficiency ${\rm lum}_{\rm eff}$~[lm/W]. The scene is scaled by a reference illuminance ${\rm lux}_{\rm ref}$~[lm/$m^2$]. Then, the simulated photon flux is 
\begin{equation}
    \Phi=\frac{{\rm lux}_{\rm ref} lA} {{\rm lum}_{\rm eff} hc} I 
    ~~~~~[{\rm photons/s]}
    \;.
    \label{eq:SUPP_SIM_lambda}
\end{equation} 
In Sec.~3.1 of the main manuscript, $\Phi$ is converted to the expected event rate 
$\lambda$ at the pixel. 

The expected event rate per pixel is then used in one of several simulators of ${\cal F}$, depending on the operation mode of Sec.~2.1 on the main manuscript. For the mode of {\em Continuous Time Readouts and Domain},  a sequence of detection events is  simulated, by an algorithm we implemented based on Ref.~\cite{suonsivu2021time} herein. This algorithm also incorporates a dark count rate (DCR), probability of after-pulsing (PAP) and timing jitter, to simulate measurement noise. We use  $A=2.5\times10^{-11} ~[{\rm m}^2]$, $q=0.9$, $l=555$~[nm], 
${\rm lum}_{\rm eff}=683$~[lm/W], ${\rm DCR}=100$~[Hz], ${\rm PAP}=0.005$, ${\text{timing jitter}}=200$~[ps], $\tau_{\rm dead}=100$~[ns]. Values are based on commercially available sensors such as Ref.~\cite{piimaging2026spadlambda} herein.
%Fig.~\ref{fig:cameraman} shows an example, using different values of ${\rm lux}_{\rm ref}$ and a fixed exposure time $T$. For this visualization, we use a simple reconstruction method from Ref.~\cite{ingle2019high}. 
This simulator is sequential. Hence, the number of simulated events is limited, in practice, by computer speed. We allocated up to a few minutes to simulate an image.

%%%%%%%%%%%%%%%%%%%%%%%%%%%%%%%%%%%%%%%%%%%%%%%
\subsection{Configurations}
\label{sec:sim_configs}

We created simulations that implement the operation modes {\em Continuous Time Readouts in Discrete Time Bins} and {\em Binary Readouts in Discrete Time Bins}.
They account for $\tau_{\rm sense}$ in addition to $\tau_{\rm dead}$. We use $\tau_{\rm sense}=100$~[ns].  
We studied three rate scenarios, detailed in Table~\ref{tab:event_configs} herein. We set the exposure time $T$, so that at least one event is produced for most pixels.

%%%%%%%%%%%%%%%%%%%%%%%%%%%%%%%%%%%%%%%%%%%%%%%
\subsection{Calculating \texorpdfstring{$\rho$ and $\zeta$}{rho and zeta}}
\label{sec:sim_hyper}

Out of the face data described in Sec.~\ref{sec:sim_train} herein, we used a subset of 20 faces that had not trained  $\bm{s}_{\bm{\theta}}$. We then ran Alg.~1 of the main manuscript on this subset, per rate scenario, for a variety of $\rho$ values. We eventually settled on a value that gave us the best subjective visual results on this subset, per rate scenario. For the high, medium and low event rate scenarios,   we used, respectively, $\rho=5\times10^4$, $15\times10^{-4}$ and $18\times10^{-4}$. 

Setting  $\zeta$ is done by substituting $I=1$ in Eq.~(\ref{eq:SUPP_SIM_lambda}) herein, and then using the value yielded by the conversion to event rate in Sec.~3.1 of the main manuscript.

%%%%%%%%%%%%%%%%%%%%%%%%%%%%%%%%%%%%%%%%%%%%%%%
\subsection{Simulation Results}
\label{sec:sim_results}

\begin{figure*}[t]
    \centering
    \includegraphics[width=1.0\textwidth]{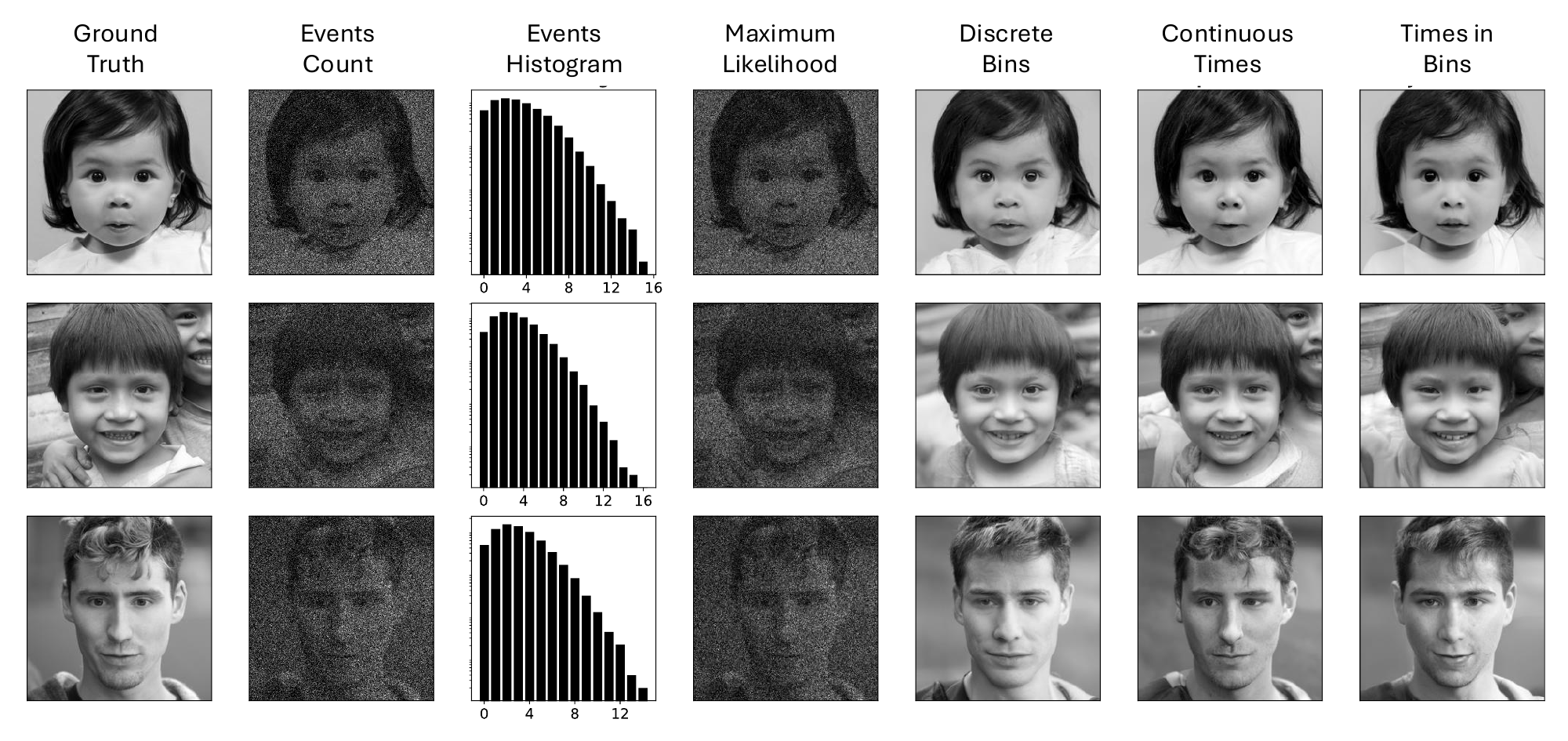}
    \caption{
    Low event rate reconstruction simulation results.
    From left to right: Ground truth image; Raw events count;  Histogram of events per pixel; ML based on {\em Continuous Time Readouts and Domain}; SBD using {\em Binary Readouts in Discrete Time Bins}. SBD using with {\em Continuous Time Readouts and Domain}. SBD using {\em Continuous Time Readouts in Discrete Time Bins}.}
    \label{fig:low_recon}
\end{figure*}

\begin{figure*}[t]
    \centering
    \includegraphics[width=1.0\textwidth]{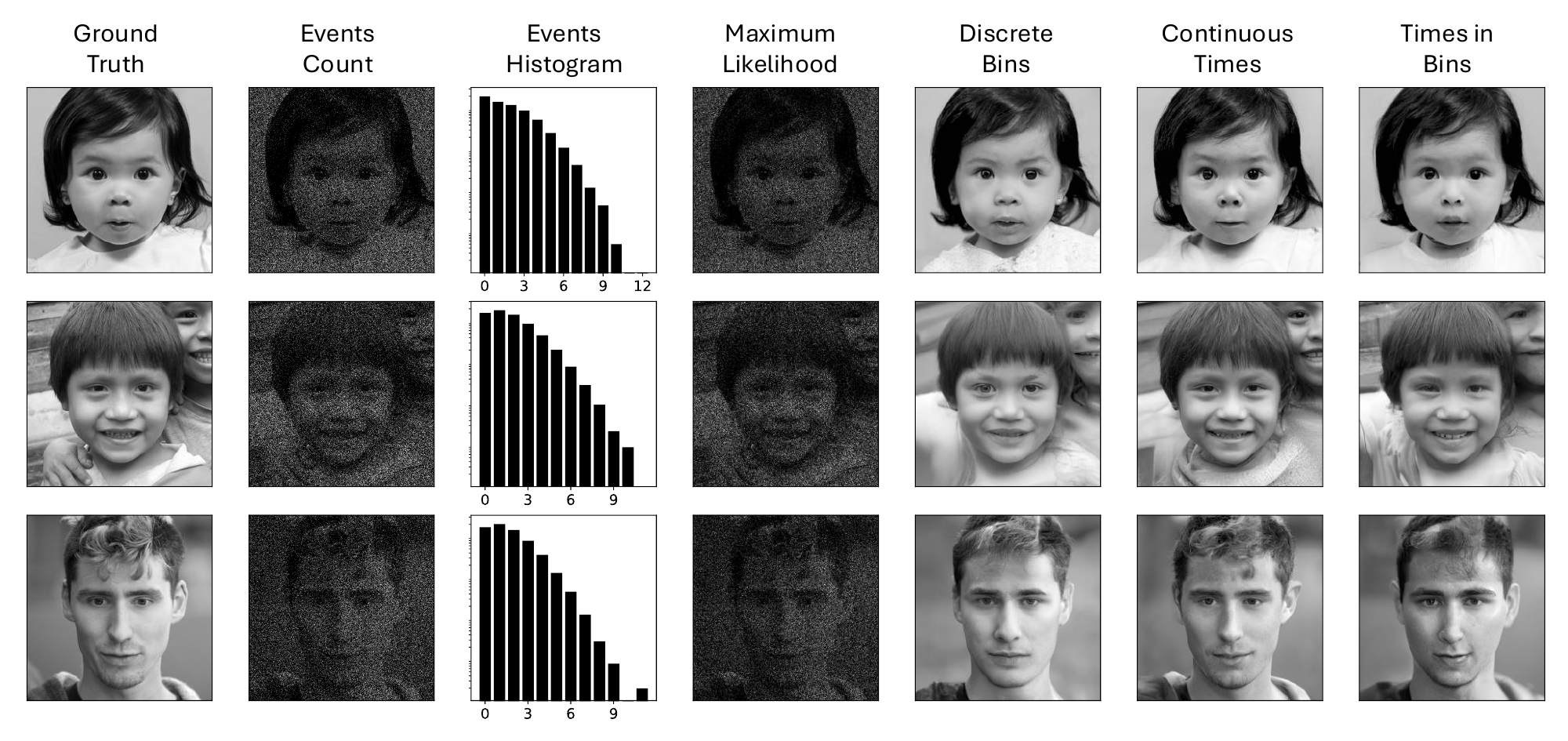}\\
    \caption{
    Medium event rate reconstruction simulation results.
    From left to right: Ground truth image; Raw events count;  Histogram of events per pixel; ML based on {\em Continuous Time Readouts and Domain}; SBD using {\em Binary Readouts in Discrete Time Bins}. SBD using with {\em Continuous Time Readouts and Domain}. SBD using {\em Continuous Time Readouts in Discrete Time Bins}.}
    \label{fig:mid_recon}
\end{figure*}

\begin{figure*}[t]
    \centering
    % Top Plot
    \includegraphics[width=1.0\textwidth]{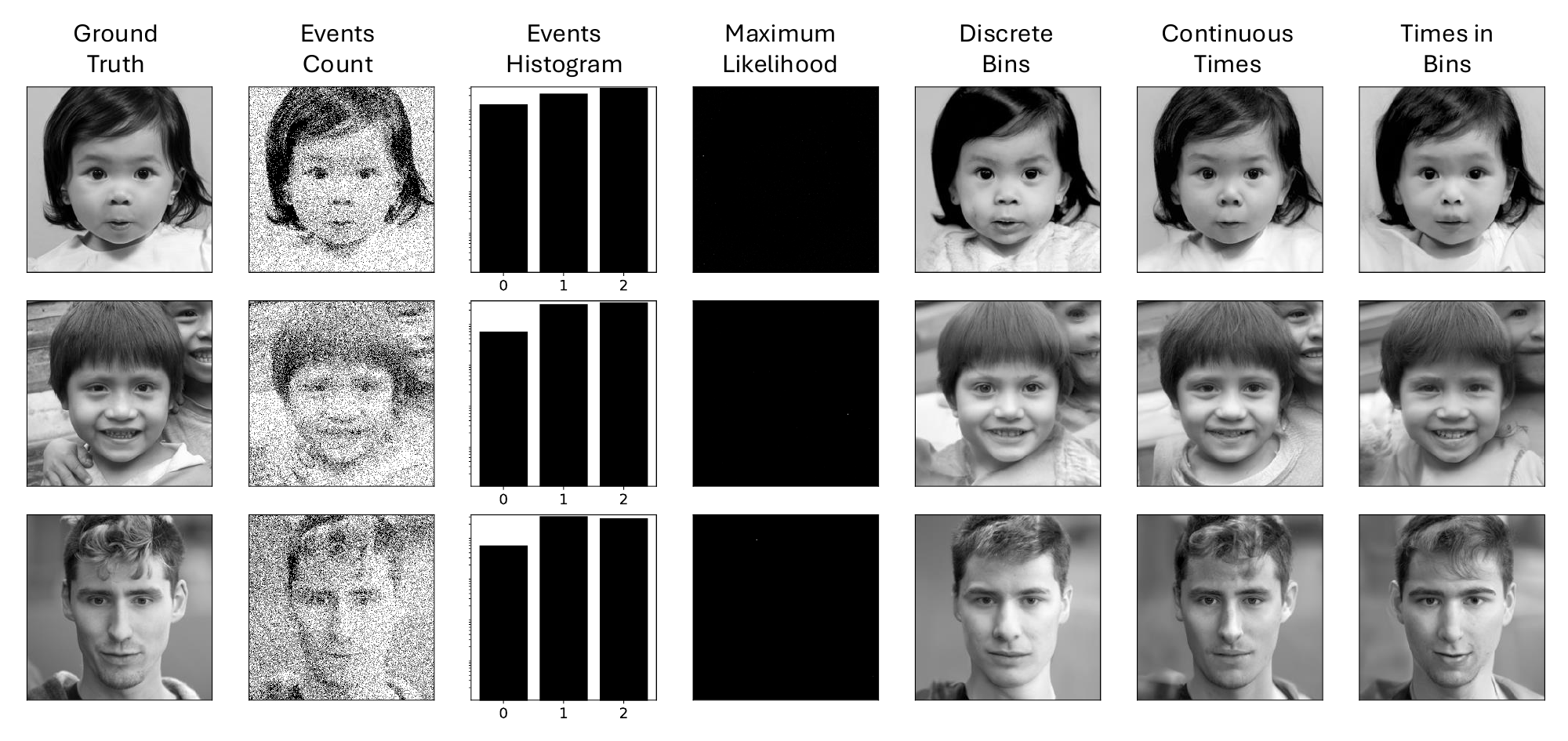}\\
    % Middle Plot
    \caption{
    High event rate reconstruction simulation results.
    From left to right: Ground truth image; Raw events count;  Histogram of events per pixel; ML based on {\em Continuous Time Readouts and Domain}; SBD using {\em Binary Readouts in Discrete Time Bins}. SBD using with {\em Continuous Time Readouts and Domain}. SBD using {\em Continuous Time Readouts in Discrete Time Bins}.}
    \label{fig:high_recon}
\end{figure*}

In this section we present more visual and statistical results for the configurations in Tab.~\ref{tab:event_configs} herein. The visual results are depicted in Figs.~\ref{fig:high_recon},\ref{fig:mid_recon},\ref{fig:low_recon} herein. Statistics are presented in Tab.~\ref{tab:SUPP_sim_results} herein.
\begin{table}[t!]
\centering
\caption{Simulations. High [Top], medium [Middle], and low [Bottom] event rate scenarios.  Continuous Times stand for {\em Continuous Time Readouts and Domain}. Discrete Bins stand for {\em Binary Readouts in Discrete Time Bins}. Times in Bins stand for {\em Continuous Time Readouts in Discrete Time Bins}.}
\label{tab:SUPP_sim_results}
% Reduced from 5.0pt to 4.2pt to pull the right edge away from the column gutter
\setlength{\tabcolsep}{4.2pt}
\small 
\begin{tabular}{@{} l 
    S[table-format=2(1), table-number-alignment=center] 
    S[table-format=1.2(1), table-number-alignment=center] 
    S[table-format=1.2(1), table-number-alignment=center] 
    S[table-format=3, table-number-alignment=center] @{}}
\toprule
Method & {PSNR\makebox[0pt][l]{$\uparrow$}} & {SSIM\makebox[0pt][l]{$\uparrow$}} & {LPIPS\makebox[0pt][l]{$\downarrow$}} & {FID\makebox[0pt][l]{$\downarrow$}} \\ 
\midrule
ML: Continuous Times & 6(2)  & 0.01(3)  & 0.72(3)  & 460 \\
SBD: Discrete Bins      & 16(3) & 0.47(1)  & 0.45(7)  &  40  \\
SBD: Times in Bins      & 20(2) & 0.59(9)  & 0.38(6)  & 42  \\ 
SBD: Continuous Times   & \bfseries 21(2) & \bfseries 0.65(8) & \bfseries 0.35(6) & \bfseries 39 \\

\specialrule{1.2pt}{2pt}{2pt} 
ML: Continuous Times & 8(2) & 0.11(4) & 0.76(5) & 305 \\
SBD: Discrete Bins      & 22(2) & 0.61(8) & 0.36(5) & 36 \\
SBD: Times in Bins      & 22(2) & 0.61(8) & 0.36(5) & 36  \\ 
SBD: Continuous Times   & \bfseries 23(2) & \bfseries 0.67(7) & \bfseries 0.33(5) & \bfseries 33 \\

\specialrule{1.2pt}{2pt}{2pt} 
ML: Continuous Times & 10(2) & 0.11(2) & 0.78(3) & 325 \\
SBD: Discrete Bins      & 22(2)  & 0.61(8) & 0.36(5) & 36 \\
SBD: Times in Bins      & 22(2) & 0.61(8) & 0.36(5) & 36 \\ 
SBD: Continuous Times   & \bfseries 23(2) & \bfseries 0.66(7) & \bfseries 0.34(5) & \bfseries 33 \\

\bottomrule
\end{tabular}
\end{table}

%%%%%%%%%%%%%%%%%%%%%%%%%%%%%%%%%%%%%%%%%%%%%%%
\subsection{Prox-DiffPir Comparison}
\label{sec:sim_proxdiffpir}

\begin{figure}[t]
    \centering
    % Top Plot
    \includegraphics[width=1.0\columnwidth]{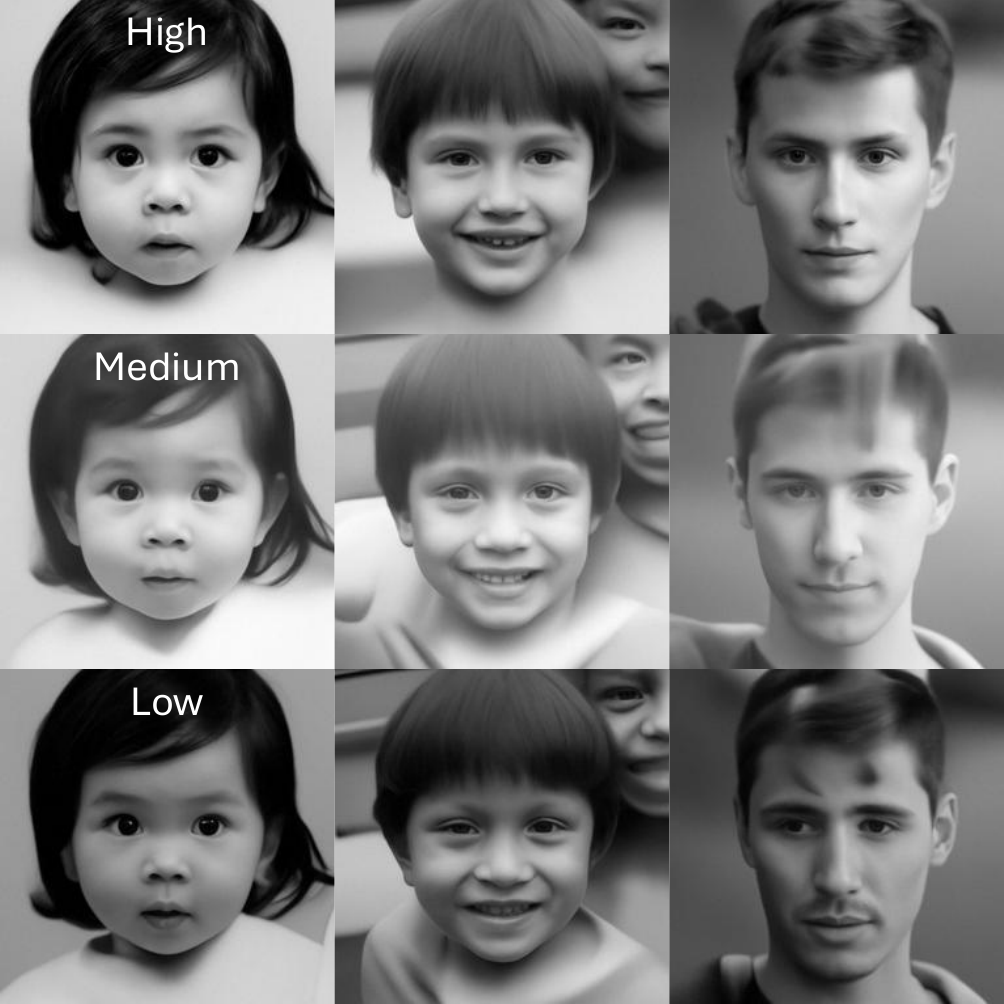}\\
    % Middle Plot
    \caption{
    Visual examples of recovery by Prox-DiffPir in simulations. Simulation configurations correspond to in Tab.~\ref{tab:event_configs} herein. From top to bottom: High event rate; Medium event rate; Low event rate.}
    \label{fig:prox_diffpir_all}
\end{figure}

In Sec.~8 of the main manuscript, we mentioned how Ref.~[28] of the main manuscript uses SBD to recover SPAD data as well. We applied the method Prox-DiffPir of Ref.~[28] of the main manuscript on the same configurations as Tab.~\ref{tab:event_configs} herein. The same test set of 900 scenes from Sec.~9 of the main manuscript is used. Examples of visual results for each event rate configuration are presented in Fig.~\ref{fig:prox_diffpir_all} herein. Statistics are presented in Tab.~\ref{tab:proxdiffpir_metrics} herein. Prox-DiffPir results in  similar statistics as the {\bf M2} operation mode, except for the FID score, which is significantly higher in Prox-DiffPir.  

% LaTeX Table: Simulation Results
\begin{table}[t!]
\centering
\caption{Prox-DiffPir recovery statistics on simulations. High [Top], medium [Middle], and low [Bottom] event rate scenarios.}
\label{tab:proxdiffpir_metrics}
\setlength{\tabcolsep}{4.2pt}
\small
\begin{tabular}{@{} l S[table-format=2(1)] S[table-format=1.2(1)] S[table-format=1.2(1)] S[table-format=3] @{}}
\toprule
Method & {PSNR\makebox[0pt][l]{$\uparrow$}} & {SSIM\makebox[0pt][l]{$\uparrow$}} & {LPIPS\makebox[0pt][l]{$\downarrow$}} & {FID\makebox[0pt][l]{$\downarrow$}} \\
\midrule
Prox-DiffPir: High &  17(2) &  0.61(8) &  0.43(6) &  81 \\
Prox-DiffPir: Medium &  22(1) &  0.64(7) &  0.40(5) &  77 \\
Prox-DiffPir: Low &  19(2) &  0.64(7) &  0.42(5) &  83 \\
\bottomrule
\end{tabular}
\end{table}

%%%%%%%%%%%%%%%%%%%%%%%%%%%%%%%%%%%%%%%%%%%%%%%
\subsection{Additional Fan Recovery}
\label{sec:sim_fan}

\begin{figure}[t!] 
    \centering
    \includegraphics[width=\columnwidth]{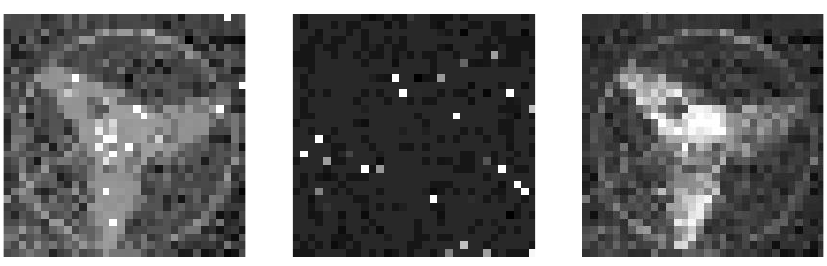}
    \caption{Recovery from real SPAD data. Exposure time $T=300$~[$\mu s$]. 
    [Left to Right]: {\bf (M2)} {\em binary Bins}; {\bf (M1)} {\em Free-running Timestamps}; {\bf (M3)} {\em Timestamped Bins}.}
    \label{fig:fans_300}
\end{figure}

This section shows additional results corresponding to the fan data of Sec.~10 of the main manuscript. The operation modes and recovery methods in the main manuscript assume a stationary event rate. However, the scene is of a rotating fan. 
% In $T=600\mu{\rm s}$, the fan hardly moves, so flux is nearly constant in most pixels. We'll also show (d), where $T=300\mu{\rm s}$ so all points move less than a pixel. That's because
A full rotation takes $\approx 1500$ frames. The frame rate is reported as $50000$ FPS. Therefore, the fan completes a rotation every $0.03$[s]. The fan radius is approximately $16$ pixels. The fan circumference is thus  $\approx 2\pi\cdot16$ pixels. A point on the fan's extremity moves a single pixel every $\frac{0.03}{2\pi\cdot16}\approx300$~[$\mu s$]. For an exposure time of $300$~[$\mu s$], the fan to be assumed stationary. The main manuscript shows results using $T=600$~[$\mu s$]. Here we produce visually consistent results using $T=300$~[$\mu s$], as depicted in Fig.~\ref{fig:fans_300} herein.

% Any acknowledgments to only be included in camera ready
\ifpeerreview \else
\section*{Acknowledgments}
We thank Andreas Velten, Atul Ingle and Trevor Seets for their hard work providing the Fan and Tunnel data, and the support. We thank Vadim Holodovsky, Ina Talmon and Miri Haramati for their invaluable assistance. Yoav Schechner is the Mark and Diane
Seiden Chair in Science at the Technion. He is a Landau
Fellow supported by the Taub Foundation. His work was
conducted in the Ollendorff Minerva Center. Minvera is
funded through the BMBF. Mohit Gupta was supported by the National Science Foundation (CAREER Award \#1943149), the Office of Naval Research (N000142412155), and by a Sony Faculty Innovation Award. This work was supported by the PMRI – Peter Munk Research Institute - Technion, and by KLA. 
\fi

% \bibliographystyle{IEEEtran}
% \bibliography{supp_ref}

\end{document}